\documentclass[manuscript,screen]{acmart}

\usepackage{overpic} 
\usepackage{tikz}
\usepackage[edges]{forest}
\definecolor{hiddendraw}{RGB}{205, 44, 36}
\definecolor{hidden-blue}{RGB}{194,232,247}
\definecolor{hidden-orange}{RGB}{243,202,120}
\definecolor{hidden-yellow}{RGB}{242,244,193}

\usepackage{amsmath}
\usepackage{amsthm}
\usepackage{booktabs}
\usepackage{algorithm}
\usepackage{algorithmic}
\usepackage[switch]{lineno}
\usepackage{tablefootnote}
\usepackage{float}     
\usepackage{subfig}   
\usepackage{overpic}
\usepackage{subcaption}
\usepackage{multirow} 
\usepackage{amsmath}
\usepackage{newfloat}
\usepackage{listings}

\usepackage{color}

\AtBeginDocument{%
  \providecommand\BibTeX{{%
    \normalfont B\kern-0.5em{\scshape i\kern-0.25em b}\kern-0.8em\TeX}}}

\setcopyright{acmlicensed}
\copyrightyear{2018}
\acmYear{2018}
\acmDOI{XXXXXXX.XXXXXXX}

\acmConference[Conference acronym 'XX]{Make sure to enter the correct
  conference title from your rights confirmation emai}{June 03--05,
  2018}{Woodstock, NY}
\acmISBN{978-1-4503-XXXX-X/18/06}

\begin{document}

\title{Adapting Large Language Models for Education:\\
Foundational Capabilities, Potentials, and Challenges}

\author{Qingyao Li}
\email{ly890306@sjtu.edu.cn}

\author{Lingyue Fu}
\affiliation{%
  \institution{Shanghai Jiao Tong University}
  \country{China}
}

\author{Weiming Zhang}
\author{Xianyu Chen}
\affiliation{%
  \institution{Shanghai Jiao Tong University}
  \country{China}}

\author{Jingwei Yu}
\affiliation{%
  \institution{Shanghai Jiao Tong University}
  \country{China}
}

\author{Wei Xia$^\dagger$}
\affiliation{%
 \institution{Huawei Noah's Ark Lab}
 \country{China}}
\email{xiawei24@huawei.com}

\author{Weinan Zhang$^\dagger$}
\affiliation{%
  \institution{Shanghai Jiao Tong University}
  \country{China}}
\email{wnzhang@sjtu.edu.cn}

\author{Ruiming Tang}
\affiliation{%
  \institution{Huawei Noah's Ark Lab}
  \country{China}}

\author{Yong Yu$^\dagger$}
\affiliation{%
  \institution{Shanghai Jiao Tong University}
  \country{China}}
\email{yyu@sjtu.edu.cn}

\renewcommand{\shortauthors}{Q.Li, et al.}

\begin{abstract}
Online education platforms, leveraging the internet to distribute education resources, seek to provide convenient education but often fall short in real-time communication with students. They often struggle to address the diverse obstacles students encounter throughout their learning journey. 
Solving the problems encountered by students poses a significant challenge for traditional deep learning models, as it requires not only a broad spectrum of subject knowledge but also the ability to understand what constitutes a student's individual difficulties. It's challenging for traditional machine learning models, as they lack the capacity to comprehend students' personalized needs.
Recently, the emergence of large language models (LLMs) offers the possibility for resolving this issue by comprehending individual requests. Although LLMs have been successful in various fields, creating an LLM-based education system is still challenging for the wide range of educational skills required. This paper reviews the recently emerged LLM researches related to educational capabilities, including mathematics, writing, programming, reasoning, and knowledge-based question answering, with the aim to explore their potential in constructing the next-generation intelligent education system. Specifically, for each capability, we focus on investigating two aspects. Firstly, we examine the current state of LLMs regarding this capability: how advanced they have become, whether they surpass human abilities, and what deficiencies might exist. Secondly, we evaluate whether the development methods for LLMs in this area are generalizable—that is, whether these methods can be applied to construct a comprehensive educational supermodel with strengths across various capabilities, rather than being effective in only a singular aspect. Based on the current development status, we further outline two approaches for an LLM-based education system: a unified approach and a mixture-of-expert (MoE) approach. Finally, we explore the challenges and future directions, providing new research opportunities and perspectives on adapting LLMs for education.
  \let\thefootnote\relax\footnotetext{$\dagger$ Weinan Zhang, Yong Yu and Wei Xia are the co-corresponding authors.}
\end{abstract}

\ccsdesc[500]{Information systems}
\ccsdesc[500]{Applied computing~Education}



\keywords{Large Language Models, Educational Data Mining}


\maketitle

\section{Introduction}
Education plays a vital role in shaping individuals' futures as it forms the foundation for providing people with knowledge, skills, and critical thinking abilities. Conventional education systems heavily rely on teachers for imparting knowledge to students, which place a significant demand on educational resources. However, the advent of online education has substantially lowered the cost of accessing these educational materials. Many people are conveniently acquiring knowledge through online courses and exercises. 

Much effort has been made to achieve personalized learning in online-education~\citep{abdelrahman2023knowledge, li2023graph, gong2020attentional}. Most of these methods are based on predicting students' knowledge states or recommending personalized learning resources using neural networks, based on sequences of student behaviors. However, this approach achieves only a coarse level of personalization. Even when students receive recommended resources, the specific difficulties they encounter in their learning process can remain unresolved. These difficulties may vary from student to student; for example, different students might struggle with different aspects of the same problem, such as misconceptions in understanding key concepts or difficulties in the reasoning process. These issues require detailed descriptions from each student for educators to understand. However, current online education systems face the issue of not being able to interact with students in real-time like a teacher. Presently, online education platforms typically offer static course videos and exercises, leaving students unable to ask questions or seek dynamic solutions to their specific issues. Therefore, developing a teaching assistant model capable of addressing students' individual concerns represents a crucial step forward in advancing online education.

The emergence of large language models (LLMs) instills optimism in creating an intelligent education system. Since the launch of ChatGPT, LLMs have shown exceptional ability in understanding human knowledge and have been widely applied in various professional fields, including recommendation systems~\citep{lin2023can}, healthcare~\citep{liu2023qilin}, economics~\citep{li2023large}, and others. By introducing expansive world knowledge, considerable reasoning capabilities, and the ability to understand human language, LLMs have the potential to introduce new forms of interactions and methodologies across these fields. However, the challenge intensifies when constructing an LLM-based education system. To solve the specific problems encountered by students, it's not only necessary for LLMs to understand the precise issues faced by the students but also essential for them to possess specialized educational-related knowledge and skills. Only then they would be able to address problems that the students themselves might not be able to resolve.
 
The act of LLMs answering students' questions can be considered as a process that involves using multiple educational capabilities. For instance, as depicted in Figure~\ref{fig-caps}, the question posed by the student requires the education assistant to possess both mathematics and programming skills simultaneously. Therefore, summarizing the methods of developing LLMs' education-related capabilities is meaningful for building the next-generation intelligent education system. In this paper, we study LLMs from the perspective of these education-related capabilities and explore the potential of an LLM-based education system, aiming to provide valuable insights into the areas that need further improvement. We present our primary investigation in Figure~\ref{fig:summary}. Our specific focus centers on five distinct capabilities of LLMs:

\begin{figure}[tbp]
  \includegraphics[width=0.98\textwidth]{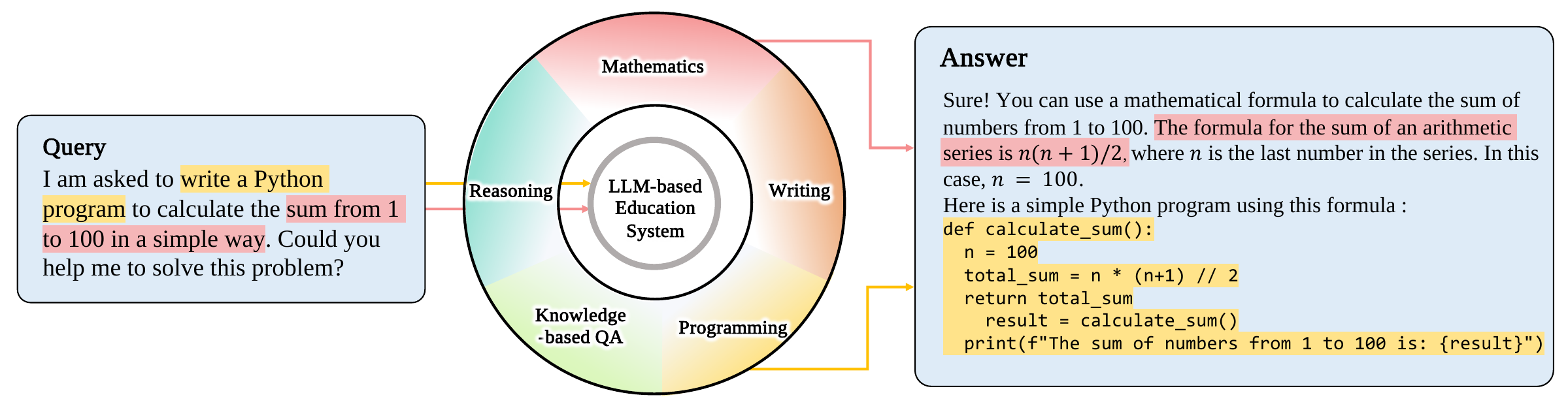}
  \caption{An example of LLM-based education systems integrating multiple abilities to solve student problems.}
  \label{fig-caps}
  \vspace{-4mm}
\end{figure}

\begin{itemize}
    \item \textbf{Mathematics:} We primarily summarize the current development of LLMs in solving mathematical problems of varying difficulties and types, encompassing basic arithmetic operations, challenging problems, multi-modal problems, and math proof.
    \item \textbf{Writing:} We investigate the performance of LLMs on a number of representative writing tasks to outline the problems LLMs face and potential future directions.
    \item \textbf{Programming:} In accordance with human programming conventions, we divide LLMs' programming process into two stages: code writing and code refinement. We review the researches in this area and summarize the remaining problems for LLMs in coding.
    \item \textbf{Reasoning:} We explore the capability of LLMs to perform reasoning in various ways, including supervised fine-tuning, prompt engineering, and hybrid methods, and explore their potential applications in the field of education.
    \item \textbf{Knowledge-based Question Answering:} We investigate the development of LLMs in open-domain and domain-specific knowledge-based question answering. We hope to offer a comprehensive view of incorporating such capabilities into the education system.
\end{itemize}

These five capabilities form the foundation of an LLM-based education system. Building upon this foundation, we propose two potential approaches for forming an LLM-based education system. One approach involves training a comprehensive language model with multiple capabilities, while the other is based on an LLM controller in a mixture-of-experts framework. 

The rest of this paper is organized as follows. In section \ref{background}, we briefly introduce the education tasks and educational LLMs. In section \ref{capabilities}, we summarize the current development status of five foundational abilities related to education, followed by discussion of the development trend of education-related LLM capabilities. In section \ref{overalldevel}, we investigate the performance of well-known LLMs in education-related capabilities. In section \ref{potential}, we introduce the potential approaches for organizing an LLM-based education system. Finally, in section \ref{future} and section \ref{conclusion}, we highlight the challenges and future directions for designing an LLM-based education system and concludes this survey.

\tikzstyle{mybox}=[
    rectangle,
    draw=hiddendraw,
    rounded corners,
    text opacity=1,
    minimum height=2em,
    inner sep=2pt,
    align=left,
    fill opacity=.5,
    ]
    
\tikzstyle{leaf}=[mybox,minimum height=1em,
fill=hidden-blue!40, text width=12em,  text=black,align=left,font=\small,
inner xsep=2pt,
inner ysep=1pt,
]

\begin{figure*}[tp]
  \centering
  \begin{forest}
    forked edges,
    for tree={
      grow=east,
      reversed=true,
      anchor=base west,
      parent anchor=east,
      child anchor=west,
      base=left,
      font=\small,
      rectangle,
      draw=hiddendraw,
      align=left,
      minimum width=2.5em,
      s sep=6pt,
      inner xsep=3pt,
      inner ysep=1pt,
      ver/.style={rotate=90, child anchor=north, parent anchor=south, anchor=center},
    },
    where level=1{text width=5.6em,font=\small}{},
    where level=2{text width=5.3em,font=\small}{},
    where level=3{text width=9.7em,font=\small}{},
    [Foundational\\ Capabilities,text width=6.3em
        [Mathematics, text width=5.5em
            [Basic Arithmetic Problems, text width=10em
               [Goat~\citep{liu2023goat}{, }GENBERT~\citep{geva2020injecting},leaf, text width=11.2em]
            ]
            [Challenging Mathematics Problems,text width=13.3em
               [WizardMath~\citep{luo2023wizardmath}{, }MeatMath~\citep{yu2023metamath},leaf, text width=13.5em]
            ]
            [Multimodal Problems,text width=8.3em
               [Mathvista~\citep{lu2023mathvista}{, }Unigeo~\citep{chen2022unigeo},leaf, text width=11.5em]
            ]
            [Mathematical Proof,text width=8em
               [LeanDojo~\citep{yang2023leandojo}{, }LEGO-Prover~\citep{xin2023lego},leaf, text width=13.5em]
            ]
        ]
        [Writing,text width=4em
            [Text Summarization,text width=8.0em
               [BRIO~\citep{liu2022brio}{, }PROM~\citep{ma2023prom},leaf, text width=10em]
            ]
            [Grammatical Error Correction,text width=11.5em
                [CoEdit~\citep{liu2022brio}{, }GrammarGPT~\citep{fan2023grammargpt},leaf, text width=12em]
            ]
        ]
       [Programming,text width=6em
            [Code Writing, text width=5.2em
               [CodeLlama~\citep{rozière2023codellama}{, }WizardCoder~\citep{luo2023wizardcoder},leaf, text width=13.5em]
            ]
            [Code Refinement, text width=7.2em
                [Codet~\citep{chen2022codet}{, }  SEIDR~\citep{fullyautonomousp},leaf, text width=10em] 
            ]
       ]
      [Reasoning, text width= 5em
            [Supervised Fine-tuning,text width=8.6em
               [CAGE~\citep{rajani2019explain}{, }Scratchpad~\citep{nye2021show},leaf, text width=11.3em]
            ]
            [Prompt Engineering,text width=7.8em
                [CoT~\citep{wei2022chain}{, }SelfConsistency~\citep{wang2022self},leaf, text width=12em] 
            ]
            [Hybrid Methodologies,text width=8.3em
                [FineTuneCoT~\citep{ho2022large}{, }STaR~\citep{zelikman2022star},leaf]
            ]
      ]
      [Knowledge-based QA , text width= 8.3em
            [Open-domain QA , text width= 7.5em
               [Retro~\citep{borgeaud2022improving}{, }FRESHPROMPT~\citep{vu2023freshllms},leaf, text width=12em]
            ]
            [Domain-specific QA,text width=7.5em
                [Reta-llm~\citep{liu2023retallm}{, }LeanContext~\citep{arefeen2023leancontext},leaf]] 
            ]
      ]
    ]
    ]
  \end{forest}
  \caption{Summary of LLM's education-related foundational capabilities. 
  }
  \label{fig:summary}
\end{figure*}
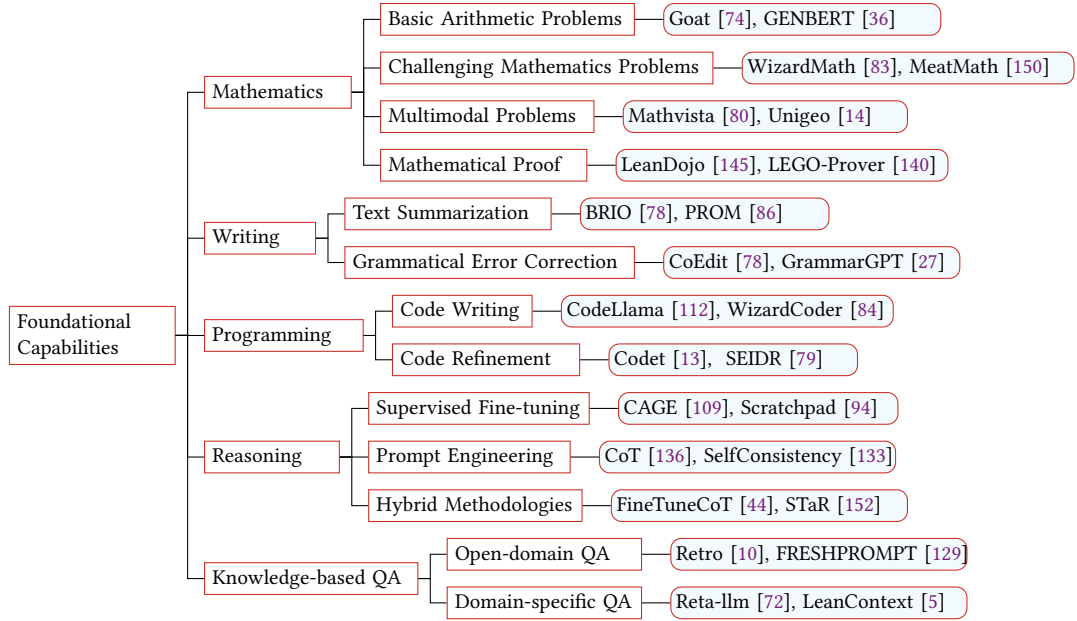

\section{Background}\label{background}
In this section, we first discuss the educational tasks and introduce of LLMs' roles in smart education. Then, we introduce the current development of educational large language models and compare our survey with previous works.

\subsection{Educational Tasks} \label{educationaltasks}
Artificial Intelligence can significantly boost the development of online education. Developing an intelligent education system involves tackling a wide array of tasks, which can be broadly categorized into two types. The first type centers around solving personalized, knowledge-based questions from students\citep{zhou2023solving, yu2023metamath, yasunaga2021qa}. This category aims to assist students in resolving their queries during the learning process, such as clarifying misunderstandings about a particular concept\citep{lazaridou2022internet}, solving a math problem\citep{wang2017deep, gou2023tora, xiong2023trigo}, or writing a piece of code to address a specific issue\citep{chen2022codet, roziere2023code, zhang2023planning}. The second type focuses on aiding students with their learning planning, such as mapping out learning paths\citep{li2023graph, chen2023set}, knowledge tracing\citep{abdelrahman2023knowledge, piech2015deep, corbett1994knowledge}, and computerized adaptive testing\citep{ghosh2021bobcat, meijer1999computerized, thompson2019framework}. These tasks are designed to support the learning process from a broader perspective, instead of addressing specific learning challenges students face.

In this paper, we mainly survey the former category which mainly develops the abilities of LLMs answering students' specific questions requiring specific skills, due to that it's the main scenario that LLMs could contribute more. The former scenario is where LLMs currently play a significant role, for two main reasons: 1) From the perspective of task characteristics, tasks like designing learning paths and tracking knowledge, though also guiding student learning, are mainly based on students' learning sequences. The reasoning process mostly happens in the background, with relatively less need for dialogue. 2) From the viewpoint of the characteristics of LLMs, the advantage of language models over traditional recommendation models lies in their extensive world knowledge, conversational abilities, and logical reasoning capacities. These capabilities are crucial for addressing the personalized and subject-specific questions that students encounter. These questions are often complex and personalized, requiring dialogue for effective understanding—a feature not possessed by previous deep learning models. However, for issues like learning path planning and knowledge tracking, deep learning models, through training, can handle them well~\citep{piech2015deep, li2023graph, ghosh2021bobcat}.

\subsection{Educational Large Language Models}
Currently, many online education companies have launched their own large educational models\citep{xunfeixinghuo, mathgpturl}. The iFLYTEK Spark\citep{xunfeixinghuo}, introduced by iFLYTEK Company, boasts capabilities in multimodal interaction, coding, text generation, solving math problems, and knowledge Q$\&$A.
The introduction of LLMs could evidently enhance the efficiency of learners~\cite{10.1145/3631802.3631806}.
MathGPT\citep{mathgpturl}, developed by TAL Education Group, is an LLM specializing in math problem-solving and lecturing. 
From the development patterns of these companies, it is evident that the primary application scenario for these large models is to address specific knowledge-based questions from students. This approach benefits the industry in two main ways: 1) It allows for more interaction with students. Answering specific questions about subjects or knowledge points involves more detailed participation in the students' learning process compared to learning path planning and knowledge tracking, thereby increasing the time students spend on the platform. 2) Solving student questions makes the model appear more intelligent. Planning learning paths and assessing students can be achieved with traditional models, but understanding and solving students' problems through dialogue is only possible with large models, making the platform's products more intelligent. In summary, because they better retain student users and make their products more intelligent, the main trend for platforms developing educational large models is to enable them to solve specific student questions. Of course, issues like learning path planning and knowledge tracking are also very important, but the transformation in the era of large models requires further exploration.

\subsection{Related Surveys}
LLMs hold vast potential in the field of education. There are already several surveys about LLMs in education, while our work is different from them. 
\citet{gan2023large} explored various roles LLMs can play in the education process. It focuses on analyzing the roles undertaken by LLMs from the perspective of different application scenarios, such as learning support tools, personalized learning experiences, content creation and generation, language learning and teaching, cross-lingual communication, and translation. The basic capabilities of LLMs are not discussed in it. 
~\citet{kasneci2023chatgpt} explores the benefits and challenges of LLMs in education from both students' and teachers' perspectives, highlighting the potential of LLMs in research, writing, and problem-solving tasks, as well as their ability to provide domain-specific language skills for specialized learning. 

Other than that, \citet{al2023chatgpt} primarily explored the use of generative AI models in education, focusing on their application as teaching aids, the generation of personalized learning materials, and the assessment of student learning outcomes. It primarily assesses ChatGPT's performance on tasks such as instructional design from an educational perspective, in a qualitative rather than quantitative manner. ~\citet{meyer2023chatgpt} is an editorial on the opportunities and challenges of LLMs in academia, analyzing the potential impacts and risks of LLMs from the perspectives of academic writing, education, and programming education. They primarily discussed the role of LLMs in education from a pedagogical perspective, whereas our work leans towards analyzing the current capabilities of LLMs in solving subject-specific questions in the educational process from a technological standpoint, offering ideas and insights for creating an LLM capable of addressing problems across various subjects. \citet{wang2024large} summarized the development of LLMs in education from the perspective of data and technology, discussing the current applications of LLMs in tasks such as study assisting, teach assisting, and adaptive learning. It is important to note that while the article mentions some technical methods for developing LLMs' problem-solving abilities in study assisting, the content covered is not systematic. The technical development for solving problems for students didn't take up much space of this article. Moreover, the disciplines explored in the article are not foundational but include advanced subjects such as Medicine and Finance. In contrast, our work primarily focuses on discussing the development of LLMs' fundamental capabilities in assisting students with solving various problems.

While previous surveys have provided ample discussion on the potential applications of LLMs in education, they exhibit two main shortcomings: 1) Their exploration of LLMs' applications in education often spans a wide range of topics, including designing learning paths, assisting teachers, and planning curricula. As discussed in Section ~\ref{educationaltasks}, although these capabilities are important, they do not represent the primary direction of current practical applications of educational LLMs. Our work, in contrast, primarily focuses on the ability of LLMs to answer subject-specific questions. 2) They have not analyzed the development of LLMs' educational capabilities from a technological perspective. Discussing the evolution of LLMs' educational abilities from a technical standpoint is crucial for building a general LLM-based intelligent education system.
Different from them, we review the evolution of LLMs from the perspective of education-related capabilities. We summarize the techniques to promote LLMs in these capabilities.  Additionally, we provide foresight into constructing feasible frameworks for LLM-based education systems. Our work emphasizes a comprehensive understanding of LLMs' educational competencies and explores frameworks that can effectively integrate these abilities into the educational landscape.

\section{Foundational Capabilities} \label{capabilities}

\subsection{Mathematics}
Mathematics demands reasoning of complex information, making it one of the disciplines that place the highest premium on the cognitive abilities. There is significant academic interest in enhancing the mathematical capabilities of LLMs \citep{lu2022survey}. In the pursuit of creating an education system based on LLMs, the goal is to equip it with the capability to tackle a wide range of mathematical problems. These problems may encompass basic numerical calculations, complex logical reasoning, or challenges that require the integration of information from multiple modalities. In this section, we summarize the developments of the mathematical capabilities of LLMs, focusing on four aspects: fundamental numerical computations, complex reasoning, the handling of multi-modal problem-solving, and mathematical proof. 

\subsubsection{Basic Arithmetic Problems}
 Recently, significant scholarly attention has focused on augmenting LLMs' proficiency in this domain \citep{patel2021nlp,wang2017deep,zhang2023evaluating,geva2020injecting}. In human learning mathematics, foundational mathematical operations serve as the basis for addressing more advanced problems. Given the robust comprehension of human language and notable textual reasoning abilities exhibited by LLMs, it is natural for many to assume that LLMs should effortlessly handle fundamental mathematical problems. However, the reality proves otherwise. \citet{yuan2023well} pointed out that ChatGPT and GPT-4~\citep{achiam2023gpt} perform well in addition and subtraction operations, but their accuracy decreases when dealing with multiplication involving larger numerical values. This limitation arises because the LLMs do not access a calculator during the computations. More importantly, when LLMs solve computational problems, their internal logic does not perform actual calculations. Instead, they generate text to predict each digit of the answer step by step. This approach means that as the number of digits in the answer increases, the probability of making an error grows cumulatively.

This problem is not unsolvable. \citet{yang2023gpt} and \citet{liu2023goat} proposed to fine-tune LLMs on high-quality datasets and found that even small language models could avoid making mistakes on multi-digit problems. \citet{lee2023teaching} found that even small transformers could solve arithmetic problems with high accuracy as long as trained on data with detailed calculation process. All in all, fine-tuning on high quality data is one workable solution. However, this approach may only work when a specialized arithmetic model need to be build while not applicable for building general LLMs since it is not feasible to fine-tune each LLM individually. How to prevent such issues during the pre-training phase of LLMs remains an open question. At the current stage, a simple and feasible solution is to have the LLMs outsource arithmetic problems to a calculator, which can ensure the accuracy of the calculations~\citep{schick2024toolformer}.

\subsubsection{Challenging Mathematics Problems}
Despite occasional errors in basic mathematical operations, the expectation for LLMs to solve more complex problems remains high, and the LLMs' capabilities in this area continue to develop. For education, the ability to handle college-level mathematics is particularly beneficial for senior students' learning, offering significant assistance in understanding challenging concepts.
For simple arithmetic problems, LLMs make mistakes mainly due to the gap between text generation and digit calculation. For complex mathematical problems, the challenge arises for requiring LLMs' advanced symbolic reasoning abilities and domain knowledge. In this regard, LLMs still need further development. \citet{wang2023scibench} introduced a benchmark SCIBENCH, which contains collegiate-level scientific problems from mathematics, chemistry, and physics textbooks, while GPT-4 could only get averagely 53.24\% on the math part. Furthermore, ~\citet{sawada2023arb} collected a harder dataset ARB that contains math problems from Harvard PhD comprehensive exams in mathematics, where GPT-4 could only get less than 10\% right. All these results demonstrate that LLMs have a lot of room for improvement.

In recent years, more and more work has been proposed to enhance this ability \citep{luo2023wizardmath, yu2023metamath,lewkowycz2022solving,wang2023scibench}. \citet{luo2023wizardmath} tried to increase the mathematical reasoning abilities of Llama-2 by applying Reinforcement Learning from Evol-Instruct Feedback (RLEIF) on complex mathematics datasets. The Evol-Instruct method made LLMs to generate easier and harder questions from the original questions to make the LLMs think deeper. In addition to training on high-quality datasets, there are also many efforts that attempt to leverage programming as an external tool to assist in solving mathematical problems. 
\citet{zhou2023solving} tried to enhance GPT-4's mathematical ability by encouraging it to use code to self-verify its answers. This approach leads to a significant improvement in the zero-shot accuracy of mathematical problem-solving. 
ToRA~\citep{gou2023tora} divided the process of LLMs solving math problems into a cyclical rationale-action process, where the action involves invoking external tools, including computation libraries and symbolic solvers, thereby amalgamating the analytical prowess of language with the computational efficiency of tools. 
Overall, solving complex mathematical problems requires reasoning abilities, computational power, and knowledge in the mathematical domain. Fine-tuning on relevant datasets primarily aims to enhance its reasoning capabilities and knowledge on related issues, while computational abilities can be supplemented by invoking external tools, such as calculators or code compilers.

\subsubsection{Problems Involving Multi-Modal Information}
Multi-modal inputs are common in mathematics problems like geometric questions. They require LLMs to understand text and image information for solutions. Research in multi-modal LLMs for mathematical reasoning is emerging \citep{lu2023mathvista, chen2022unigeo, peng2023geodrl}. 
This type of task poses a high requirement for the formation and quality of the training data. \citet{chen2022unigeo} introduced a Unified Geometry problem benchmark combining calculation and proving tasks. Based on this dataset, the study presented a framework capable of simultaneously solving calculations and proving tasks through a sequence generation approach. Furthermore, \citet{lu2023mathvista} proposed MATHVISTA, a benchmark for diverse mathematical and visual challenges. \citet{zhang2024geoeval} developed a benchmark GeoEval for testing geometry problem-solving ability of LLMs. Their results showed that WizardMath and GPT-4V excels in handling multi-modal mathematics problems.

Various methods have been proposed for this task. For geometric problems, \citet{zhang2023multi} converted diagrams into text clauses, using a convolutional neural network and a language model for encoding and a GRU-based framework for answer generation. \citet{gao2023g} argued that the reason current models fail on solving geometry problems is that they struggle to accurately comprehending basic geometric elements and their relationships. So they built a augmented dataset Geo170K containing high-quality descriptions of geometric information and developed a model G-LLaVA on it, which demonstrated exceptional performance in solving geometric problems, significantly outperforming GPT-4-V on the MathVista benchmark with only 7B parameters.

Besides geometry problems that involves processing images and text, \citet{lu2022dynamic} presented the Tabular Math Word Problems (TABWP) dataset, requiring textual and tabular data reasoning, and introduced PROMPTPG, a policy gradient-based selector for training and prompt construction for test samples.

\subsubsection{Mathematical Proof}
Unlike other types of mathematical problems where LLMs primarily focus on providing answers, LLMs' role in mathematical proofs emphasizes the integration with proof assistants such as Coq~\citep{barras1997coq}, Isabelle~\citep{nipkow2002isabelle}, and Lean~\citep{de2015lean}. These proof assistants correspond to specific programming languages, requiring users to formulate proofs in the required languages, after which the assistant can verify the proof's correctness. Many LLM-based methods are proposed to help theorem proving \citep{jiang2023multilingual, liu2023fimo, xiong2023trigo}. Based on these proof assistants, there are two main approaches to utilizing LLMs for mathematical proofs. 

The first approach is formal proof search, exemplified by models like GPT-f~\citep{polu2020generative}, which involves prompting LLMs to produce the next proof step (also called 'tactic' in proof assistants) based on the current proof state and some optional context. In conjunction with proof assistants, it transforms the proof of mathematical propositions into a process of executing actions. Here, an action can be the application of a mathematical theorem or a method of variable substitution, which can transform and decompose the original proposition. The LLM is responsible for generating actions, sampling multiple actions in each round and iterating over multiple rounds to produce a tree structure. It utilizes the proof assistant's functionality to verify the validity of proofs to evaluate branches, thereby employing tree search methods to find proof strategies. Following GPT-f, Thor~\citep{jiang2022thor} was further proposed to help select the premise for theorem proving. \citet{yang2023leandojo} introduced an open-source framework named LeanDojo based on the Lean proof assistant. This framework comprises data, toolkits, models, and benchmarks, and it has led to the development of ReProver (Retrieval-Augmented Prover), which enhanced proof accuracy by using retrieval methods to extract premises for LLMs to base their mathematical proofs.

The second is natural proof translation, also known as autoformalization, which is to convert math proofs written in natural language into formalized versions. 
In such schemes, the responsibility of LLMs is not to generate proof steps. Due to the extremely low prevalence of these proof assistants in the human corpus, LLMs struggle to directly undertake the task of generating proof steps. This approach primarily tackles the challenge of insufficient data for formal mathematical proofs. By leveraging autoformalization, a significant increase in this type of data can be achieved, consequently enhancing the proof-generating capabilities of neural provers that have been fine-tuned on this expanded dataset. Initially, ~\citet{wu2022autoformalization} demonstrated that LLMs perform well in autoformalization. They employed LLMs for autoformalization, transforming mathematical proofs and problems expressed in natural language into formal specifications and proofs in Isabelle language. The generated data was used to train a neural theorem prover, enhancing the effectiveness of the original prover. Following it,~\citet{cunningham2023towards} utilized an encoder-decoder framework based on the universal transformer architecture, converting both problem statements and mathematical proofs written in LaTeX into the language of the Coq interactive prover. ~\citet{jiang2022draft} built a pipeline of Draft, Sketch, Prove (DSP), where the informal and incomplete proof is first generated (Draft) and given to LLMs for autoformalization (Sketch), and finally passed to off-the-shelf prover to be completed (Prove). 

In mathematical education, proof problems are indispensable. Currently, LLMs for math proof primarily operate in the form of interactive theorem proving. In this approach, LLMs complete proofs by interacting with software proof assistants. To realize completely automated theorem proving with LLMs, it is essential that these models possess not only strong reasoning skills but also the capability to formalize concepts effectively. There is no room for hallucination in mathematical proofs, which poses a formidable challenge for LLMs.

\subsubsection{Summary}
When examining the progress of large language models (LLMs) in terms of their mathematical abilities, it becomes evident that the primary obstacle lies in the inherent conflict between the principles of mathematical logic and those of text generation. This discrepancy manifests itself not only in the outcomes (e.g., LLMs encountering difficulties with multiplication involving large numbers and struggling with complex mathematical problems) but also in the training data itself. Mathematical problems, in their symbolic form, represent only a minor fraction of the training corpora used for these expansive models. Consequently, the current approaches to bolstering the mathematical capabilities of LLMs can be broadly categorized into two main strategies: 1) Data enhancement: The most straightforward method to improve LLMs' performance on mathematical tasks is to provide them with high-quality, relevant data during the fine-tuning phase of their training. By exposing the models to a more comprehensive and representative set of mathematical problems, their ability to handle such challenges can be significantly enhanced. 2) Tool integration: Another effective approach is to leverage external tools, such as calculators and code compilers, to compensate for the inherent limitations of LLMs. By strategically invoking these tools at the points where the models struggle, their functional shortcomings can be effectively mitigated, allowing for a more comprehensive and accurate handling of mathematical problems.

\subsection{Writing}
Writing proficiency is crucial for LLMs, underpinning their ability to comprehend inputs deeply and produce semantically and syntactically accurate outputs \citep{surveyDong2023,chang2023survey}. In education, the writing capability of LLMs holds the potential to transform how writing is taught. They can assist in content creation, simplify complex topics for students, and offer personalized educational materials. In this part, we dive into LLM's writing capability on two education-related tasks: text summarization and grammatical error correction.

\subsubsection{Text Summarization}
Text summarization is a task that requires LLMs to compress lengthy texts into concise summaries while maintaining the essential information. This process presents a significant challenge for LLMs, as they must effectively comprehend and distill the key points from a wide range of diverse content, such as news articles and texts written in multiple languages. In the context of education, students are often confronted with an overwhelming volume of intricate learning materials. A well-crafted summary can prove invaluable in helping them grasp the core concepts quickly and efficiently, saving them considerable time and effort. For instance, a summary can break down a complex piece of code into its fundamental components, making it easier for students to understand its structure and functionality. Similarly, a summary can highlight the main ideas and key takeaways from a lengthy chapter, allowing students to focus on the most critical information without getting lost in the details. It is evident that traditional fine-tuning methods are less effective with the advent of advanced LLMs \citep{pu2023summarization,liu2022brio}. \citet{pu2023summarization} and \citet{liang2022holistic} showed LLMs like ChatGPT initially lag behind fine-tuned models like T5 \citep{raffel2020exploring} and BART \citep{lewis2019bart} in ROUGE scores \citep{lin2004rouge} for text summarization. However, when human judges evaluate overall quality, LLMs outperform fine-tuned models and even standard human summaries, superior in aspects like factual consistency, fluency, and diversity. This discovery underscored the limitations of traditional evaluation methods and suggested a need for new paradigms to guide summarization tasks in the LLM era. For example, BRIO \citep{liu2022brio} implemented a ranking task to foster more diverse summarizations. Furthermore, \citet{liu2023learning} utilized a GPT model based on BRIO to directly generate training data to guide the learning process of other models, which is similar to the process of RLHF \citep{stiennon2020learning}.

Given the outstanding performance of LLMs in the domain of text summarization, researchers have already begun to tackle more challenging tasks. \citet{liu2023benchmarking} benchmarked LLMs on instruction controllable text summarization. In this task, the input provided to the model consists of two components: the source article that needs to be summarized and a set of instructions in natural language that specify the desired characteristics of the summary output. The goal is to evaluate how well LLMs can generate summaries that adhere to these specific requirements. In a related study, ~\citet{shen2023large} investigated whether LLMs could potentially replace human evaluators in assessing the quality of abstractive summarization. Abstractive summarization involves generating a summary that captures the main ideas of the source text while potentially using different words and phrases. The researchers found that, at present, LLMs are not capable of serving as reliable substitutes for human evaluators in this task. LLM evaluators rate each candidate system inconsistently and are dimension-dependent. Moreover, LLMs face challenges when comparing candidate summaries that have similar levels of performance. They find it difficult to make fine-grained distinctions between summaries of comparable quality, which limits their ability to provide accurate comparative assessments. The correlation between the ratings provided by LLMs and those given by human evaluators becomes lower when dealing with higher-quality summaries. Although LLMs can surpass humans in text summarization tasks, they are not without flaws. Current LLMs make fewer silly mistakes (e.g., entity confusion, irrelevant information generation) but more sophisticated ones~\citep{pu2023summarization}. For example, they fill in the details related to but not directly supported by the source text, which is a kind of ``hallucination''. ~\citet{liu2022improving} tried to employ human feedback to enhance the summarization factual consistency. The dataset DeFacto they built contained human demonstrations and informational natural language feedback consisting of
corrective instructions, edited summaries, and explanations with respect to the factual consistency of the summary.
~\citet{feng2023improving} tried to resolve this hallucination problem by disentangling the
comprehension and embellishment abilities of LLMs. It trained the embellishment to be consistent with the facts presented in the original text.  

Overall, LLMs perform well in text summarization tasks, even surpassing humans in simple summaries, but this does not mean they are flawless. In the educational domain, helping students summarize learning materials should ensure there is no conflict between the summary and the original content. LLMs still face issues with hallucinations in this aspect. While these hallucination issues can be mitigated through post-processing techniques, hallucinations remain a fundamental problem with LLMs that extends beyond the task of text summarization. Addressing the issue of hallucinations in LLMs is an ongoing research challenge that requires further investigation and development of novel approaches. Until a satisfactory solution is found, it is important to exercise caution when using LLM-generated summaries in educational contexts and to have mechanisms in place to verify the accuracy and consistency of the summaries with the original learning materials.

\subsubsection{Grammatical Error Correction} 
We are well aware of the remarkable capability of LLMs to generate fluent and coherent conversations. However, from an educational perspective, the importance of producing grammatically correct dialogues cannot be overstated, especially for students learning a new language. The correctness of grammar in conversations plays a vital role in language acquisition, providing students with reliable examples to emulate and learn from. Numerous studies have evaluated the effectiveness of LLMs in grammatical error correction (GEC). Several works \citep{wu2023chatgpt,fang2023chatgpt,mohammed2023chatgpt} first evaluated the error correction performance of closed-source LLMs such as ChatGPT. Although there exists a pronounced gap between ChatGPT and the previous state-of-the-art \citep{omelianchuk2020gector,Grammarly} models on the overall F0.5 metric, closer analysis shows that ChatGPT underperforms other models in terms of precision but far exceeds other models in terms of recall. That said, LLMs like ChatGPT are good at error detection. A detailed manual analysis of ChatGPT's outputs revealed that, in most cases, it maintained grammar accuracy better than the previous methods. However, it often overcorrects sentences to increase diversity and fluency, resulting in a decrease in the recall score. As a result of this characteristic, LLMs perform better when evaluated on higher-order metrics such as fluency, which assess modifications to text. However, for issues requiring minimal edit corrections, they may not necessarily outperform traditional models. Some efforts have attempted to mitigate the problem of over-correction in large models by employing instruction-tuning techniques, encouraging them to make only the necessary, such as CoEdit \citep{raheja2023coedit}, which covered multiple text editing tasks (including GEC) by fine-tuning LLMs to integrate the capabilities brought by these tasks. GrammarGPT \citep{fan2023grammargpt}  collected grammatically incorrect sentences and performed instruction tuning on LLMs to improve the ability of positioning grammar errors. 

Overall, LLMs perform well in the area of GEC, with their main issue being over-correction. In scenarios such as copywriting or article writing, this problem is not critical, as LLMs can assist individuals in correcting grammatical errors while crafting more fluent sentences. However, in educational settings, LLMs' GEC capabilities are more often utilized to aid students in learning grammar. This requires LLMs to accurately identify grammatical errors in sentences. The issue of over-correction could potentially mislead students, making further adjustments necessary.

\subsubsection{Summary}
Leveraging LLMs' proficiency in text summarization and grammatical error correction can significantly benefit education. Their capability to condense complex material into concise summaries facilitates efficient learning, while error correction tools help improve students' writing and language skills. However, critical challenges need to be resolved to integrate these writing-related capabilities to help education. It becomes evident that more refined evaluation metrics and task-specific optimizations are essential for LLMs.

\subsection{Programming}
Programming is a process of writing code and correcting code if unexpected results are obtained. Incorporating LLMs in programming education is reshaping the future of AI-assisted programming learning. LLMs could play multiple roles: as instructors providing guidance, as teaching assistants offering personalized tutorials, and as collaborative coding partners. Studies like \citep{ma2023hypocompass} demonstrated improved performance (17\%) and efficiency (13\%) among programming novices using LLM-based assistants. 
Research from \citep{phung2023generative} focused on programming education tasks and benchmarks like \citep{fu2023codeapex} and \citep{ding2023crosscodeeval}, which were used to evaluate the effectiveness of LLMs. This section mainly discusses the LLMs' coding capability development from two perspectives: code writing and code refinement, corresponding to the two stages in programming.


\subsubsection{Code Writing}
Unlike natural language tasks, generating code requires a more rigorous token syntax and places higher demands on the training stage.
A common method to improve LLMs' performance in generating code is to train or fine-tune them on extensive code datasets \citep{codex,nijkamp2023codegen}.
WizardCoder \citep{luo2023wizardcoder} introduced the Evol-Instruct \citep{xu2023wizardlm} method to generate complex and diverse instruction datasets of code-related tasks.
To emulate the iterative process of humans repeatedly modifying and reviewing code, InCoder \citep{fried2023incoder} utilized bidirectional encoding instead of left-to-right encoding. 
In addition to next-token prediction, training or fine-tuning code-aimed LLMs on additional code-related tasks could enhance their programming capabilities.
LLMs first learn language patterns and representations from a large amount of text data through unsupervised learning. 
Then, they could be fine-tuned on labeled code tasks, allowing them to learn targeted code representations and gain a deep understanding of code structure and semantics based on the provided labels.
CodeT5+ \citep{wang2023codet5} introduced the concepts of unimodal and bimodal alignment, increasing the model's adaptability to function in different modes for various downstream tasks. During the bimodal alignment phase, the model synchronizes the representations of text-code pairs using multiple tasks, which improves its ability to understand and generate content across different modalities.
CodeLlama \citep{rozière2023codellama} also applied the multi-task objectives, including autoregression and causal infilling prediction, which achieves better performance among open models.
MFTCoder \citep{liu2023mftcoder} utilized the Multi-Task Learning (MTL) technique and incorporated a training loss computation algorithm to alleviate the instability and imbalance of multi-task training.
Given that code text has its unique syntax and structure compared to natural language text, the methods mentioned above all attempt to construct datasets of code and perform fine-tuning. This is the most direct and effective approach to enhancing the code capabilities of LLMs.

By fine-tuning on code datasets, LLMs can enhance the probability of generating a correct piece of code. Considering the human programming process, besides relying on the programmer's coding skills, it also involves various stages such as consulting documentation, designing code frameworks, implementing and testing submodules, which entail numerous decision-making processes. Therefore, many works perceive LLMs as agents, viewing the coding process as a continuous series of decisions and external tool invocations. 
~\citet{zhang2023planning} attempts to enhance the effectiveness of model code generation by employing a tree search approach, without altering the parameters of LLM itself. Specifically, each token in the code is regarded as an action, with the generated code serving as the state. The LLM makes decisions step by step, while utilizing Monte Carlo Tree Search (MCTS) to calculate the value of each action (token) in the current state, thereby selecting the optimal action and significantly improving its pass rate in code generation. Similarly, ~\citet{zhou2023language} also treats LLM as an agent, where each action involves generating a complete piece of code. It employs MCTS to estimate the value of each action as well. ~\citet{shinn2024reflexion} introduced Reflexion framework, which enhances LLM agents through linguistic feedback. The method assigned LLMs the roles of both generating code as actors and evaluating it the code as evaluators. Additionally, it utilizes self-reflection to generate verbal reinforcement cues aimed at assisting the actor in self-improvement. ~\cite{zhou2022docprompting} introduces a document retriever as a precursor to the code generator, extracting relevant function descriptions from documents to provide external information to the LLM. This enables the generated code to utilize the latest library functions. By considering the LLM as an agent and utilizing external documents or tree search algorithms, the accuracy of the agent's decisions can be improved without the need to update the model's parameters, which reduces training costs. However, this approach also has a downside: it increases the time required for decision-making during the code generation process, resulting in lower inference efficiency compared to using the LLM alone.

In addition to single-agent approaches, multi-agent systems have also made significant progress in code generation tasks. ~\citet{qian2023communicative} developed ChatDev that divided the process of writing code into four stages: designing, coding, testing, and documenting. Each stage is managed by a group of ``software agents'', with the entire chat chain acting as a facilitator, assigning specific sub-tasks to each stage.  The system achieved efficient code writing. Furthermore,  ~\citet{hong2023metagpt} proposed MetaGPT, a Multi-agent system for coding, which assigns diverse roles to different agents, such as Product Managers, Architects, Engineers, etc. They introduce Standardized Operating Procedures (SOPs) into prompt sequences, effectively enhancing the effectiveness of code generation. 
Although chat-based multi-agent systems have shown significant effectiveness in code generation tasks, they impose high demands on the fundamental capabilities of LLMs due to the need for dialogue-based coordination between agents. The effectiveness of these systems generally increases with the rise in the parameter size and capability of LLMs.

\subsubsection{Code Refinement}
In most cases, LLMs could not generate the correct code at once.
We could enable LLMs to generate a code sketch (either actual code or pseudocode) and utilize various methods to guide the model to modify and refine the code.
By leveraging the inherent code correction ability of LLMs, the overall precision and quality of the code could be significantly enhanced. 

We investigate the development of LLMs in code refinement from two perspectives. One aspect is the advancement of LLMs in the task of fixing code bugs, known as Automated Program Repair (APR). \citet{sobania2023analysis} conducted experiments analyzing the performance of ChatGPT on the APR task. They found that it can achieve competitive results compared to previous deep learning-based methods, and incorporating additional information through dialogue can surpass previous approaches. \citet{xia2023conversational} introduced conversational APR, enabling LLMs to obtain bug feedback through dialogue, effectively enhancing the performance of various LLMs in APR. In addition to the APR task itself, the second aspect involves integrating code refinement into the process of code generation, leveraging bug feedback to enhance the effectiveness of code generation. \citet{fullyautonomousp} constructed a pipeline: Synthesize, Execute, Instruct, Debug, and Rank (SEIDR). 
It first generates multiple different codes and undergoes the process of code filtering and debugging, ultimately selecting the best code among them.
According to \citet{magister2022teaching},  teaching an LLM to debug its program draft via few-shot demonstrations could improve the performance on code generation tasks.
Another method for LLM debug is generating unit tests by LLM itself and checking its code \citep{chen2022codet}.
By mimicking the human coding process, LLM's programming ability is greatly enhanced. 
However, these methods lead to an increased number of calls to LLMs, resulting in a significant increase in inference time.

In the context of coding education, the support and guidance provided by LLMs do not yet match the level of assistance offered by human instructors. One major reason for this gap is that LLMs still have substantial room for improvement in their coding capabilities.
While LLMs can generate functional code for relatively simple tasks, when it comes to generating complex algorithms, LLMs' performance rapidly declines compared to the functionality achieved by humans \citep{chen2021evaluating}. 
Additionally, due to the lack of real-world data, LLMs struggle to learn the intermediate thinking process of code writing, making it difficult for them to provide relevant explanations and instructions to beginners. 
As a result, the use of LLMs in programming education still needs to be improved, especially in terms of interpretability.

\subsubsection{Summary}
Code data is more abundant in the training corpora of LLMs compared to mathematical data. This is primarily due to the inherent nature of coding, which heavily relies on computers and the internet. As LLMs are trained using data scraped from the web, they are exposed to a significant amount of code-related information during the training process. Many improvements to LLMs' code generation abilities are inspired by the human programming process. For instance, programmers often refer to documentation and resources to gain a better understanding of the problem and potential solutions. For complex coding tasks, the solving process typically involves a cycle of design, writing, and debugging. These thought processes can be utilized to enhance the programming effectiveness of LLMs. From an educational perspective, it is crucial for LLMs to not only generate correct code but also possess the ability to analyze and provide feedback on code written by students. This involves identifying issues, suggesting improvements, and offering explanations to help students learn and grow as programmers.

\subsection{Reasoning}
 
The reasoning capability of LLMs offers significant potential for educational use, serving as advanced tools that enhance students' cognitive processes, provide personalized mentorship, and offer tailored learning support. 
This section reviews LLMs' general reasoning ability development strategies.

\subsubsection{Supervised Fine-tuning for Reasoning}
Previous studies have primarily focused on fully supervised fine-tuning LLMs to enhance their reasoning capabilities. This approach aligns model outputs closely with labeled datasets, allowing the models to produce highly accurate predictions within specific domains. One such study by \citep{rajani2019explain} demonstrated the efficacy of fine-tuning a pre-trained GPT model, which generated rationales for predictions on the CoS-E dataset \citep{talmor2018commonsenseqa}. The results revealed that models trained with explanations exhibited improved performance in commonsense question-answering tasks. However, the effectiveness of fine-tuning methods heavily relies on the availability of a specific dataset that includes explicit reasoning steps. Acquiring such a dataset can prove to be challenging. Moreover, the scope of inference from fine-tuned models is restricted to the dataset's domain, hinging largely on the data's inferential quality. This constraint highlights the benefits and limitations of fully supervised fine-tuning, as it narrows the model's reasoning abilities to the dataset's specific domain. Consequently, it underscores the need to explore methods that harness LLMs' intrinsic reasoning capabilities, potentially providing broader relevance and deeper insights beyond the limitations of domain-specific datasets.

\subsubsection{Prompt Engineering for Reasoning}
Efforts have been made in recent research to tackle the constraints inherent in the fine-tuning process of LLMs. These fine-tuning methods tend to overfit specific dataset distributions, reducing their effectiveness on more diverse datasets. In response to this issue, a variety of strategies have been proposed. These strategies are designed to draw upon the robust reasoning abilities inherent in LLMs by leveraging their extensively pre-trained parameters. One approach involves guiding LLMs to generate inference and reasoning through demonstrations or prompts. For example, \citet{wei2022chain} introduced the ``Chain of Thought'' (CoT) method, which utilized natural language reasoning steps as prompts for the model. By integrating CoT within a few-shot prompting framework, the model leveraged its extensive parameters to produce analogous chains of reasoning. Consequently, this approach empowered the model to adeptly navigate complex reasoning tasks across diverse domains, obviating additional training or fine-tuning. This innovation underscored the model's inherent capability to generate deductive pathways, significantly enhancing its applicability and versatility in problem-solving scenarios without extensive domain-specific adaptations. Similarly, \citet{wang2022self} introduced a self-consistency strategy that enhances model performance by sampling various reasoning paths and selecting the most consistent answer. This approach diversified the exploration of reasoning strategies. It ensured the conclusions' reliability, showcasing an innovative way to leverage the model's capabilities for improved decision-making and problem-solving across different contexts. Confronting the limitations of relying on static, manually annotated demonstrations, which can restrict the adaptability of LLMs to the varying complexities of real-world tasks, \citet{diao2023active} introduced an active selection approach. This technique dynamically pinpointed the most pertinent demonstrations aligned with the specific demands of a task from a broad set of queries. In this way, the approach enhanced the flexibility and effectiveness of LLMs in adapting to diverse and evolving problem contexts. Concurrently, \citet{zhou2022least} devised a prompting methodology that broke down intricate problems into their simpler constituent sub-problems. This tactic not only promoted a step-by-step problem-solving process but also hold promise for augmenting the efficacy of LLMs in handling complex tasks. 

Building upon the CoT methodology, subsequent developments have introduced more sophisticated frameworks for enhancing the reasoning capabilities of LLMs. The Tree of Thoughts (ToT) \citep{yao2024tree} framework extended CoT by enabling LLMs to explore multiple reasoning paths through a hierarchical structure, thereby improving decision-making for tasks requiring strategic planning. Following the ToT, the Boosting of Thoughts (BoT) \citep{chen2024boosting} framework introduced a novel approach by iteratively exploring and self-evaluating multiple trees of thoughts. This process accumulated an ensemble of trial-and-error reasoning experiences, offering a new form of prompting designed to tackle complex problems. Starting with simple prompts, BOT iteratively refined reasoning steps through error analysis, significantly improving the generation of reasoning paths and achieving higher problem-solving rates on complex tasks than existing advanced prompting strategies. To structure thoughts through prompts without depending on fine-tuning, the Graph of Thoughts (GoT) framework \citep{besta2023graph} introduced a new angle by arranging thoughts generated by LLMs into a graph structure. This setup fostered a dynamic interaction among different thought units, facilitating synthesizing synergistic results, simplifying intricate thought networks, and refining ideas via feedback mechanisms. GoT's graph-based approach presented a versatile tool for problem-solving, allowing for a more nuanced and interconnected reasoning process that mirrors the complexity of human thought.

The emergence of CoT and related prompting strategies marked a significant evolution in utilizing LLMs for advanced reasoning, transitioning away from dependence on fine-tuning. These methods exploited the LLMs' inherent capabilities to enhance their flexibility and effectiveness across various tasks without reliance on domain-specific tuning.

\subsubsection{Hybrid Methodologies for Reasoning}

Despite the success of prompt engineering in leveraging the intrinsic characteristics and capabilities of LLMs to enhance their performance, this approach falls short of fundamentally augmenting the model's core reasoning abilities, as it does not alter the model's underlying parameters. This inherent limitation points to a need for strategies that not only exploit the pre-existing strengths of LLMs but also seek to expand their innate capabilities. Innovative approaches that integrate the specificity of fine-tuning with the flexibility of prompt engineering have been developed to bridge this gap. These hybrid methodologies aim to bolster the LLMs' responsiveness to complex prompts and substantially improve their intrinsic reasoning capacities, offering a more comprehensive enhancement of their problem-solving prowess. One practical approach is to utilize LLMs to ``teach'' language models with smaller model sizes. \citet{ho2022large, magister2022teaching} explored to fine-tune a student model on the chain of thought outputs generated by a larger teacher model and proved that enriching the fine-tuning data with such diverse reasoning results in a substantial performance boost across datasets even for very small models. Moreover, \citet{zelikman2022star} reported significant performance improvements across multiple datasets by generating step-by-step rationales and fine-tuning models based on correct answers, thus facilitating model learning from its reasoning. Similarly, \citet{huang2022large} proposed that by employing chain-of-thought prompting \citep{wei2022chain} and self-consistency \citep{wang2022self} to generate rationale-augmented answers, and then used these answers for fine-tuning, LLMs autonomously refined their reasoning capabilities. This approach highlighted the significant ability of LLMs to advance their knowledge and problem-solving skills independently.

\subsubsection{Summary}

In the educational domain, reasoning tasks possess unique characteristics that necessitate not only the accurate processing of information but also the capability to navigate and elucidate complex concepts in a manner that is accessible and educational for learners. As mentioned above, the discussed methods have significantly advanced the reasoning capability of LLMs, optimally utilizing their unique features for diverse reasoning tasks. This enhancement can greatly benefit educational applications. However, it's crucial to recognize the limitations. As underscored by \citet{valmeekam2022large} and \citet{ruis2022large}, LLMs struggle with complex reasoning tasks and those requiring implicit expressions. For example, LLMs can struggle with complex reasoning scenarios, leading to a notable decrease in performance. This is particularly relevant in educational contexts, where incorrect problem-solving modeled by LLMs could misguide students and lead to misunderstandings or flawed comprehension. Thus, despite LLMs' immense educational potential, their limitations must be carefully considered to ensure they facilitate rather than obstruct learning.

\subsection{Knowledge-based Question Answering}

In the context of Knowledge-based Question Answering using LLM, the user presents a question to LLM, and LLM leverages knowledge-based methods and responds with the corresponding answer. 
Previous work by \cite{ren2023investigating} showed that LLMs have an inaccurate perception of factual boundaries and often exhibit overconfidence. Many studies have explored and utilized external knowledge from open-world and domain-specific databases to enhance the knowledge base of these LLMs.

\subsubsection{Open-domain QA}
Open-domain question answering requires LLMs to accurately determine the reliability of information in the open world and craft their responses based on that understanding. 
The critical requirements for open-domain QA are real-time responsiveness and authenticity. LLMs exhibit disadvantages in both of these aspects. Due to the fixed parameters of the model, ensuring real-time information solely through the LLM itself is challenging, and LLMs commonly suffer from severe hallucination issues \citep{xu2024hallucination}, posing a challenge to their authenticity as well.
\citet{jiang2021can} evaluated the accuracy of LLM responses to a particular question from the perspective of calibration. Through experiments, the researchers discovered that models such as T5, BERT, and GPT-2 are not well-calibrated in QA tasks. While suggesting that incorporating calibration-related methods into the fine-tuning process can effectively enhance performance in QA tasks, it is evident that solely pre-trained language models still face significant challenges in open-domain tasks. To overcome this challenge, many works tried to add additional information to help the LLMs answer correctly~\citep{khandelwal2019generalization, guu2020retrieval, borgeaud2022improving}. A common information source used is from the web. \citet{lazaridou2022internet} employed information gathered from web searches as prompt input for LLMs, conditioning it to generate answers to questions. This approach effectively enables LLMs to use open-world information to answer questions.  ~\citet{vu2023freshllms} introduced FreshPrompt, which incorporated web pages collected from the internet into prompts given to large-scale models. This allowed them to leverage the latest information when answering questions. ~\citet{kasai2024realtime} developed a QA platform REALTIME QA that updated itself weekly. Through evaluation on this platform, they found that GPT-3 could update its generation results based on newly retrieved documents. However, when retrieved documents fail to provide sufficient information to find the answer, GPT-3 may provide outdated answers. 

The development of LLM-based open-domain question answering highlights significant challenges, particularly in dealing with hallucinations. In the context of establishing an LLM-based education system, this issue becomes more critical for providing seemingly correct yet incorrect answers, leading to misleading students. Drawing insights from approaches that introduce additional information from sources such as the web or textbooks can offer valuable lessons for the development of an LLM-based education system.

\subsubsection{Domain-specific QA}
Although LLMs are trained on vast corpora, they may still exhibit gaps in understanding specific domains. 
The primary challenge faced by LLMs in this task is the lack of domain knowledge. For specific domain questions, providing good answers often requires a considerable amount of expertise or skills in that field, while professional data is relatively scarce in the corpus of large-scale models. One straightforward approach is to fine-tune LLMs on specialized datasets.
Typically, there are dedicated knowledge repositories for professional content that consolidate domain-specific knowledge, such as MedlinePlus\footnote{https://medlineplus.gov}, GeeksforGeeks\footnote{https://www.geeksforgeeks.org}, etc. 
\citet{choi2023conversational} utilized an external knowledge base to generate a set of question-answer pairs and then employed fine-tuning to transfer financial knowledge to LLMs, significantly improving financial question-answering tasks. 
Another common approach is to leverage the in-context learning capability of LLMs by incorporating retrieved knowledge from the knowledge base into prompts. 
\citet{peng2023embeddingbased} demonstrate this approach in their work on pest identification. They first use text embeddings, which are dense vector representations of text, to retrieve relevant information from a knowledge base. Text embeddings allow for efficient and accurate retrieval of similar or related content based on the semantic similarity between the query and the stored information. Once the relevant knowledge is retrieved, it is incorporated into the prompts provided to the LLMs. The LLMs then utilize their automatic feature extraction capabilities to process and understand the retrieved information in the context of the pest identification task.
\citet{zhang2023} utilized K-nearest neighbors (KNN)~\citep{guo2003knn} to search for the most similar K records from an accounting database, serving as k-shot examples, and greatly improved accounting efficiency. 
There are also works that train and improve the retriever encoder \citep{zhang2023retrieve}, as well as distill and refine the data in the database \citep{jeronymo2023inparsv2}. 
Such retrieval frameworks have lower costs and can be more flexible in applications across different domains. 
\citet{liu2023retallm} proposed the RETA-LLM, a system that leveraged an information retrieval system based on Google Search to initially retrieve the top-k documents relevant to a user's query, allowing LLMs to generate answers based on these retrieved documents. Furthermore, the system included plug-and-play modules that enable users to construct their own domain-specific LLMs. These modules covered various functionalities, including request rewriting, document retrieval, passage extraction, answer generation, and fact-checking.

By integrating Information Retrieval (IR) systems, LLMs can enhance their capabilities with professional knowledge, gaining valuable and precise supplemental information.
Furthermore, according to \citet{ren2023investigating}, retrieval augmentation can also be employed to improve LLMs' ability to perceive facts within the boundaries of their legal knowledge, mitigating the issue of hallucinations.
During the education process, different majors or courses involve different professional content.
Applying external knowledge repositories as an enhancement mechanism can provide more accurate guidance in domain-specific contexts and mitigate the issues caused by misleading information. 
Therefore, domain-specific question-answering ability is crucial for developing an LLM-based education system.

\subsubsection{Summary}
While LLMs have mastered a broad range of open-world knowledge through extensive corpus training, their fixed parameters make it challenging to handle real-time, high-demand open-domain questions. Severe hallucination issues further compromise their accuracy in both open-domain and domain-specific queries. In the field of education, where authenticity is paramount, students may pose questions about textbook knowledge points. If the accuracy of these responses cannot be guaranteed, it could potentially mislead students. Therefore, a feasible solution to the inherent hallucination issues in foundational LLMs is to integrate external information, such as authoritative documents, allowing LLMs to base their responses on such external sources to mitigate hallucination problems.

\begin{figure*}[htbp]
    \centering
    \includegraphics[width=0.98\textwidth]{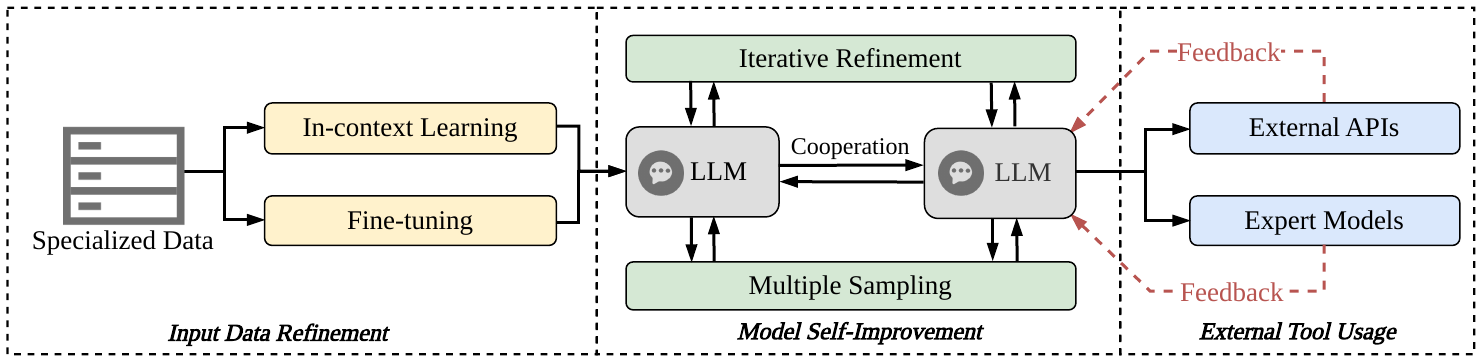}
    \caption{A summary framework diagram for the approaches of LLMs in the development of education-related abilities. It categorizes previous enhancement strategies into three parts: Input Data Refinement, Model Self-Improvement, and External Tool Usage.}
    \label{fig:comsfig}
\end{figure*}

\subsection{Discussion}
Despite the varied contexts and specific challenges associated with each capability, certain strategies and insights resonate universally among researchers working to harness the capabilities of LLMs for educational purposes. Here we discuss the trends or commonalities we discovered across the development of capabilities.

\subsubsection{High-quality data could help LLMs develop capabilities effectively.}
Since the advent of the deep learning era, high-quality training data has significantly improved model performance. This is especially true for LLMs, and the approach extends beyond mere training. Through in-context learning, where desired inputs and outputs are formed into demonstrations and provided as prompts to LLMs, if the demonstrations are well-chosen, they can also greatly enhance the model's capabilities.

There is a notable trend towards leveraging smaller models fine-tuned on high-quality data to surpass the performance of larger language models. This approach emphasizes the importance of focused, domain-specific training rather than simply relying on the vast number of parameters in large-scale models.
The term ``high-quality data'' here refers to data that provides detailed supervisory signals for specific domain problems. For instance, even powerful LLMs like GPT-4 encounter high error rates with large number multiplication in basic arithmetic problems. However, with a dataset that includes detailed steps for multiplication and addition, even a small transformer model can effectively solve these problems. Similarly, for reasoning tasks, it's possible to improve LLMs' performance on these issues without altering model parameters, simply by incorporating specific reasoning steps into the demonstrations in the prompt. This highlights that the application of ``high-quality data'' in LLMs extends beyond just fine-tuning.

There is a concerted effort among researchers to explore how smaller models can achieve equivalent or superior outcomes in certain scenarios where LLMs have already exceeded human benchmarks. This pursuit reflects a broader shift towards optimizing computational efficiency and model scalability, ensuring that the advancements in LLM technology remain accessible and sustainable. The primary scenarios and objectives for training small models can be divided into two categories:
\begin{itemize}
    \item Training specialized models with high-quality data. For applications within specific, narrow fields, such as programming, developing a small, specialized model through data collection can facilitate deployment and reduce the computational resources needed for inference, among other benefits.
    \item Training small models through knowledge distillation. In cases where the required knowledge and skills for an application are more general, it may be challenging to construct a specific dataset for training a small model. By learning to match the outputs of the teacher model, the student model can effectively absorb the knowledge and skills of the larger model, without the need for a specific, curated dataset. This allows the smaller model to inherit the generalization capabilities of the large teacher model, while being more computationally efficient and easier to deploy in resource-constrained environments.
\end{itemize}
Efficiency is an unavoidable issue for LLMs during training and real-world deployment. Training-wise, low-loss, high-efficiency training schemes like LoRA (Low-Rank Adaptation)~\citep{hu2021lora} are continually being introduced. These can significantly reduce the number of trainable parameters required during fine-tuning. However, small models are still needed. In practical applications, where the trained models need to be deployed and used for generating predictions or outputs, the size of the model still plays a critical role.

\subsubsection{LLMs could achieve self-improvement.}

The inference abilities and text comprehension skills of LLMs enable them to conveniently obtain feedback, which allows them to refine their outputs. This process is known as the self-improvement, which is a general approach in improving LLMs' answers to all the capabilities.  LLMs can achieve self-improvement through a methodical approach that involves iterative refinement and multiple sampling. For instance, to provide better responses to queries, an LLM may first generate an initial output. Then, it evaluates the output's effectiveness or accuracy. Leveraging multiple sampling, the LLM explores different solution pathways or creative responses, which expands its potential answers pool. Through iterative refinement, it compares, contrasts, and consolidates these possibilities, learning which strategies yield the best results. This could involve internal processes such as adjusting parameters based on feedback loops, where it might integrate data from new examples or corrections provided by human users. Over time, this enhances the LLM's ability to provide more precise, informative, and contextually relevant answers, thus gradually improving its problem-solving and content creation skills. 
In addition to individual LLMs enhancing their capabilities through feedback, the method of coordinating multiple LLMs to improve overall output is also being explored across various fields. For example, in solving programming problems, the process can be segmented into different stages, with each stage managed by a distinct LLM. These LLMs communicate and collaborate to complete the task collectively. Alternatively, one LLM may act as a generator to produce answers, while another serves as an evaluator to provide feedback. Through continuous dialogue between the two, the response can be consistently assessed and refined, thereby improving the quality of the answer. This cooperative approach leverages the strengths of different models to achieve a more effective and sophisticated problem-solving mechanism.
Although the method of self-improvement is effective, it often results in a longer time to produce responses. For an LLM-based education system, the importance of providing accurate answers to students’ questions (to avoid misleading them) outweighs the need for speed. Therefore, employing a multi-agent approach to enhance answer quality through LLMs' collaboration, or using sampling and iterative optimization for self-improvement, is an appropriate strategy for developing an education system. This ensures that the system prioritizes the correctness and reliability of information, which is crucial in educational settings where students' learning effect is depend on the accuracy of the content provided.

\subsubsection{Calling external tools is an universal method.}
The integration of external tools into the LLM framework is a widely adopted method. This strategy not only enhances the models' ability to access and incorporate real-time information and authoritative sources but also mitigates some of the inherent limitations of LLMs, such as their tendency towards factual inaccuracies or hallucinations.  
We can divide the use of external tools by LLMs into two perspectives:
\begin{itemize}
    \item LLMs inherently have certain limitations that cannot be resolved through training alone, and external tools can be used to address these deficiencies. In this scenario, tools serve the LLMs. For example, LLMs' high error rates in large number multiplication can easily be mitigated by employing an external calculator. Similarly, LLMs' inability to access real-time information can be compensated for by retrieving the latest web pages via web API calls. 
    \item Utilizing the reasoning and decision-making capabilities of LLMs, the invocation of external tools can influence the real world. In this approach, the primary task of LLMs is to make informed decisions about when and which tools to utilize in order to accomplish specific tasks. The key responsibility of LLMs in this approach is to act as intelligent agents that can analyze a given situation, understand the requirements and constraints, and determine the most appropriate course of action.
\end{itemize}
The LLM that solves problems by invoking external tools is a type of LLM agent, where the external tool is not necessarily an API but can also be an expert model. Fine-tuning LLMs on specific datasets can yield excellent results on the corresponding task, but it's impractical to fine-tune LLMs across datasets for all capabilities. A viable solution is to use fine-tuned small language models as expert models, which serve as external tools for the central LLM. Compared to APIs, the advantage of trained expert language models is their ability to understand more granular and flexible demands from the LLM, providing targeted feedback.

\begin{table*}
\centering
  \caption{Overview of LLMs' performance on foundational education-related capabilities.}
  \label{tab-OverallStatus}
  \resizebox{\textwidth}{!}{
  \begin{tabular}{c|c|cccccc}
    \toprule
     \textbf{Models} & \textbf{Reference}  &\textbf{Mathematics} & \textbf{Writing} & \textbf{Programming}  & \textbf{Reasoning} & \textbf{Knowledge-based QA} & \textbf{General} \\
                 &   & (GSM8K)  & (OpenCompass) & (HumanEval) & (HelloSwag) & (TruthfulQA) & (C-Eval) \\
               &     & (Pass@1)  & (Avg Score) & (Pass@1) & (Acc) & (MC2) & (Avg Score) \\
    \midrule
      GPT-4 & \citet{openai2023gpt4}    &\textbf{92.00}& \textbf{62.00} & \textbf{67.00}   & \textbf{91.40}  & 59.00 & 68.70 \\
      ChatGPT &  \citet{chatgpt}   & 57.10 & 48.60 & 48.10   & 79.50  & 47.00     & 54.40\\
      TigerBot-70B-Chat-V2(70B)  & \citet{tigerbot}  & 54.36  & 61.30  & 30.50 & 82.83 & \textbf{75.40} & - \\
      LLaMA2(70B) & \citet{touvron2023llama2} & 60.27 & 51.60 & 29.90   & 82.30  & 56.18     & 55.20 \\
      LLaMA(65B)& \citet{touvron2023llama} & 43.37 & 47.10 & 23.70   & 82.30  & 55.09     & 38.80\\
      Yi(34B) &\citet{yillm} & 50.64 & 48.90  & 26.20  & 82.00  & 56.23 & \textbf{81.40}  \\
      Vicuna(33B)& \citet{vicuna} & 13.72 & 44.90  & 15.20   & 83.00  & 56.16   & 39.80\\
      WizardLM(30B) &\citet{xu2023wizardlm} & 34.42 & - & 26.08 & 76.32  & 49.14 & -  \\
      Moss(16B)&\citet{moss}  & 6.90 & 39.00 & -  & 55.80  & 49.00 & 33.10  \\
      Qwen(14B) & \citet{bai2023qwen}  & 58.98 & 52.7 & 43.90   & 80.20  & 49.43 & 72.10 \\
      Baichuan2(13B)& \citet{yang2023baichuan} & 55.30 & 51.50  & 17.07   & 66.90  & 48.98  & 40.00 \\
      Alpaca(7B) & \citet{alpacablog} & 0.15 & 39.50 & 9.10   & 75.71  & 36.28  &29.90 \\
      ChatGLM3(6B) & \citet{zeng2022glm} & 72.30 & 43.10  & 44.50 & 76.50  & -  & 69.00 \\
    \bottomrule
  \end{tabular}}
\end{table*}

\section{Overall Development Status}\label{overalldevel}
Before exploring the possibilities of building an education system based on LLMs, we first need to investigate the performance of LLMs in capabilities related to education. We select representative benchmarks to assess the current development of LLMs across education-related capabilities. Specifically, we mainly collect the results from three sources: Huggingface\footnote{https://huggingface.co/spaces/HuggingFaceH4/open\_llm\_leaderboard}, OpenCompass\footnote{https://opencompass.org.cn/leaderboard-llm} and C-Eval\footnote{https://cevalbenchmark.com/static/leaderboard.html}. The formal two are comprehensive leaderboards. C-Eval is a Chinese evaluation suite for foundation models spanning 52 diverse disciplines. We collect performance data from popular general LLMs on these benchmarks, and the compiled results are presented in Table~\ref{tab-OverallStatus}, where one can observe that:
\begin{itemize}
    \item It is hard for a single LLM to be superior across all capabilities. Among the current LLMs, GPT-4 has shown the most impressive overall performance. However, utilizing GPT-4 comes with higher associated costs compared to other LLMs, which can be a significant consideration for users and organizations with limited budgets, and it has been surpassed by TigerBot in knowledge-based QA tasks. For mathematics, GPT-4 achieves optimal performance on the representative dataset GSM8K, but it exhibits a higher error rate in basic arithmetic tasks, such as large number multiplication.
    
    \item LLMs still lag significantly behind humans in some crucial abilities. One notable example of this gap is illustrated by their performance on TruthfulQA~\citep{lin2021truthfulqa}, a benchmark designed to evaluate the ability of models to provide truthful and accurate answers, where human achieves achieving 94\% accuracy while GPT-4 only got 59\% correct.  
    
    \item Most LLMs display considerable variation in developing these skills. While certain models (such as Alpaca and Yi) might excel in text comprehension tasks, their effectiveness often diminishes in areas requiring deep understanding and reasoning, like Mathematics and Programming. This reveals the substantial challenges in building a unified education-focused LLM since it may fail in certain areas.
\end{itemize}

\section{Potential of LLM-based Education System} \label{potential}
LLMs can potentially transform online education by understanding a wide range of student questions, similar to human teachers. They aim to provide support across different subjects and skill levels. With the latest developments in LLMs, we suggest two approaches for creating LLM-based education systems. The first involves training a comprehensive and unified LLM that can handle questions from various subjects. The second approach uses a mixture-of-experts (MoE) framework, integrating specialized models to support the system with an LLM controller to manage interactive dialogues with students.

\subsection{Unified Approach}
The most straightforward idea for establishing an LLM-based education system is to train a language model capable of answering students' questions across all subjects. As shown in Figure \ref{fig:twoframes} (a), the foundational capabilities are included in the unified LLM, and the student can directly communicate with it and ask questions.

Research on whether general LLMs can handle educational tasks has been underway. \citet{wang2023chatgpt} introduced three teacher coaching tasks for generative AI: (A) scoring transcript segments using classroom observation instruments, (B) identifying highlights and missed opportunities for effective instructional strategies, and (C) offering actionable suggestions to encourage more student reasoning. And evaluated by human teachers, ChatGPT on these tasks for elementary math classroom transcripts generates responses that are relevant to improving instruction, but they are often not novel or insightful. Beyond that, \citet{phung2023generative} assessed the programming education ability of ChatGPT and GPT-4 by comparing them with human tutors. The result shows that GPT-4 performs way better than ChatGPT, even close to human tutors in some scenarios, while it also highlights some situations in which GPT-4  struggles. In particular, for the grading feedback and task creation scenarios that have a substantial gap in the performance of GPT-4 compared to that of human tutors. 

From the methods proposed by researchers in developing LLMs for educational capabilities, we can extract some common, scalable approaches to lay the groundwork for developing a unified LLM-based educational system:
\begin{itemize}
    \item High-quality demonstrations. While collecting high-quality data from various fields for fine-tuning LLMs is impractical, achieving better responses through prompt engineering as a form of demonstration is feasible.
    \item API Tool Learning. For inherent challenges within LLMs, such as large number computations and the absence of real-time information, these can be addressed by incorporating external APIs as tools.
    \item Search-based methods. Attempts have been made across various fields to improve the task completion accuracy of LLMs using search-based methods, leveraging the probabilistic nature of LLMs. For challenging questions, LLMs might waver among multiple possible responses. Here, employing search-based methods to evaluate and filter all options can effectively enhance accuracy, offering a generally applicable solution.
\end{itemize}
The benefit of developing a unified LLM-based educational system is that it centers around a general LLM handling the core reasoning tasks. This setup means that all major language-based interactions are directly between the LLM and the students, making the system easier to deploy. The most significant effort and resources are invested during the training phase. This critical period is where the LLM gains expert-level skills in a wide range of subjects, preparing it to effectively support and educate students across various disciplines.

\subsection{MoE Approach}
Section ~\ref{capabilities} reviewed the current development of LLMs across various capabilities. Unfortunately, despite the existence of comprehensive language models, such as GPT-4, these models often exhibit notable deficiencies in certain abilities. This situation poses a challenge, indicating that relying solely on an LLM itself for educational guidance involving all these capabilities is currently a difficult task. 
Yet, LLMs can achieve excellent results through fine-tuning individual capabilities, and their ability to comprehend human language is exceptionally strong. Therefore, we can aggregate models with distinct capabilities using a mixture-of-experts approach. By establishing an LLM-based controller for language interaction and task assignment with students, a currently feasible education system can be generated. 

\begin{figure*}[htbp]
    \centering
    \subfloat[Unified Approach Sketch]{\includegraphics[width=0.46\textwidth]{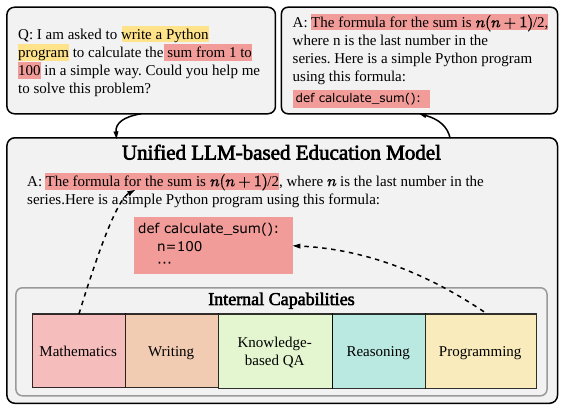}}\hspace{25pt}
    \subfloat[MoE Approach Sketch]{\includegraphics[width=0.46\textwidth]{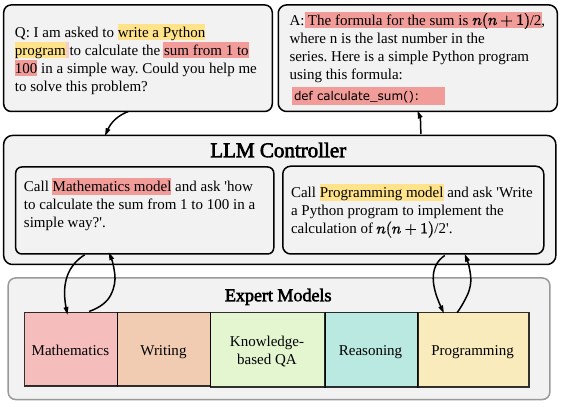}}
    \\

    \centering
    \caption{Two frameworks towards LLM-based educational framework. (a) depicts the unified approach, where a single LLM addresses all aspects of educational-related queries, utilizing its internal capabilities such as mathematics, writing, knowledge-based question answering, reasoning, and programming. (b) illustrates the Mixture of Experts (MoE) approach, where an LLM controller is tasked with task distribution, delegating specific questions to specialized expert models that are proficient in individual areas. }
    \label{fig:twoframes}
\end{figure*}

An education framework implemented with a mixture-of-experts (MoE) approach is illustrated in Figure~\ref{fig:twoframes}(b), consisting of multiple models that excel in individual capabilities (not necessarily LLMs) and an LLM controller. The controller is mainly responsible for three tasks: 
\begin{itemize}
    \item Understand the student's request and decide which specific area or areas the request is about. 
    \item Re-form the request to fit the input of the specific areas' expert models.
    \item Aggregate the output of the related experts and generate the final response to the student. 
\end{itemize}

The advantage of the MoE approach is that training is less challenging. The result is a suite of models, each excelling in its specific domain or capability, which, when combined, offer a comprehensive educational tool. This specialization means training can be more focused and less onerous, optimizing resources towards developing excellence in distinct areas of knowledge and skills. However, one significant drawback is the increased potential for misunderstandings or errors during the system's inference phase, primarily due to the complexity of interactions between the different specialized models and the LLM controller. Errors can arise from the LLM controller misinterpreting student inputs or incorrectly assigning tasks to the specialized models. Moreover, integrating outputs from various experts into a coherent response can also introduce discrepancies, as differences in context or terminology used by each model can lead to inconsistencies in the overall communication with students.

Despite these challenges, the approach represents a practical pathway toward realizing an LLM-based educational assistant system. By leveraging specialized models for different capabilities, it's possible to create a more flexible and efficient system that can adapt to a wide range of educational needs and learning styles. The key to success lies in improving the integration and communication between the specialized models and the overarching LLM controller, ensuring that the system can handle complex inquiries and deliver accurate, useful responses to students. Currently, this approach appears to be a viable strategy for achieving the ambitious goal of an effective LLM-based educational assistant, promising a future where personalized education is accessible and adaptable to every learner's need.

\section{Challenges and Future Directions} \label{future}
Recently, more and more researchers have been trying to apply LLMs to handle education tasks, such as course design, student evaluation, lesson plan design, and others. Nevertheless, there are still numerous challenges and opportunities that need to be addressed.





\begin{itemize}
    \item \textbf{Planning for Students.} Solving subject-related questions for students can significantly address the issue of students not receiving targeted guidance from teachers. Furthermore, a higher-level task involves assessing students' knowledge status and planning their learning paths. These tasks are continuously evolving in the era of deep learning, with the adaptation and application of LLMs in these areas requiring further exploration. The primary challenge in planning learning paths for students lies in integrating knowledge from two aspects: first, the human knowledge system, which involves the structural relationships between knowledge points, requiring LLMs to understand the meaning of these knowledge points. Second, the personalized information of students, including their knowledge state, learning interests, and habits. Previous deep learning models for this task have been trained on sequences of student behaviors, which are often constructed as IDs rather than text. Since the foundation of LLMs is their ability to process text, the gap in data form presents a significant challenge in applying LLMs to this task.
    \item \textbf{Interdisciplinary Reasoning Ability.} Students may encounter interdisciplinary reasoning problems during real-world learning, requiring the education system to integrate multiple capabilities to formulate responses. As illustrated in Figure~\ref{fig-caps}, the student intends to write a program to solve a mathematical problem, and the model needs first to comprehend the mathematical problem, devise a solution, and then generate the code. This process necessitates the model to synthesize both mathematical and programming capabilities. However, there is currently limited research in the integration of multiple interdisciplinary capabilities for LLMs at this stage, including both datasets and algorithms. \citet{boyko2023interdisciplinary} examined how LLMs augment scientific inquiry, code development, scientific writing process, etc., and they propose that LLMs can foster interdisciplinary work by bridging knowledge gaps across scientific fields. However, they mainly discuss the LLMs' ability to help researchers' interdisciplinary collaboration instead of their ability to answer interdisciplinary questions. Cultivating an LLM to obtain this ability would help to develop a unified education system, which is an essential research direction.
    \item \textbf{Student Modeling.} Before the era of LLMs, in the age of deep learning, modeling student behavior was primarily achieved through sequential models, such as RNNs~\citep{sherstinsky2020fundamentals} and Transformers. A drawback of this approach was the inability to obtain student feedback, and the results lacked interpretability. Establishing an LLM-based education system allows students to articulate their personalized needs through dialog. Through such conversations, we can extract or infer personalized features about students, such as their current mastery of topics and preferences in learning styles.  Besides modeling students from conversations, some researches~\citep{aher2023using, argyle2023out} have shown that LLMs have certain abilities in simulating humans and generating human samples. Applying to education, this ability indicates a potential for LLM-based student simulation. In this way, for the students with few interaction records, the LLM-based simulator could generate more samples and provide data to help the expert model better understand the student. It could help human teachers develop teaching skills better.

    \item \textbf{Social Bias of LLMs}  Even after training through Reinforcement Learning from Human Feedback (RLHF)~\citep{ouyang2022training}, LLMs can to some extent avoid answers that do not align with human cultural habits and values~\citep{jiang2022communitylm, feng2023pretraining}, it has been observed that LLMs still manifest a certain degree of value bias in their responses. ~\citet{feng2023pretraining} pointed out that the training of LLMs can lead to a certain degree of political bias. In the field of education, although most questions posed by students are related to scientific knowledge, issues such as writing and text reasoning should ideally be avoided by researchers developing foundational LLM models.  In educational applications, the social biases inherent in LLMs pose a risk of inadvertently impart imparting skewed value systems to students. To safeguard the educational integrity and ensure the neutral and fair dissemination of knowledge, it is crucial to implement stringent measures. These could include developing advanced content review systems, establishing clear guidelines for the ethical use of LLMs in educational settings, and continuously monitoring the quality and nature of the LLM-generated content. Through these efforts, the educational community can leverage the benefits of LLMs while minimizing the risk of perpetuating biases, thus maintaining a balanced and objective learning environment.
    
    \item \textbf{Preventing Cheating in Education.} The texts generated by LLMs are indistinguishable from or even surpass those produced by humans in terms of fluency and usage. Although the primary aim of this article is to survey the development of LLMs in educational capacities, offering insights for the creation of an educational supermodel, it's crucial to recognize that in certain educational contexts, the over-reliance on LLMs is not desirable as it could hinder the natural learning process. For instance, while it's acceptable for students to seek assistance from LLMs to aid understanding during homework tasks, relying on LLMs to complete assignments without thoughtful consideration prevents students from receiving the necessary practice and learning. Therefore, identifying content generated by LLMs holds significant importance in the educational domain to prevent cheating and ensure the integrity of the learning process. By striking a balance between leveraging LLMs for educational enhancement and maintaining rigorous educational standards, educators and technologists can create an environment where students benefit from technology without compromising their learning journey. Recent studies have proposed detectors to identify LLM-generated texts. The foundational ideas could be broadly categorized into two primary strategies: statistical outlier detection methods and supervised classifiers. The former strategy focuses on uncovering statistical differences in linguistic features between texts written by humans and those generated by LLMs. This involves analyzing patterns, such as syntactic structures, vocabulary diversity, and stylistic nuances, that distinguish LLM-generated texts from human-crafted writings. These statistical indicators serve as markers for automated systems to detect content that deviates from human norms, potentially signaling LLM involvement. On the other hand, supervised classifiers rely on a different mechanism. This approach employs machine learning algorithms that have been trained on a labeled dataset containing examples of both human-written and LLM-generated texts. The battle against detecting LLM-generated texts is dynamic, necessitating ongoing research and adaptation of detection methodologies. As LLMs become increasingly sophisticated, the strategies for distinguishing their outputs from human-created content will need to evolve, embracing a combination of statistical insights, machine learning innovations, and perhaps new, yet-to-be-discovered methods.
    
    \item \textbf{Multi-modal Education.} In education, multi-modal information is common, like geometry problems combining text and images or textbook concepts with illustrations. Building a general intelligent education system requires handling such multi-modal data. Notably, the development of multi-modal LLMs is rapidly advancing~\citep{ye2023mplug, du2022survey}. Different kinds of architectures and pre-train tasks are proposed~\citep{du2022survey}. However, the education domain often exhibits unique distribution characteristics in multi-modal information. Firstly, in education, images and text often have a high level of detail matching; for example, geometry questions often describe the specific parameters of shapes in images in great detail. Therefore, multi-modal large models need to have a high capability of capturing details in image information. Secondly, the multi-modal information in education often requires the model to have a high capacity for cross-modal reasoning, but such data is less common in multi-modal datasets, leading to a potential shortfall in the reasoning capabilities of multi-modal language models across different modalities. Addressing this gap may require targeted datasets, and inspired by Chain of Thought (CoT) and its variants, the data should ideally contain detailed steps of multi-modal reasoning. Efforts are currently being made to address these data deficiencies in the field. Moreover, the characteristics of image and text data in education could limit the choice of structures for multi-modal models. For instance, a popular approach in the field of multi-modal large models involves dividing images into patches to create ``image tokens'',~\citep{mckinzie2024mm1} which are then processed alongside text tokens as input. However, in the educational context, such division might disrupt certain key geometric structures within the images, thereby affecting their interpretation. This drawback could be more pronounced in educational multi-modal scenarios. 
\end{itemize}

\section{Conclusion} \label{conclusion}
In this paper, we presented an overview of the development of the LLM-based education system. We first reviewed the important development of LLMs' education-related abilities. Then, we analyzed the potential of it and proposed two different ways of building such a system. We also highlighted the future directions that are worth working on. We hope this survey provides some insight into future research in this direction.

\section*{Acknowledgments}
The SJTU team is partially supported by National Natural Science Foundation of China (62177033).

\bibliographystyle{ACM-Reference-Format}
\bibliography{sample-base}


\begin{thebibliography}{163}


\ifx \showCODEN    \undefined \def \showCODEN     #1{\unskip}     \fi
\ifx \showDOI      \undefined \def \showDOI       #1{#1}\fi
\ifx \showISBNx    \undefined \def \showISBNx     #1{\unskip}     \fi
\ifx \showISBNxiii \undefined \def \showISBNxiii  #1{\unskip}     \fi
\ifx \showISSN     \undefined \def \showISSN      #1{\unskip}     \fi
\ifx \showLCCN     \undefined \def \showLCCN      #1{\unskip}     \fi
\ifx \shownote     \undefined \def \shownote      #1{#1}          \fi
\ifx \showarticletitle \undefined \def \showarticletitle #1{#1}   \fi
\ifx \showURL      \undefined \def \showURL       {\relax}        \fi
\providecommand\bibfield[2]{#2}
\providecommand\bibinfo[2]{#2}
\providecommand\natexlab[1]{#1}
\providecommand\showeprint[2][]{arXiv:#2}

\bibitem[Abdelrahman et~al\mbox{.}(2023)]%
        {abdelrahman2023knowledge}
\bibfield{author}{\bibinfo{person}{Ghodai Abdelrahman}, \bibinfo{person}{Qing Wang}, {and} \bibinfo{person}{Bernardo Nunes}.} \bibinfo{year}{2023}\natexlab{}.
\newblock \showarticletitle{Knowledge tracing: A survey}.
\newblock \bibinfo{journal}{\emph{Comput. Surveys}} \bibinfo{volume}{55}, \bibinfo{number}{11} (\bibinfo{year}{2023}), \bibinfo{pages}{1--37}.
\newblock


\bibitem[Achiam et~al\mbox{.}(2023)]%
        {achiam2023gpt}
\bibfield{author}{\bibinfo{person}{Josh Achiam}, \bibinfo{person}{Steven Adler}, \bibinfo{person}{Sandhini Agarwal}, \bibinfo{person}{Lama Ahmad}, \bibinfo{person}{Ilge Akkaya}, \bibinfo{person}{Florencia~Leoni Aleman}, \bibinfo{person}{Diogo Almeida}, \bibinfo{person}{Janko Altenschmidt}, \bibinfo{person}{Sam Altman}, \bibinfo{person}{Shyamal Anadkat}, {et~al\mbox{.}}} \bibinfo{year}{2023}\natexlab{}.
\newblock \showarticletitle{Gpt-4 technical report}.
\newblock \bibinfo{journal}{\emph{arXiv preprint arXiv:2303.08774}} (\bibinfo{year}{2023}).
\newblock


\bibitem[Aher et~al\mbox{.}(2023)]%
        {aher2023using}
\bibfield{author}{\bibinfo{person}{Gati~V Aher}, \bibinfo{person}{Rosa~I Arriaga}, {and} \bibinfo{person}{Adam~Tauman Kalai}.} \bibinfo{year}{2023}\natexlab{}.
\newblock \showarticletitle{Using large language models to simulate multiple humans and replicate human subject studies}. In \bibinfo{booktitle}{\emph{International Conference on Machine Learning}}. PMLR, \bibinfo{pages}{337--371}.
\newblock


\bibitem[AL-Smadi(2023)]%
        {al2023chatgpt}
\bibfield{author}{\bibinfo{person}{Mohammad AL-Smadi}.} \bibinfo{year}{2023}\natexlab{}.
\newblock \showarticletitle{ChatGPT and Beyond: The Generative AI Revolution in Education}.
\newblock \bibinfo{journal}{\emph{arXiv preprint arXiv:2311.15198}} (\bibinfo{year}{2023}).
\newblock


\bibitem[Arefeen et~al\mbox{.}(2023)]%
        {arefeen2023leancontext}
\bibfield{author}{\bibinfo{person}{Md~Adnan Arefeen}, \bibinfo{person}{Biplob Debnath}, {and} \bibinfo{person}{Srimat Chakradhar}.} \bibinfo{year}{2023}\natexlab{}.
\newblock \showarticletitle{Leancontext: Cost-efficient domain-specific question answering using llms}.
\newblock \bibinfo{journal}{\emph{arXiv preprint arXiv:2309.00841}} (\bibinfo{year}{2023}).
\newblock


\bibitem[Argyle et~al\mbox{.}(2023)]%
        {argyle2023out}
\bibfield{author}{\bibinfo{person}{Lisa~P Argyle}, \bibinfo{person}{Ethan~C Busby}, \bibinfo{person}{Nancy Fulda}, \bibinfo{person}{Joshua~R Gubler}, \bibinfo{person}{Christopher Rytting}, {and} \bibinfo{person}{David Wingate}.} \bibinfo{year}{2023}\natexlab{}.
\newblock \showarticletitle{Out of one, many: Using language models to simulate human samples}.
\newblock \bibinfo{journal}{\emph{Political Analysis}} \bibinfo{volume}{31}, \bibinfo{number}{3} (\bibinfo{year}{2023}), \bibinfo{pages}{337--351}.
\newblock


\bibitem[Bai et~al\mbox{.}(2023)]%
        {bai2023qwen}
\bibfield{author}{\bibinfo{person}{Jinze Bai}, \bibinfo{person}{Shuai Bai}, \bibinfo{person}{Yunfei Chu}, \bibinfo{person}{Zeyu Cui}, \bibinfo{person}{Kai Dang}, \bibinfo{person}{Xiaodong Deng}, \bibinfo{person}{Yang Fan}, \bibinfo{person}{Wenbin Ge}, \bibinfo{person}{Yu Han}, \bibinfo{person}{Fei Huang}, {et~al\mbox{.}}} \bibinfo{year}{2023}\natexlab{}.
\newblock \showarticletitle{Qwen technical report}.
\newblock \bibinfo{journal}{\emph{arXiv preprint arXiv:2309.16609}} (\bibinfo{year}{2023}).
\newblock


\bibitem[Barras et~al\mbox{.}(1997)]%
        {barras1997coq}
\bibfield{author}{\bibinfo{person}{Bruno Barras}, \bibinfo{person}{Samuel Boutin}, \bibinfo{person}{Cristina Cornes}, \bibinfo{person}{Judica{\"e}l Courant}, \bibinfo{person}{Jean-Christophe Filliatre}, \bibinfo{person}{Eduardo Gimenez}, \bibinfo{person}{Hugo Herbelin}, \bibinfo{person}{Gerard Huet}, \bibinfo{person}{Cesar Munoz}, \bibinfo{person}{Chetan Murthy}, {et~al\mbox{.}}} \bibinfo{year}{1997}\natexlab{}.
\newblock \emph{\bibinfo{title}{The Coq proof assistant reference manual: Version 6.1}}.
\newblock \bibinfo{thesistype}{Ph.\,D. Dissertation}. \bibinfo{school}{Inria}.
\newblock


\bibitem[Besta et~al\mbox{.}(2023)]%
        {besta2023graph}
\bibfield{author}{\bibinfo{person}{Maciej Besta}, \bibinfo{person}{Nils Blach}, \bibinfo{person}{Ales Kubicek}, \bibinfo{person}{Robert Gerstenberger}, \bibinfo{person}{Lukas Gianinazzi}, \bibinfo{person}{Joanna Gajda}, \bibinfo{person}{Tomasz Lehmann}, \bibinfo{person}{Michal Podstawski}, \bibinfo{person}{Hubert Niewiadomski}, \bibinfo{person}{Piotr Nyczyk}, {et~al\mbox{.}}} \bibinfo{year}{2023}\natexlab{}.
\newblock \showarticletitle{Graph of thoughts: Solving elaborate problems with large language models}.
\newblock \bibinfo{journal}{\emph{arXiv preprint arXiv:2308.09687}} (\bibinfo{year}{2023}).
\newblock


\bibitem[Borgeaud et~al\mbox{.}(2022)]%
        {borgeaud2022improving}
\bibfield{author}{\bibinfo{person}{Sebastian Borgeaud}, \bibinfo{person}{Arthur Mensch}, \bibinfo{person}{Jordan Hoffmann}, \bibinfo{person}{Trevor Cai}, \bibinfo{person}{Eliza Rutherford}, \bibinfo{person}{Katie Millican}, \bibinfo{person}{George~Bm Van Den~Driessche}, \bibinfo{person}{Jean-Baptiste Lespiau}, \bibinfo{person}{Bogdan Damoc}, \bibinfo{person}{Aidan Clark}, {et~al\mbox{.}}} \bibinfo{year}{2022}\natexlab{}.
\newblock \showarticletitle{Improving language models by retrieving from trillions of tokens}. In \bibinfo{booktitle}{\emph{International conference on machine learning}}. PMLR, \bibinfo{pages}{2206--2240}.
\newblock


\bibitem[Boyko et~al\mbox{.}(2023)]%
        {boyko2023interdisciplinary}
\bibfield{author}{\bibinfo{person}{James Boyko}, \bibinfo{person}{Joseph Cohen}, \bibinfo{person}{Nathan Fox}, \bibinfo{person}{Maria~Han Veiga}, \bibinfo{person}{Jennifer~I Li}, \bibinfo{person}{Jing Liu}, \bibinfo{person}{Bernardo Modenesi}, \bibinfo{person}{Andreas~H Rauch}, \bibinfo{person}{Kenneth~N Reid}, \bibinfo{person}{Soumi Tribedi}, {et~al\mbox{.}}} \bibinfo{year}{2023}\natexlab{}.
\newblock \showarticletitle{An Interdisciplinary Outlook on Large Language Models for Scientific Research}.
\newblock \bibinfo{journal}{\emph{arXiv preprint arXiv:2311.04929}} (\bibinfo{year}{2023}).
\newblock


\bibitem[Chang et~al\mbox{.}(2023)]%
        {chang2023survey}
\bibfield{author}{\bibinfo{person}{Yupeng Chang}, \bibinfo{person}{Xu Wang}, \bibinfo{person}{Jindong Wang}, \bibinfo{person}{Yuan Wu}, \bibinfo{person}{Kaijie Zhu}, \bibinfo{person}{Hao Chen}, \bibinfo{person}{Linyi Yang}, \bibinfo{person}{Xiaoyuan Yi}, \bibinfo{person}{Cunxiang Wang}, \bibinfo{person}{Yidong Wang}, {et~al\mbox{.}}} \bibinfo{year}{2023}\natexlab{}.
\newblock \showarticletitle{A survey on evaluation of large language models}.
\newblock \bibinfo{journal}{\emph{arXiv preprint arXiv:2307.03109}} (\bibinfo{year}{2023}).
\newblock


\bibitem[Chen et~al\mbox{.}(2022b)]%
        {chen2022codet}
\bibfield{author}{\bibinfo{person}{Bei Chen}, \bibinfo{person}{Fengji Zhang}, \bibinfo{person}{Anh Nguyen}, \bibinfo{person}{Daoguang Zan}, \bibinfo{person}{Zeqi Lin}, \bibinfo{person}{Jian-Guang Lou}, {and} \bibinfo{person}{Weizhu Chen}.} \bibinfo{year}{2022}\natexlab{b}.
\newblock \bibinfo{title}{CodeT: Code Generation with Generated Tests}.
\newblock
\newblock
\showeprint[arxiv]{2207.10397}~[cs.CL]


\bibitem[Chen et~al\mbox{.}(2022a)]%
        {chen2022unigeo}
\bibfield{author}{\bibinfo{person}{Jiaqi Chen}, \bibinfo{person}{Tong Li}, \bibinfo{person}{Jinghui Qin}, \bibinfo{person}{Pan Lu}, \bibinfo{person}{Liang Lin}, \bibinfo{person}{Chongyu Chen}, {and} \bibinfo{person}{Xiaodan Liang}.} \bibinfo{year}{2022}\natexlab{a}.
\newblock \showarticletitle{UniGeo: Unifying Geometry Logical Reasoning via Reformulating Mathematical Expression}.
\newblock \bibinfo{journal}{\emph{arXiv preprint arXiv:2212.02746}} (\bibinfo{year}{2022}).
\newblock


\bibitem[Chen et~al\mbox{.}(2021a)]%
        {codex}
\bibfield{author}{\bibinfo{person}{Mark Chen}, \bibinfo{person}{Jerry Tworek}, \bibinfo{person}{Heewoo Jun}, \bibinfo{person}{Qiming Yuan}, \bibinfo{person}{Henrique~Ponde de Oliveira~Pinto}, \bibinfo{person}{Jared Kaplan}, \bibinfo{person}{Harri Edwards}, \bibinfo{person}{Yuri Burda}, \bibinfo{person}{Nicholas Joseph}, \bibinfo{person}{Greg Brockman}, \bibinfo{person}{Alex Ray}, \bibinfo{person}{Raul Puri}, \bibinfo{person}{Gretchen Krueger}, \bibinfo{person}{Michael Petrov}, \bibinfo{person}{Heidy Khlaaf}, \bibinfo{person}{Girish Sastry}, \bibinfo{person}{Pamela Mishkin}, \bibinfo{person}{Brooke Chan}, \bibinfo{person}{Scott Gray}, \bibinfo{person}{Nick Ryder}, \bibinfo{person}{Mikhail Pavlov}, \bibinfo{person}{Alethea Power}, \bibinfo{person}{Lukasz Kaiser}, \bibinfo{person}{Mohammad Bavarian}, \bibinfo{person}{Clemens Winter}, \bibinfo{person}{Philippe Tillet}, \bibinfo{person}{Felipe~Petroski Such}, \bibinfo{person}{Dave Cummings}, \bibinfo{person}{Matthias Plappert}, \bibinfo{person}{Fotios Chantzis},
  \bibinfo{person}{Elizabeth Barnes}, \bibinfo{person}{Ariel Herbert-Voss}, \bibinfo{person}{William~Hebgen Guss}, \bibinfo{person}{Alex Nichol}, \bibinfo{person}{Alex Paino}, \bibinfo{person}{Nikolas Tezak}, \bibinfo{person}{Jie Tang}, \bibinfo{person}{Igor Babuschkin}, \bibinfo{person}{Suchir Balaji}, \bibinfo{person}{Shantanu Jain}, \bibinfo{person}{William Saunders}, \bibinfo{person}{Christopher Hesse}, \bibinfo{person}{Andrew~N. Carr}, \bibinfo{person}{Jan Leike}, \bibinfo{person}{Josh Achiam}, \bibinfo{person}{Vedant Misra}, \bibinfo{person}{Evan Morikawa}, \bibinfo{person}{Alec Radford}, \bibinfo{person}{Matthew Knight}, \bibinfo{person}{Miles Brundage}, \bibinfo{person}{Mira Murati}, \bibinfo{person}{Katie Mayer}, \bibinfo{person}{Peter Welinder}, \bibinfo{person}{Bob McGrew}, \bibinfo{person}{Dario Amodei}, \bibinfo{person}{Sam McCandlish}, \bibinfo{person}{Ilya Sutskever}, {and} \bibinfo{person}{Wojciech Zaremba}.} \bibinfo{year}{2021}\natexlab{a}.
\newblock \bibinfo{title}{Evaluating Large Language Models Trained on Code}.
\newblock
\newblock
\showeprint[arxiv]{2107.03374}~[cs.LG]


\bibitem[Chen et~al\mbox{.}(2021b)]%
        {chen2021evaluating}
\bibfield{author}{\bibinfo{person}{Mark Chen}, \bibinfo{person}{Jerry Tworek}, \bibinfo{person}{Heewoo Jun}, \bibinfo{person}{Qiming Yuan}, \bibinfo{person}{Henrique~Ponde de Oliveira~Pinto}, \bibinfo{person}{Jared Kaplan}, \bibinfo{person}{Harri Edwards}, \bibinfo{person}{Yuri Burda}, \bibinfo{person}{Nicholas Joseph}, \bibinfo{person}{Greg Brockman}, \bibinfo{person}{Alex Ray}, \bibinfo{person}{Raul Puri}, \bibinfo{person}{Gretchen Krueger}, \bibinfo{person}{Michael Petrov}, \bibinfo{person}{Heidy Khlaaf}, \bibinfo{person}{Girish Sastry}, \bibinfo{person}{Pamela Mishkin}, \bibinfo{person}{Brooke Chan}, \bibinfo{person}{Scott Gray}, \bibinfo{person}{Nick Ryder}, \bibinfo{person}{Mikhail Pavlov}, \bibinfo{person}{Alethea Power}, \bibinfo{person}{Lukasz Kaiser}, \bibinfo{person}{Mohammad Bavarian}, \bibinfo{person}{Clemens Winter}, \bibinfo{person}{Philippe Tillet}, \bibinfo{person}{Felipe~Petroski Such}, \bibinfo{person}{Dave Cummings}, \bibinfo{person}{Matthias Plappert}, \bibinfo{person}{Fotios Chantzis},
  \bibinfo{person}{Elizabeth Barnes}, \bibinfo{person}{Ariel Herbert-Voss}, \bibinfo{person}{William~Hebgen Guss}, \bibinfo{person}{Alex Nichol}, \bibinfo{person}{Alex Paino}, \bibinfo{person}{Nikolas Tezak}, \bibinfo{person}{Jie Tang}, \bibinfo{person}{Igor Babuschkin}, \bibinfo{person}{Suchir Balaji}, \bibinfo{person}{Shantanu Jain}, \bibinfo{person}{William Saunders}, \bibinfo{person}{Christopher Hesse}, \bibinfo{person}{Andrew~N. Carr}, \bibinfo{person}{Jan Leike}, \bibinfo{person}{Josh Achiam}, \bibinfo{person}{Vedant Misra}, \bibinfo{person}{Evan Morikawa}, \bibinfo{person}{Alec Radford}, \bibinfo{person}{Matthew Knight}, \bibinfo{person}{Miles Brundage}, \bibinfo{person}{Mira Murati}, \bibinfo{person}{Katie Mayer}, \bibinfo{person}{Peter Welinder}, \bibinfo{person}{Bob McGrew}, \bibinfo{person}{Dario Amodei}, \bibinfo{person}{Sam McCandlish}, \bibinfo{person}{Ilya Sutskever}, {and} \bibinfo{person}{Wojciech Zaremba}.} \bibinfo{year}{2021}\natexlab{b}.
\newblock \bibinfo{title}{Evaluating Large Language Models Trained on Code}.
\newblock
\newblock
\showeprint[arxiv]{2107.03374}~[cs.LG]


\bibitem[Chen et~al\mbox{.}(2024)]%
        {chen2024boosting}
\bibfield{author}{\bibinfo{person}{Sijia Chen}, \bibinfo{person}{Baochun Li}, {and} \bibinfo{person}{Di Niu}.} \bibinfo{year}{2024}\natexlab{}.
\newblock \showarticletitle{Boosting of thoughts: Trial-and-error problem solving with large language models}.
\newblock \bibinfo{journal}{\emph{arXiv preprint arXiv:2402.11140}} (\bibinfo{year}{2024}).
\newblock


\bibitem[Chen et~al\mbox{.}(2023)]%
        {chen2023set}
\bibfield{author}{\bibinfo{person}{Xianyu Chen}, \bibinfo{person}{Jian Shen}, \bibinfo{person}{Wei Xia}, \bibinfo{person}{Jiarui Jin}, \bibinfo{person}{Yakun Song}, \bibinfo{person}{Weinan Zhang}, \bibinfo{person}{Weiwen Liu}, \bibinfo{person}{Menghui Zhu}, \bibinfo{person}{Ruiming Tang}, \bibinfo{person}{Kai Dong}, {et~al\mbox{.}}} \bibinfo{year}{2023}\natexlab{}.
\newblock \showarticletitle{Set-to-sequence ranking-based concept-aware learning path recommendation}. In \bibinfo{booktitle}{\emph{Proceedings of the AAAI Conference on Artificial Intelligence}}, Vol.~\bibinfo{volume}{37}. \bibinfo{pages}{5027--5035}.
\newblock


\bibitem[Choi et~al\mbox{.}(2023)]%
        {choi2023conversational}
\bibfield{author}{\bibinfo{person}{Stephen Choi}, \bibinfo{person}{William Gazeley}, \bibinfo{person}{Siu~Ho Wong}, {and} \bibinfo{person}{Tingting Li}.} \bibinfo{year}{2023}\natexlab{}.
\newblock \bibinfo{title}{Conversational Financial Information Retrieval Model (ConFIRM)}.
\newblock
\newblock
\showeprint[arxiv]{2310.13001}~[cs.IR]


\bibitem[Corbett and Anderson(1994)]%
        {corbett1994knowledge}
\bibfield{author}{\bibinfo{person}{Albert~T Corbett} {and} \bibinfo{person}{John~R Anderson}.} \bibinfo{year}{1994}\natexlab{}.
\newblock \showarticletitle{Knowledge tracing: Modeling the acquisition of procedural knowledge}.
\newblock \bibinfo{journal}{\emph{User modeling and user-adapted interaction}}  \bibinfo{volume}{4} (\bibinfo{year}{1994}), \bibinfo{pages}{253--278}.
\newblock


\bibitem[Cunningham et~al\mbox{.}(2023)]%
        {cunningham2023towards}
\bibfield{author}{\bibinfo{person}{Garett Cunningham}, \bibinfo{person}{Razvan~C Bunescu}, {and} \bibinfo{person}{David Juedes}.} \bibinfo{year}{2023}\natexlab{}.
\newblock \showarticletitle{Towards Autoformalization of Mathematics and Code Correctness: Experiments with Elementary Proofs}.
\newblock \bibinfo{journal}{\emph{arXiv preprint arXiv:2301.02195}} (\bibinfo{year}{2023}).
\newblock


\bibitem[de~Moura et~al\mbox{.}(2015)]%
        {de2015lean}
\bibfield{author}{\bibinfo{person}{Leonardo de Moura}, \bibinfo{person}{Soonho Kong}, \bibinfo{person}{Jeremy Avigad}, \bibinfo{person}{Floris Van~Doorn}, {and} \bibinfo{person}{Jakob von Raumer}.} \bibinfo{year}{2015}\natexlab{}.
\newblock \showarticletitle{The Lean theorem prover (system description)}. In \bibinfo{booktitle}{\emph{Automated Deduction-CADE-25: 25th International Conference on Automated Deduction, Berlin, Germany, August 1-7, 2015, Proceedings 25}}. Springer, \bibinfo{pages}{378--388}.
\newblock


\bibitem[Diao et~al\mbox{.}(2023)]%
        {diao2023active}
\bibfield{author}{\bibinfo{person}{Shizhe Diao}, \bibinfo{person}{Pengcheng Wang}, \bibinfo{person}{Yong Lin}, {and} \bibinfo{person}{Tong Zhang}.} \bibinfo{year}{2023}\natexlab{}.
\newblock \showarticletitle{Active prompting with chain-of-thought for large language models}.
\newblock \bibinfo{journal}{\emph{arXiv preprint arXiv:2302.12246}} (\bibinfo{year}{2023}).
\newblock


\bibitem[Ding et~al\mbox{.}(2023)]%
        {ding2023crosscodeeval}
\bibfield{author}{\bibinfo{person}{Yangruibo Ding}, \bibinfo{person}{Zijian Wang}, \bibinfo{person}{Wasi~Uddin Ahmad}, \bibinfo{person}{Hantian Ding}, \bibinfo{person}{Ming Tan}, \bibinfo{person}{Nihal Jain}, \bibinfo{person}{Murali~Krishna Ramanathan}, \bibinfo{person}{Ramesh Nallapati}, \bibinfo{person}{Parminder Bhatia}, \bibinfo{person}{Dan Roth}, {and} \bibinfo{person}{Bing Xiang}.} \bibinfo{year}{2023}\natexlab{}.
\newblock \bibinfo{title}{CrossCodeEval: A Diverse and Multilingual Benchmark for Cross-File Code Completion}.
\newblock
\newblock
\showeprint[arxiv]{2310.11248}~[cs.LG]


\bibitem[Dong et~al\mbox{.}(2022)]%
        {surveyDong2023}
\bibfield{author}{\bibinfo{person}{Chenhe Dong}, \bibinfo{person}{Yinghui Li}, \bibinfo{person}{Haifan Gong}, \bibinfo{person}{Miaoxin Chen}, \bibinfo{person}{Junxin Li}, \bibinfo{person}{Ying Shen}, {and} \bibinfo{person}{Min Yang}.} \bibinfo{year}{2022}\natexlab{}.
\newblock \showarticletitle{A Survey of Natural Language Generation}.
\newblock \bibinfo{journal}{\emph{ACM Comput. Surv.}} \bibinfo{volume}{55}, \bibinfo{number}{8}, Article \bibinfo{articleno}{173} (\bibinfo{date}{dec} \bibinfo{year}{2022}), \bibinfo{numpages}{38}~pages.
\newblock
\showISSN{0360-0300}
\urldef\tempurl%
\url{https://doi.org/10.1145/3554727}
\showDOI{\tempurl}


\bibitem[Du et~al\mbox{.}(2022)]%
        {du2022survey}
\bibfield{author}{\bibinfo{person}{Yifan Du}, \bibinfo{person}{Zikang Liu}, \bibinfo{person}{Junyi Li}, {and} \bibinfo{person}{Wayne~Xin Zhao}.} \bibinfo{year}{2022}\natexlab{}.
\newblock \showarticletitle{A survey of vision-language pre-trained models}.
\newblock \bibinfo{journal}{\emph{arXiv preprint arXiv:2202.10936}} (\bibinfo{year}{2022}).
\newblock


\bibitem[Fan et~al\mbox{.}(2023)]%
        {fan2023grammargpt}
\bibfield{author}{\bibinfo{person}{Yaxin Fan}, \bibinfo{person}{Feng Jiang}, \bibinfo{person}{Peifeng Li}, {and} \bibinfo{person}{Haizhou Li}.} \bibinfo{year}{2023}\natexlab{}.
\newblock \showarticletitle{GrammarGPT: Exploring Open-Source LLMs for Native Chinese Grammatical Error Correction with Supervised Fine-Tuning}. In \bibinfo{booktitle}{\emph{CCF International Conference on Natural Language Processing and Chinese Computing}}. Springer, \bibinfo{pages}{69--80}.
\newblock


\bibitem[Fang et~al\mbox{.}(2023)]%
        {fang2023chatgpt}
\bibfield{author}{\bibinfo{person}{Tao Fang}, \bibinfo{person}{Shu Yang}, \bibinfo{person}{Kaixin Lan}, \bibinfo{person}{Derek~F Wong}, \bibinfo{person}{Jinpeng Hu}, \bibinfo{person}{Lidia~S Chao}, {and} \bibinfo{person}{Yue Zhang}.} \bibinfo{year}{2023}\natexlab{}.
\newblock \showarticletitle{Is chatgpt a highly fluent grammatical error correction system? a comprehensive evaluation}.
\newblock \bibinfo{journal}{\emph{arXiv preprint arXiv:2304.01746}} (\bibinfo{year}{2023}).
\newblock


\bibitem[Feng et~al\mbox{.}(2023a)]%
        {feng2023improving}
\bibfield{author}{\bibinfo{person}{Huawen Feng}, \bibinfo{person}{Yan Fan}, \bibinfo{person}{Xiong Liu}, \bibinfo{person}{Ting-En Lin}, \bibinfo{person}{Zekun Yao}, \bibinfo{person}{Yuchuan Wu}, \bibinfo{person}{Fei Huang}, \bibinfo{person}{Yongbin Li}, {and} \bibinfo{person}{Qianli Ma}.} \bibinfo{year}{2023}\natexlab{a}.
\newblock \showarticletitle{Improving Factual Consistency of Text Summarization by Adversarially Decoupling Comprehension and Embellishment Abilities of LLMs}.
\newblock \bibinfo{journal}{\emph{arXiv preprint arXiv:2310.19347}} (\bibinfo{year}{2023}).
\newblock


\bibitem[Feng et~al\mbox{.}(2023b)]%
        {feng2023pretraining}
\bibfield{author}{\bibinfo{person}{Shangbin Feng}, \bibinfo{person}{Chan~Young Park}, \bibinfo{person}{Yuhan Liu}, {and} \bibinfo{person}{Yulia Tsvetkov}.} \bibinfo{year}{2023}\natexlab{b}.
\newblock \showarticletitle{From pretraining data to language models to downstream tasks: Tracking the trails of political biases leading to unfair NLP models}.
\newblock \bibinfo{journal}{\emph{arXiv preprint arXiv:2305.08283}} (\bibinfo{year}{2023}).
\newblock


\bibitem[Fried et~al\mbox{.}(2023)]%
        {fried2023incoder}
\bibfield{author}{\bibinfo{person}{Daniel Fried}, \bibinfo{person}{Armen Aghajanyan}, \bibinfo{person}{Jessy Lin}, \bibinfo{person}{Sida Wang}, \bibinfo{person}{Eric Wallace}, \bibinfo{person}{Freda Shi}, \bibinfo{person}{Ruiqi Zhong}, \bibinfo{person}{Wen tau Yih}, \bibinfo{person}{Luke Zettlemoyer}, {and} \bibinfo{person}{Mike Lewis}.} \bibinfo{year}{2023}\natexlab{}.
\newblock \bibinfo{title}{InCoder: A Generative Model for Code Infilling and Synthesis}.
\newblock
\newblock
\showeprint[arxiv]{2204.05999}~[cs.SE]


\bibitem[Fu et~al\mbox{.}(2023)]%
        {fu2023codeapex}
\bibfield{author}{\bibinfo{person}{Lingyue Fu}, \bibinfo{person}{Huacan Chai}, \bibinfo{person}{Shuang Luo}, \bibinfo{person}{Kounianhua Du}, \bibinfo{person}{Weiming Zhang}, \bibinfo{person}{Longteng Fan}, \bibinfo{person}{Jiayi Lei}, \bibinfo{person}{Renting Rui}, \bibinfo{person}{Jianghao Lin}, \bibinfo{person}{Yuchen Fang}, \bibinfo{person}{Yifan Liu}, \bibinfo{person}{Jingkuan Wang}, \bibinfo{person}{Siyuan Qi}, \bibinfo{person}{Kangning Zhang}, \bibinfo{person}{Weinan Zhang}, {and} \bibinfo{person}{Yong Yu}.} \bibinfo{year}{2023}\natexlab{}.
\newblock \bibinfo{title}{CodeApex: A Bilingual Programming Evaluation Benchmark for Large Language Models}.
\newblock
\newblock
\showeprint[arxiv]{2309.01940}~[cs.CL]


\bibitem[FudanUniversity(2023)]%
        {moss}
\bibfield{author}{\bibinfo{person}{FudanUniversity}.} \bibinfo{year}{2023}\natexlab{}.
\newblock \bibinfo{title}{moss}.
\newblock \bibinfo{howpublished}{\url{https://github.com/OpenLMLab/MOSS}}.
\newblock


\bibitem[Gan et~al\mbox{.}(2023)]%
        {gan2023large}
\bibfield{author}{\bibinfo{person}{Wensheng Gan}, \bibinfo{person}{Zhenlian Qi}, \bibinfo{person}{Jiayang Wu}, {and} \bibinfo{person}{Jerry Chun-Wei Lin}.} \bibinfo{year}{2023}\natexlab{}.
\newblock \showarticletitle{Large language models in education: Vision and opportunities}.
\newblock \bibinfo{journal}{\emph{arXiv preprint arXiv:2311.13160}} (\bibinfo{year}{2023}).
\newblock


\bibitem[Gao et~al\mbox{.}(2023)]%
        {gao2023g}
\bibfield{author}{\bibinfo{person}{Jiahui Gao}, \bibinfo{person}{Renjie Pi}, \bibinfo{person}{Jipeng Zhang}, \bibinfo{person}{Jiacheng Ye}, \bibinfo{person}{Wanjun Zhong}, \bibinfo{person}{Yufei Wang}, \bibinfo{person}{Lanqing Hong}, \bibinfo{person}{Jianhua Han}, \bibinfo{person}{Hang Xu}, \bibinfo{person}{Zhenguo Li}, {et~al\mbox{.}}} \bibinfo{year}{2023}\natexlab{}.
\newblock \showarticletitle{G-llava: Solving geometric problem with multi-modal large language model}.
\newblock \bibinfo{journal}{\emph{arXiv preprint arXiv:2312.11370}} (\bibinfo{year}{2023}).
\newblock


\bibitem[Geva et~al\mbox{.}(2020)]%
        {geva2020injecting}
\bibfield{author}{\bibinfo{person}{Mor Geva}, \bibinfo{person}{Ankit Gupta}, {and} \bibinfo{person}{Jonathan Berant}.} \bibinfo{year}{2020}\natexlab{}.
\newblock \showarticletitle{Injecting numerical reasoning skills into language models}.
\newblock \bibinfo{journal}{\emph{arXiv preprint arXiv:2004.04487}} (\bibinfo{year}{2020}).
\newblock


\bibitem[Ghosh and Lan(2021)]%
        {ghosh2021bobcat}
\bibfield{author}{\bibinfo{person}{Aritra Ghosh} {and} \bibinfo{person}{Andrew Lan}.} \bibinfo{year}{2021}\natexlab{}.
\newblock \showarticletitle{Bobcat: Bilevel optimization-based computerized adaptive testing}.
\newblock \bibinfo{journal}{\emph{arXiv preprint arXiv:2108.07386}} (\bibinfo{year}{2021}).
\newblock


\bibitem[Gong et~al\mbox{.}(2020)]%
        {gong2020attentional}
\bibfield{author}{\bibinfo{person}{Jibing Gong}, \bibinfo{person}{Shen Wang}, \bibinfo{person}{Jinlong Wang}, \bibinfo{person}{Wenzheng Feng}, \bibinfo{person}{Hao Peng}, \bibinfo{person}{Jie Tang}, {and} \bibinfo{person}{Philip~S Yu}.} \bibinfo{year}{2020}\natexlab{}.
\newblock \showarticletitle{Attentional graph convolutional networks for knowledge concept recommendation in moocs in a heterogeneous view}. In \bibinfo{booktitle}{\emph{Proceedings of the 43rd international ACM SIGIR conference on research and development in information retrieval}}. \bibinfo{pages}{79--88}.
\newblock


\bibitem[Gou et~al\mbox{.}(2023)]%
        {gou2023tora}
\bibfield{author}{\bibinfo{person}{Zhibin Gou}, \bibinfo{person}{Zhihong Shao}, \bibinfo{person}{Yeyun Gong}, \bibinfo{person}{Yujiu Yang}, \bibinfo{person}{Minlie Huang}, \bibinfo{person}{Nan Duan}, \bibinfo{person}{Weizhu Chen}, {et~al\mbox{.}}} \bibinfo{year}{2023}\natexlab{}.
\newblock \showarticletitle{Tora: A tool-integrated reasoning agent for mathematical problem solving}.
\newblock \bibinfo{journal}{\emph{arXiv preprint arXiv:2309.17452}} (\bibinfo{year}{2023}).
\newblock


\bibitem[Grammarly(2023)]%
        {Grammarly}
\bibfield{author}{\bibinfo{person}{Grammarly}.} \bibinfo{year}{2023}\natexlab{}.
\newblock \bibinfo{title}{Grammarly}.
\newblock
\newblock
\urldef\tempurl%
\url{https://www.grammarly.com/}
\showURL{%
\tempurl}


\bibitem[Group(2024)]%
        {mathgpturl}
\bibfield{author}{\bibinfo{person}{TAL~Education Group}.} \bibinfo{year}{2024}\natexlab{}.
\newblock \bibinfo{title}{MathGPT}.
\newblock
\newblock
\urldef\tempurl%
\url{https://www.mathgpt.com/}
\showURL{%
\tempurl}
\newblock
\shownote{2024-03-30}.


\bibitem[Guo et~al\mbox{.}(2003)]%
        {guo2003knn}
\bibfield{author}{\bibinfo{person}{Gongde Guo}, \bibinfo{person}{Hui Wang}, \bibinfo{person}{David Bell}, \bibinfo{person}{Yaxin Bi}, {and} \bibinfo{person}{Kieran Greer}.} \bibinfo{year}{2003}\natexlab{}.
\newblock \showarticletitle{KNN model-based approach in classification}. In \bibinfo{booktitle}{\emph{On The Move to Meaningful Internet Systems 2003: CoopIS, DOA, and ODBASE: OTM Confederated International Conferences, CoopIS, DOA, and ODBASE 2003, Catania, Sicily, Italy, November 3-7, 2003. Proceedings}}. Springer, \bibinfo{pages}{986--996}.
\newblock


\bibitem[Guu et~al\mbox{.}(2020)]%
        {guu2020retrieval}
\bibfield{author}{\bibinfo{person}{Kelvin Guu}, \bibinfo{person}{Kenton Lee}, \bibinfo{person}{Zora Tung}, \bibinfo{person}{Panupong Pasupat}, {and} \bibinfo{person}{Mingwei Chang}.} \bibinfo{year}{2020}\natexlab{}.
\newblock \showarticletitle{Retrieval augmented language model pre-training}. In \bibinfo{booktitle}{\emph{International conference on machine learning}}. PMLR, \bibinfo{pages}{3929--3938}.
\newblock


\bibitem[Ho et~al\mbox{.}(2022)]%
        {ho2022large}
\bibfield{author}{\bibinfo{person}{Namgyu Ho}, \bibinfo{person}{Laura Schmid}, {and} \bibinfo{person}{Se-Young Yun}.} \bibinfo{year}{2022}\natexlab{}.
\newblock \showarticletitle{Large language models are reasoning teachers}.
\newblock \bibinfo{journal}{\emph{arXiv preprint arXiv:2212.10071}} (\bibinfo{year}{2022}).
\newblock


\bibitem[Hong et~al\mbox{.}(2023)]%
        {hong2023metagpt}
\bibfield{author}{\bibinfo{person}{Sirui Hong}, \bibinfo{person}{Xiawu Zheng}, \bibinfo{person}{Jonathan Chen}, \bibinfo{person}{Yuheng Cheng}, \bibinfo{person}{Jinlin Wang}, \bibinfo{person}{Ceyao Zhang}, \bibinfo{person}{Zili Wang}, \bibinfo{person}{Steven Ka~Shing Yau}, \bibinfo{person}{Zijuan Lin}, \bibinfo{person}{Liyang Zhou}, {et~al\mbox{.}}} \bibinfo{year}{2023}\natexlab{}.
\newblock \showarticletitle{Metagpt: Meta programming for multi-agent collaborative framework}.
\newblock \bibinfo{journal}{\emph{arXiv preprint arXiv:2308.00352}} (\bibinfo{year}{2023}).
\newblock


\bibitem[Hu et~al\mbox{.}(2021)]%
        {hu2021lora}
\bibfield{author}{\bibinfo{person}{Edward~J Hu}, \bibinfo{person}{Yelong Shen}, \bibinfo{person}{Phillip Wallis}, \bibinfo{person}{Zeyuan Allen-Zhu}, \bibinfo{person}{Yuanzhi Li}, \bibinfo{person}{Shean Wang}, \bibinfo{person}{Lu Wang}, {and} \bibinfo{person}{Weizhu Chen}.} \bibinfo{year}{2021}\natexlab{}.
\newblock \showarticletitle{Lora: Low-rank adaptation of large language models}.
\newblock \bibinfo{journal}{\emph{arXiv preprint arXiv:2106.09685}} (\bibinfo{year}{2021}).
\newblock


\bibitem[Huang et~al\mbox{.}(2022)]%
        {huang2022large}
\bibfield{author}{\bibinfo{person}{Jiaxin Huang}, \bibinfo{person}{Shixiang~Shane Gu}, \bibinfo{person}{Le Hou}, \bibinfo{person}{Yuexin Wu}, \bibinfo{person}{Xuezhi Wang}, \bibinfo{person}{Hongkun Yu}, {and} \bibinfo{person}{Jiawei Han}.} \bibinfo{year}{2022}\natexlab{}.
\newblock \showarticletitle{Large language models can self-improve}.
\newblock \bibinfo{journal}{\emph{arXiv preprint arXiv:2210.11610}} (\bibinfo{year}{2022}).
\newblock


\bibitem[iFLYTEK(2024)]%
        {xunfeixinghuo}
\bibfield{author}{\bibinfo{person}{iFLYTEK}.} \bibinfo{year}{2024}\natexlab{}.
\newblock \bibinfo{title}{AutoSpark}.
\newblock
\newblock
\urldef\tempurl%
\url{https://xinghuo.xfyun.cn/}
\showURL{%
\tempurl}
\newblock
\shownote{2024-03-30}.


\bibitem[Jeronymo et~al\mbox{.}(2023)]%
        {jeronymo2023inparsv2}
\bibfield{author}{\bibinfo{person}{Vitor Jeronymo}, \bibinfo{person}{Luiz Bonifacio}, \bibinfo{person}{Hugo Abonizio}, \bibinfo{person}{Marzieh Fadaee}, \bibinfo{person}{Roberto Lotufo}, \bibinfo{person}{Jakub Zavrel}, {and} \bibinfo{person}{Rodrigo Nogueira}.} \bibinfo{year}{2023}\natexlab{}.
\newblock \bibinfo{title}{InPars-v2: Large Language Models as Efficient Dataset Generators for Information Retrieval}.
\newblock
\newblock
\showeprint[arxiv]{2301.01820}~[cs.IR]


\bibitem[Jiang et~al\mbox{.}(2023)]%
        {jiang2023multilingual}
\bibfield{author}{\bibinfo{person}{Albert~Q Jiang}, \bibinfo{person}{Wenda Li}, {and} \bibinfo{person}{Mateja Jamnik}.} \bibinfo{year}{2023}\natexlab{}.
\newblock \showarticletitle{Multilingual Mathematical Autoformalization}.
\newblock \bibinfo{journal}{\emph{arXiv preprint arXiv:2311.03755}} (\bibinfo{year}{2023}).
\newblock


\bibitem[Jiang et~al\mbox{.}(2022b)]%
        {jiang2022thor}
\bibfield{author}{\bibinfo{person}{Albert~Qiaochu Jiang}, \bibinfo{person}{Wenda Li}, \bibinfo{person}{Szymon Tworkowski}, \bibinfo{person}{Konrad Czechowski}, \bibinfo{person}{Tomasz Odrzyg{\'o}{\'z}d{\'z}}, \bibinfo{person}{Piotr Mi{\l}o{\'s}}, \bibinfo{person}{Yuhuai Wu}, {and} \bibinfo{person}{Mateja Jamnik}.} \bibinfo{year}{2022}\natexlab{b}.
\newblock \showarticletitle{Thor: Wielding hammers to integrate language models and automated theorem provers}.
\newblock \bibinfo{journal}{\emph{Advances in Neural Information Processing Systems}}  \bibinfo{volume}{35} (\bibinfo{year}{2022}), \bibinfo{pages}{8360--8373}.
\newblock


\bibitem[Jiang et~al\mbox{.}(2022c)]%
        {jiang2022draft}
\bibfield{author}{\bibinfo{person}{Albert~Q Jiang}, \bibinfo{person}{Sean Welleck}, \bibinfo{person}{Jin~Peng Zhou}, \bibinfo{person}{Wenda Li}, \bibinfo{person}{Jiacheng Liu}, \bibinfo{person}{Mateja Jamnik}, \bibinfo{person}{Timoth{\'e}e Lacroix}, \bibinfo{person}{Yuhuai Wu}, {and} \bibinfo{person}{Guillaume Lample}.} \bibinfo{year}{2022}\natexlab{c}.
\newblock \showarticletitle{Draft, sketch, and prove: Guiding formal theorem provers with informal proofs}.
\newblock \bibinfo{journal}{\emph{arXiv preprint arXiv:2210.12283}} (\bibinfo{year}{2022}).
\newblock


\bibitem[Jiang et~al\mbox{.}(2022a)]%
        {jiang2022communitylm}
\bibfield{author}{\bibinfo{person}{Hang Jiang}, \bibinfo{person}{Doug Beeferman}, \bibinfo{person}{Brandon Roy}, {and} \bibinfo{person}{Deb Roy}.} \bibinfo{year}{2022}\natexlab{a}.
\newblock \showarticletitle{CommunityLM: Probing partisan worldviews from language models}.
\newblock \bibinfo{journal}{\emph{arXiv preprint arXiv:2209.07065}} (\bibinfo{year}{2022}).
\newblock


\bibitem[Jiang et~al\mbox{.}(2021)]%
        {jiang2021can}
\bibfield{author}{\bibinfo{person}{Zhengbao Jiang}, \bibinfo{person}{Jun Araki}, \bibinfo{person}{Haibo Ding}, {and} \bibinfo{person}{Graham Neubig}.} \bibinfo{year}{2021}\natexlab{}.
\newblock \showarticletitle{How can we know when language models know? on the calibration of language models for question answering}.
\newblock \bibinfo{journal}{\emph{Transactions of the Association for Computational Linguistics}}  \bibinfo{volume}{9} (\bibinfo{year}{2021}), \bibinfo{pages}{962--977}.
\newblock


\bibitem[Kasai et~al\mbox{.}(2024)]%
        {kasai2024realtime}
\bibfield{author}{\bibinfo{person}{Jungo Kasai}, \bibinfo{person}{Keisuke Sakaguchi}, \bibinfo{person}{Ronan Le~Bras}, \bibinfo{person}{Akari Asai}, \bibinfo{person}{Xinyan Yu}, \bibinfo{person}{Dragomir Radev}, \bibinfo{person}{Noah~A Smith}, \bibinfo{person}{Yejin Choi}, \bibinfo{person}{Kentaro Inui}, {et~al\mbox{.}}} \bibinfo{year}{2024}\natexlab{}.
\newblock \showarticletitle{RealTime QA: What's the Answer Right Now?}
\newblock \bibinfo{journal}{\emph{Advances in Neural Information Processing Systems}}  \bibinfo{volume}{36} (\bibinfo{year}{2024}).
\newblock


\bibitem[Kasneci et~al\mbox{.}(2023)]%
        {kasneci2023chatgpt}
\bibfield{author}{\bibinfo{person}{Enkelejda Kasneci}, \bibinfo{person}{Kathrin Se{\ss}ler}, \bibinfo{person}{Stefan K{\"u}chemann}, \bibinfo{person}{Maria Bannert}, \bibinfo{person}{Daryna Dementieva}, \bibinfo{person}{Frank Fischer}, \bibinfo{person}{Urs Gasser}, \bibinfo{person}{Georg Groh}, \bibinfo{person}{Stephan G{\"u}nnemann}, \bibinfo{person}{Eyke H{\"u}llermeier}, {et~al\mbox{.}}} \bibinfo{year}{2023}\natexlab{}.
\newblock \showarticletitle{ChatGPT for good? On opportunities and challenges of large language models for education}.
\newblock \bibinfo{journal}{\emph{Learning and individual differences}}  \bibinfo{volume}{103} (\bibinfo{year}{2023}), \bibinfo{pages}{102274}.
\newblock


\bibitem[Kazemitabaar et~al\mbox{.}(2024)]%
        {10.1145/3631802.3631806}
\bibfield{author}{\bibinfo{person}{Majeed Kazemitabaar}, \bibinfo{person}{Xinying Hou}, \bibinfo{person}{Austin Henley}, \bibinfo{person}{Barbara~Jane Ericson}, \bibinfo{person}{David Weintrop}, {and} \bibinfo{person}{Tovi Grossman}.} \bibinfo{year}{2024}\natexlab{}.
\newblock \showarticletitle{How Novices Use LLM-based Code Generators to Solve CS1 Coding Tasks in a Self-Paced Learning Environment}. In \bibinfo{booktitle}{\emph{Proceedings of the 23rd Koli Calling International Conference on Computing Education Research}} (<conf-loc>, <city>Koli</city>, <country>Finland</country>, </conf-loc>) \emph{(\bibinfo{series}{Koli Calling '23})}. \bibinfo{publisher}{Association for Computing Machinery}, \bibinfo{address}{New York, NY, USA}, Article \bibinfo{articleno}{3}, \bibinfo{numpages}{12}~pages.
\newblock
\showISBNx{9798400716539}
\urldef\tempurl%
\url{https://doi.org/10.1145/3631802.3631806}
\showDOI{\tempurl}


\bibitem[Khandelwal et~al\mbox{.}(2019)]%
        {khandelwal2019generalization}
\bibfield{author}{\bibinfo{person}{Urvashi Khandelwal}, \bibinfo{person}{Omer Levy}, \bibinfo{person}{Dan Jurafsky}, \bibinfo{person}{Luke Zettlemoyer}, {and} \bibinfo{person}{Mike Lewis}.} \bibinfo{year}{2019}\natexlab{}.
\newblock \showarticletitle{Generalization through memorization: Nearest neighbor language models}.
\newblock \bibinfo{journal}{\emph{arXiv preprint arXiv:1911.00172}} (\bibinfo{year}{2019}).
\newblock


\bibitem[Lazaridou et~al\mbox{.}(2022)]%
        {lazaridou2022internet}
\bibfield{author}{\bibinfo{person}{Angeliki Lazaridou}, \bibinfo{person}{Elena Gribovskaya}, \bibinfo{person}{Wojciech Stokowiec}, {and} \bibinfo{person}{Nikolai Grigorev}.} \bibinfo{year}{2022}\natexlab{}.
\newblock \showarticletitle{Internet-augmented language models through few-shot prompting for open-domain question answering}.
\newblock \bibinfo{journal}{\emph{arXiv preprint arXiv:2203.05115}} (\bibinfo{year}{2022}).
\newblock


\bibitem[Lee et~al\mbox{.}(2023)]%
        {lee2023teaching}
\bibfield{author}{\bibinfo{person}{Nayoung Lee}, \bibinfo{person}{Kartik Sreenivasan}, \bibinfo{person}{Jason~D Lee}, \bibinfo{person}{Kangwook Lee}, {and} \bibinfo{person}{Dimitris Papailiopoulos}.} \bibinfo{year}{2023}\natexlab{}.
\newblock \showarticletitle{Teaching arithmetic to small transformers}.
\newblock \bibinfo{journal}{\emph{arXiv preprint arXiv:2307.03381}} (\bibinfo{year}{2023}).
\newblock


\bibitem[Lewis et~al\mbox{.}(2019)]%
        {lewis2019bart}
\bibfield{author}{\bibinfo{person}{Mike Lewis}, \bibinfo{person}{Yinhan Liu}, \bibinfo{person}{Naman Goyal}, \bibinfo{person}{Marjan Ghazvininejad}, \bibinfo{person}{Abdelrahman Mohamed}, \bibinfo{person}{Omer Levy}, \bibinfo{person}{Ves Stoyanov}, {and} \bibinfo{person}{Luke Zettlemoyer}.} \bibinfo{year}{2019}\natexlab{}.
\newblock \showarticletitle{Bart: Denoising sequence-to-sequence pre-training for natural language generation, translation, and comprehension}.
\newblock \bibinfo{journal}{\emph{arXiv preprint arXiv:1910.13461}} (\bibinfo{year}{2019}).
\newblock


\bibitem[Lewkowycz et~al\mbox{.}(2022)]%
        {lewkowycz2022solving}
\bibfield{author}{\bibinfo{person}{Aitor Lewkowycz}, \bibinfo{person}{Anders Andreassen}, \bibinfo{person}{David Dohan}, \bibinfo{person}{Ethan Dyer}, \bibinfo{person}{Henryk Michalewski}, \bibinfo{person}{Vinay Ramasesh}, \bibinfo{person}{Ambrose Slone}, \bibinfo{person}{Cem Anil}, \bibinfo{person}{Imanol Schlag}, \bibinfo{person}{Theo Gutman-Solo}, {et~al\mbox{.}}} \bibinfo{year}{2022}\natexlab{}.
\newblock \showarticletitle{Solving quantitative reasoning problems with language models, 2022}.
\newblock \bibinfo{journal}{\emph{URL https://arxiv. org/abs/2206.14858}} (\bibinfo{year}{2022}).
\newblock


\bibitem[Li et~al\mbox{.}(2023a)]%
        {li2023large}
\bibfield{author}{\bibinfo{person}{Nian Li}, \bibinfo{person}{Chen Gao}, \bibinfo{person}{Yong Li}, {and} \bibinfo{person}{Qingmin Liao}.} \bibinfo{year}{2023}\natexlab{a}.
\newblock \showarticletitle{Large language model-empowered agents for simulating macroeconomic activities}.
\newblock \bibinfo{journal}{\emph{arXiv preprint arXiv:2310.10436}} (\bibinfo{year}{2023}).
\newblock


\bibitem[Li et~al\mbox{.}(2023b)]%
        {li2023graph}
\bibfield{author}{\bibinfo{person}{Qingyao Li}, \bibinfo{person}{Wei Xia}, \bibinfo{person}{Li'ang Yin}, \bibinfo{person}{Jian Shen}, \bibinfo{person}{Renting Rui}, \bibinfo{person}{Weinan Zhang}, \bibinfo{person}{Xianyu Chen}, \bibinfo{person}{Ruiming Tang}, {and} \bibinfo{person}{Yong Yu}.} \bibinfo{year}{2023}\natexlab{b}.
\newblock \showarticletitle{Graph Enhanced Hierarchical Reinforcement Learning for Goal-oriented Learning Path Recommendation}. In \bibinfo{booktitle}{\emph{Proceedings of the 32nd ACM International Conference on Information and Knowledge Management}}. \bibinfo{pages}{1318--1327}.
\newblock


\bibitem[Liang et~al\mbox{.}(2022)]%
        {liang2022holistic}
\bibfield{author}{\bibinfo{person}{Percy Liang}, \bibinfo{person}{Rishi Bommasani}, \bibinfo{person}{Tony Lee}, \bibinfo{person}{Dimitris Tsipras}, \bibinfo{person}{Dilara Soylu}, \bibinfo{person}{Michihiro Yasunaga}, \bibinfo{person}{Yian Zhang}, \bibinfo{person}{Deepak Narayanan}, \bibinfo{person}{Yuhuai Wu}, \bibinfo{person}{Ananya Kumar}, {et~al\mbox{.}}} \bibinfo{year}{2022}\natexlab{}.
\newblock \showarticletitle{Holistic evaluation of language models}.
\newblock \bibinfo{journal}{\emph{arXiv preprint arXiv:2211.09110}} (\bibinfo{year}{2022}).
\newblock


\bibitem[Lin(2004)]%
        {lin2004rouge}
\bibfield{author}{\bibinfo{person}{Chin-Yew Lin}.} \bibinfo{year}{2004}\natexlab{}.
\newblock \showarticletitle{Rouge: A package for automatic evaluation of summaries}. In \bibinfo{booktitle}{\emph{Text summarization branches out}}. \bibinfo{pages}{74--81}.
\newblock


\bibitem[Lin et~al\mbox{.}(2023)]%
        {lin2023can}
\bibfield{author}{\bibinfo{person}{Jianghao Lin}, \bibinfo{person}{Xinyi Dai}, \bibinfo{person}{Yunjia Xi}, \bibinfo{person}{Weiwen Liu}, \bibinfo{person}{Bo Chen}, \bibinfo{person}{Xiangyang Li}, \bibinfo{person}{Chenxu Zhu}, \bibinfo{person}{Huifeng Guo}, \bibinfo{person}{Yong Yu}, \bibinfo{person}{Ruiming Tang}, {et~al\mbox{.}}} \bibinfo{year}{2023}\natexlab{}.
\newblock \showarticletitle{How Can Recommender Systems Benefit from Large Language Models: A Survey}.
\newblock \bibinfo{journal}{\emph{arXiv preprint arXiv:2306.05817}} (\bibinfo{year}{2023}).
\newblock


\bibitem[Lin et~al\mbox{.}(2021)]%
        {lin2021truthfulqa}
\bibfield{author}{\bibinfo{person}{Stephanie Lin}, \bibinfo{person}{Jacob Hilton}, {and} \bibinfo{person}{Owain Evans}.} \bibinfo{year}{2021}\natexlab{}.
\newblock \showarticletitle{Truthfulqa: Measuring how models mimic human falsehoods}.
\newblock \bibinfo{journal}{\emph{arXiv preprint arXiv:2109.07958}} (\bibinfo{year}{2021}).
\newblock


\bibitem[lingyiwanwu(2023)]%
        {yillm}
\bibfield{author}{\bibinfo{person}{lingyiwanwu}.} \bibinfo{year}{2023}\natexlab{}.
\newblock \bibinfo{title}{Yi}.
\newblock \bibinfo{howpublished}{\url{https://www.lingyiwanwu.com/}}.
\newblock


\bibitem[Liu et~al\mbox{.}(2023a)]%
        {liu2023mftcoder}
\bibfield{author}{\bibinfo{person}{Bingchang Liu}, \bibinfo{person}{Chaoyu Chen}, \bibinfo{person}{Cong Liao}, \bibinfo{person}{Zi Gong}, \bibinfo{person}{Huan Wang}, \bibinfo{person}{Zhichao Lei}, \bibinfo{person}{Ming Liang}, \bibinfo{person}{Dajun Chen}, \bibinfo{person}{Min Shen}, \bibinfo{person}{Hailian Zhou}, \bibinfo{person}{Hang Yu}, {and} \bibinfo{person}{Jianguo Li}.} \bibinfo{year}{2023}\natexlab{a}.
\newblock \bibinfo{title}{MFTCoder: Boosting Code LLMs with Multitask Fine-Tuning}.
\newblock
\newblock
\showeprint[arxiv]{2311.02303}~[cs.LG]


\bibitem[Liu et~al\mbox{.}(2023e)]%
        {liu2023fimo}
\bibfield{author}{\bibinfo{person}{Chengwu Liu}, \bibinfo{person}{Jianhao Shen}, \bibinfo{person}{Huajian Xin}, \bibinfo{person}{Zhengying Liu}, \bibinfo{person}{Ye Yuan}, \bibinfo{person}{Haiming Wang}, \bibinfo{person}{Wei Ju}, \bibinfo{person}{Chuanyang Zheng}, \bibinfo{person}{Yichun Yin}, \bibinfo{person}{Lin Li}, {et~al\mbox{.}}} \bibinfo{year}{2023}\natexlab{e}.
\newblock \showarticletitle{Fimo: A challenge formal dataset for automated theorem proving}.
\newblock \bibinfo{journal}{\emph{arXiv preprint arXiv:2309.04295}} (\bibinfo{year}{2023}).
\newblock


\bibitem[Liu et~al\mbox{.}(2023d)]%
        {liu2023retallm}
\bibfield{author}{\bibinfo{person}{Jiongnan Liu}, \bibinfo{person}{Jiajie Jin}, \bibinfo{person}{Zihan Wang}, \bibinfo{person}{Jiehan Cheng}, \bibinfo{person}{Zhicheng Dou}, {and} \bibinfo{person}{Ji-Rong Wen}.} \bibinfo{year}{2023}\natexlab{d}.
\newblock \bibinfo{title}{RETA-LLM: A Retrieval-Augmented Large Language Model Toolkit}.
\newblock
\newblock
\showeprint[arxiv]{2306.05212}~[cs.IR]


\bibitem[Liu et~al\mbox{.}(2023f)]%
        {liu2023qilin}
\bibfield{author}{\bibinfo{person}{Junling Liu}, \bibinfo{person}{Ziming Wang}, \bibinfo{person}{Qichen Ye}, \bibinfo{person}{Dading Chong}, \bibinfo{person}{Peilin Zhou}, {and} \bibinfo{person}{Yining Hua}.} \bibinfo{year}{2023}\natexlab{f}.
\newblock \showarticletitle{Qilin-Med-VL: Towards Chinese Large Vision-Language Model for General Healthcare}.
\newblock \bibinfo{journal}{\emph{arXiv preprint arXiv:2310.17956}} (\bibinfo{year}{2023}).
\newblock


\bibitem[Liu and Low(2023)]%
        {liu2023goat}
\bibfield{author}{\bibinfo{person}{Tiedong Liu} {and} \bibinfo{person}{Bryan Kian~Hsiang Low}.} \bibinfo{year}{2023}\natexlab{}.
\newblock \showarticletitle{Goat: Fine-tuned LLaMA Outperforms GPT-4 on Arithmetic Tasks}.
\newblock \bibinfo{journal}{\emph{arXiv preprint arXiv:2305.14201}} (\bibinfo{year}{2023}).
\newblock


\bibitem[Liu et~al\mbox{.}(2022a)]%
        {liu2022improving}
\bibfield{author}{\bibinfo{person}{Yixin Liu}, \bibinfo{person}{Budhaditya Deb}, \bibinfo{person}{Milagro Teruel}, \bibinfo{person}{Aaron Halfaker}, \bibinfo{person}{Dragomir Radev}, {and} \bibinfo{person}{Ahmed~H Awadallah}.} \bibinfo{year}{2022}\natexlab{a}.
\newblock \showarticletitle{On improving summarization factual consistency from natural language feedback}.
\newblock \bibinfo{journal}{\emph{arXiv preprint arXiv:2212.09968}} (\bibinfo{year}{2022}).
\newblock


\bibitem[Liu et~al\mbox{.}(2023b)]%
        {liu2023benchmarking}
\bibfield{author}{\bibinfo{person}{Yixin Liu}, \bibinfo{person}{Alexander~R Fabbri}, \bibinfo{person}{Jiawen Chen}, \bibinfo{person}{Yilun Zhao}, \bibinfo{person}{Simeng Han}, \bibinfo{person}{Shafiq Joty}, \bibinfo{person}{Pengfei Liu}, \bibinfo{person}{Dragomir Radev}, \bibinfo{person}{Chien-Sheng Wu}, {and} \bibinfo{person}{Arman Cohan}.} \bibinfo{year}{2023}\natexlab{b}.
\newblock \showarticletitle{Benchmarking generation and evaluation capabilities of large language models for instruction controllable summarization}.
\newblock \bibinfo{journal}{\emph{arXiv preprint arXiv:2311.09184}} (\bibinfo{year}{2023}).
\newblock


\bibitem[Liu et~al\mbox{.}(2023c)]%
        {liu2023learning}
\bibfield{author}{\bibinfo{person}{Yixin Liu}, \bibinfo{person}{Alexander~R Fabbri}, \bibinfo{person}{Pengfei Liu}, \bibinfo{person}{Dragomir Radev}, {and} \bibinfo{person}{Arman Cohan}.} \bibinfo{year}{2023}\natexlab{c}.
\newblock \showarticletitle{On Learning to Summarize with Large Language Models as References}.
\newblock \bibinfo{journal}{\emph{arXiv preprint arXiv:2305.14239}} (\bibinfo{year}{2023}).
\newblock


\bibitem[Liu et~al\mbox{.}(2022b)]%
        {liu2022brio}
\bibfield{author}{\bibinfo{person}{Yixin Liu}, \bibinfo{person}{Pengfei Liu}, \bibinfo{person}{Dragomir Radev}, {and} \bibinfo{person}{Graham Neubig}.} \bibinfo{year}{2022}\natexlab{b}.
\newblock \showarticletitle{BRIO: Bringing order to abstractive summarization}.
\newblock \bibinfo{journal}{\emph{arXiv preprint arXiv:2203.16804}} (\bibinfo{year}{2022}).
\newblock


\bibitem[Liventsev et~al\mbox{.}(2023)]%
        {fullyautonomousp}
\bibfield{author}{\bibinfo{person}{Vadim Liventsev}, \bibinfo{person}{Anastasiia Grishina}, \bibinfo{person}{Aki Härmä}, {and} \bibinfo{person}{Leon Moonen}.} \bibinfo{year}{2023}\natexlab{}.
\newblock \showarticletitle{Fully Autonomous Programming with Large Language Models}. In \bibinfo{booktitle}{\emph{Proceedings of the Genetic and Evolutionary Computation Conference}}. \bibinfo{publisher}{{ACM}}.
\newblock
\urldef\tempurl%
\url{https://doi.org/10.1145/3583131.3590481}
\showDOI{\tempurl}


\bibitem[Lu et~al\mbox{.}(2023)]%
        {lu2023mathvista}
\bibfield{author}{\bibinfo{person}{Pan Lu}, \bibinfo{person}{Hritik Bansal}, \bibinfo{person}{Tony Xia}, \bibinfo{person}{Jiacheng Liu}, \bibinfo{person}{Chunyuan Li}, \bibinfo{person}{Hannaneh Hajishirzi}, \bibinfo{person}{Hao Cheng}, \bibinfo{person}{Kai-Wei Chang}, \bibinfo{person}{Michel Galley}, {and} \bibinfo{person}{Jianfeng Gao}.} \bibinfo{year}{2023}\natexlab{}.
\newblock \showarticletitle{MathVista: Evaluating Mathematical Reasoning of Foundation Models in Visual Contexts}.
\newblock \bibinfo{journal}{\emph{arXiv preprint arXiv:2310.02255}} (\bibinfo{year}{2023}).
\newblock


\bibitem[Lu et~al\mbox{.}(2022a)]%
        {lu2022dynamic}
\bibfield{author}{\bibinfo{person}{Pan Lu}, \bibinfo{person}{Liang Qiu}, \bibinfo{person}{Kai-Wei Chang}, \bibinfo{person}{Ying~Nian Wu}, \bibinfo{person}{Song-Chun Zhu}, \bibinfo{person}{Tanmay Rajpurohit}, \bibinfo{person}{Peter Clark}, {and} \bibinfo{person}{Ashwin Kalyan}.} \bibinfo{year}{2022}\natexlab{a}.
\newblock \showarticletitle{Dynamic prompt learning via policy gradient for semi-structured mathematical reasoning}.
\newblock \bibinfo{journal}{\emph{arXiv preprint arXiv:2209.14610}} (\bibinfo{year}{2022}).
\newblock


\bibitem[Lu et~al\mbox{.}(2022b)]%
        {lu2022survey}
\bibfield{author}{\bibinfo{person}{Pan Lu}, \bibinfo{person}{Liang Qiu}, \bibinfo{person}{Wenhao Yu}, \bibinfo{person}{Sean Welleck}, {and} \bibinfo{person}{Kai-Wei Chang}.} \bibinfo{year}{2022}\natexlab{b}.
\newblock \showarticletitle{A survey of deep learning for mathematical reasoning}.
\newblock \bibinfo{journal}{\emph{arXiv preprint arXiv:2212.10535}} (\bibinfo{year}{2022}).
\newblock


\bibitem[Luo et~al\mbox{.}(2023a)]%
        {luo2023wizardmath}
\bibfield{author}{\bibinfo{person}{Haipeng Luo}, \bibinfo{person}{Qingfeng Sun}, \bibinfo{person}{Can Xu}, \bibinfo{person}{Pu Zhao}, \bibinfo{person}{Jianguang Lou}, \bibinfo{person}{Chongyang Tao}, \bibinfo{person}{Xiubo Geng}, \bibinfo{person}{Qingwei Lin}, \bibinfo{person}{Shifeng Chen}, {and} \bibinfo{person}{Dongmei Zhang}.} \bibinfo{year}{2023}\natexlab{a}.
\newblock \showarticletitle{Wizardmath: Empowering mathematical reasoning for large language models via reinforced evol-instruct}.
\newblock \bibinfo{journal}{\emph{arXiv preprint arXiv:2308.09583}} (\bibinfo{year}{2023}).
\newblock


\bibitem[Luo et~al\mbox{.}(2023b)]%
        {luo2023wizardcoder}
\bibfield{author}{\bibinfo{person}{Ziyang Luo}, \bibinfo{person}{Can Xu}, \bibinfo{person}{Pu Zhao}, \bibinfo{person}{Qingfeng Sun}, \bibinfo{person}{Xiubo Geng}, \bibinfo{person}{Wenxiang Hu}, \bibinfo{person}{Chongyang Tao}, \bibinfo{person}{Jing Ma}, \bibinfo{person}{Qingwei Lin}, {and} \bibinfo{person}{Daxin Jiang}.} \bibinfo{year}{2023}\natexlab{b}.
\newblock \bibinfo{title}{WizardCoder: Empowering Code Large Language Models with Evol-Instruct}.
\newblock
\newblock
\showeprint[arxiv]{2306.08568}~[cs.CL]


\bibitem[Ma et~al\mbox{.}(2023b)]%
        {ma2023hypocompass}
\bibfield{author}{\bibinfo{person}{Qianou Ma}, \bibinfo{person}{Hua Shen}, \bibinfo{person}{Kenneth Koedinger}, {and} \bibinfo{person}{Tongshuang Wu}.} \bibinfo{year}{2023}\natexlab{b}.
\newblock \bibinfo{title}{HypoCompass: Large-Language-Model-based Tutor for Hypothesis Construction in Debugging for Novices}.
\newblock
\newblock
\showeprint[arxiv]{2310.05292}~[cs.HC]


\bibitem[Ma et~al\mbox{.}(2023a)]%
        {ma2023prom}
\bibfield{author}{\bibinfo{person}{Xinbei Ma}, \bibinfo{person}{Yeyun Gong}, \bibinfo{person}{Pengcheng He}, \bibinfo{person}{Hai Zhao}, {and} \bibinfo{person}{Nan Duan}.} \bibinfo{year}{2023}\natexlab{a}.
\newblock \showarticletitle{PROM: A Phrase-level Copying Mechanism with Pre-training for Abstractive Summarization}.
\newblock \bibinfo{journal}{\emph{arXiv preprint arXiv:2305.06647}} (\bibinfo{year}{2023}).
\newblock


\bibitem[Magister et~al\mbox{.}(2022)]%
        {magister2022teaching}
\bibfield{author}{\bibinfo{person}{Lucie~Charlotte Magister}, \bibinfo{person}{Jonathan Mallinson}, \bibinfo{person}{Jakub Adamek}, \bibinfo{person}{Eric Malmi}, {and} \bibinfo{person}{Aliaksei Severyn}.} \bibinfo{year}{2022}\natexlab{}.
\newblock \showarticletitle{Teaching small language models to reason}.
\newblock \bibinfo{journal}{\emph{arXiv preprint arXiv:2212.08410}} (\bibinfo{year}{2022}).
\newblock


\bibitem[McKinzie et~al\mbox{.}(2024)]%
        {mckinzie2024mm1}
\bibfield{author}{\bibinfo{person}{Brandon McKinzie}, \bibinfo{person}{Zhe Gan}, \bibinfo{person}{Jean-Philippe Fauconnier}, \bibinfo{person}{Sam Dodge}, \bibinfo{person}{Bowen Zhang}, \bibinfo{person}{Philipp Dufter}, \bibinfo{person}{Dhruti Shah}, \bibinfo{person}{Xianzhi Du}, \bibinfo{person}{Futang Peng}, \bibinfo{person}{Floris Weers}, {et~al\mbox{.}}} \bibinfo{year}{2024}\natexlab{}.
\newblock \showarticletitle{MM1: Methods, Analysis \& Insights from Multimodal LLM Pre-training}.
\newblock \bibinfo{journal}{\emph{arXiv preprint arXiv:2403.09611}} (\bibinfo{year}{2024}).
\newblock


\bibitem[Meijer and Nering(1999)]%
        {meijer1999computerized}
\bibfield{author}{\bibinfo{person}{Rob~R Meijer} {and} \bibinfo{person}{Michael~L Nering}.} \bibinfo{year}{1999}\natexlab{}.
\newblock \showarticletitle{Computerized adaptive testing: Overview and introduction}.
\newblock \bibinfo{journal}{\emph{Applied psychological measurement}} \bibinfo{volume}{23}, \bibinfo{number}{3} (\bibinfo{year}{1999}), \bibinfo{pages}{187--194}.
\newblock


\bibitem[Meyer et~al\mbox{.}(2023)]%
        {meyer2023chatgpt}
\bibfield{author}{\bibinfo{person}{Jesse~G Meyer}, \bibinfo{person}{Ryan~J Urbanowicz}, \bibinfo{person}{Patrick~CN Martin}, \bibinfo{person}{Karen O’Connor}, \bibinfo{person}{Ruowang Li}, \bibinfo{person}{Pei-Chen Peng}, \bibinfo{person}{Tiffani~J Bright}, \bibinfo{person}{Nicholas Tatonetti}, \bibinfo{person}{Kyoung~Jae Won}, \bibinfo{person}{Graciela Gonzalez-Hernandez}, {et~al\mbox{.}}} \bibinfo{year}{2023}\natexlab{}.
\newblock \showarticletitle{ChatGPT and large language models in academia: opportunities and challenges}.
\newblock \bibinfo{journal}{\emph{BioData Mining}} \bibinfo{volume}{16}, \bibinfo{number}{1} (\bibinfo{year}{2023}), \bibinfo{pages}{20}.
\newblock


\bibitem[Mohammed et~al\mbox{.}(2023)]%
        {mohammed2023chatgpt}
\bibfield{author}{\bibinfo{person}{Osamah Mohammed}, \bibinfo{person}{Thaeer~Mueen Sahib}, \bibinfo{person}{Israa~M Hayder}, \bibinfo{person}{Sani Salisu}, \bibinfo{person}{Misbah Shahid}, {et~al\mbox{.}}} \bibinfo{year}{2023}\natexlab{}.
\newblock \showarticletitle{ChatGPT Evaluation: Can It Replace Grammarly and Quillbot Tools?}
\newblock \bibinfo{journal}{\emph{British Journal of Applied Linguistics}} \bibinfo{volume}{3}, \bibinfo{number}{2} (\bibinfo{year}{2023}), \bibinfo{pages}{34--46}.
\newblock


\bibitem[Nijkamp et~al\mbox{.}(2023)]%
        {nijkamp2023codegen}
\bibfield{author}{\bibinfo{person}{Erik Nijkamp}, \bibinfo{person}{Bo Pang}, \bibinfo{person}{Hiroaki Hayashi}, \bibinfo{person}{Lifu Tu}, \bibinfo{person}{Huan Wang}, \bibinfo{person}{Yingbo Zhou}, \bibinfo{person}{Silvio Savarese}, {and} \bibinfo{person}{Caiming Xiong}.} \bibinfo{year}{2023}\natexlab{}.
\newblock \bibinfo{title}{CodeGen: An Open Large Language Model for Code with Multi-Turn Program Synthesis}.
\newblock
\newblock
\showeprint[arxiv]{2203.13474}~[cs.LG]


\bibitem[Nipkow et~al\mbox{.}(2002)]%
        {nipkow2002isabelle}
\bibfield{author}{\bibinfo{person}{Tobias Nipkow}, \bibinfo{person}{Markus Wenzel}, {and} \bibinfo{person}{Lawrence~C Paulson}.} \bibinfo{year}{2002}\natexlab{}.
\newblock \bibinfo{booktitle}{\emph{Isabelle/HOL: a proof assistant for higher-order logic}}.
\newblock \bibinfo{publisher}{Springer}.
\newblock


\bibitem[Nye et~al\mbox{.}(2021)]%
        {nye2021show}
\bibfield{author}{\bibinfo{person}{Maxwell Nye}, \bibinfo{person}{Anders~Johan Andreassen}, \bibinfo{person}{Guy Gur-Ari}, \bibinfo{person}{Henryk Michalewski}, \bibinfo{person}{Jacob Austin}, \bibinfo{person}{David Bieber}, \bibinfo{person}{David Dohan}, \bibinfo{person}{Aitor Lewkowycz}, \bibinfo{person}{Maarten Bosma}, \bibinfo{person}{David Luan}, {et~al\mbox{.}}} \bibinfo{year}{2021}\natexlab{}.
\newblock \showarticletitle{Show your work: Scratchpads for intermediate computation with language models}.
\newblock \bibinfo{journal}{\emph{arXiv preprint arXiv:2112.00114}} (\bibinfo{year}{2021}).
\newblock


\bibitem[Omelianchuk et~al\mbox{.}(2020)]%
        {omelianchuk2020gector}
\bibfield{author}{\bibinfo{person}{Kostiantyn Omelianchuk}, \bibinfo{person}{Vitaliy Atrasevych}, \bibinfo{person}{Artem Chernodub}, {and} \bibinfo{person}{Oleksandr Skurzhanskyi}.} \bibinfo{year}{2020}\natexlab{}.
\newblock \showarticletitle{GECToR--grammatical error correction: tag, not rewrite}.
\newblock \bibinfo{journal}{\emph{arXiv preprint arXiv:2005.12592}} (\bibinfo{year}{2020}).
\newblock


\bibitem[OpenAI(2023a)]%
        {chatgpt}
\bibfield{author}{\bibinfo{person}{OpenAI}.} \bibinfo{year}{2023}\natexlab{a}.
\newblock \bibinfo{title}{chatgpt}.
\newblock \bibinfo{howpublished}{\url{https://chat.openai.com/}}.
\newblock


\bibitem[OpenAI(2023b)]%
        {openai2023gpt4}
\bibfield{author}{\bibinfo{person}{OpenAI}.} \bibinfo{year}{2023}\natexlab{b}.
\newblock \bibinfo{title}{GPT-4 Technical Report}.
\newblock
\newblock
\showeprint[arxiv]{2303.08774}~[cs.CL]


\bibitem[Ouyang et~al\mbox{.}(2022)]%
        {ouyang2022training}
\bibfield{author}{\bibinfo{person}{Long Ouyang}, \bibinfo{person}{Jeffrey Wu}, \bibinfo{person}{Xu Jiang}, \bibinfo{person}{Diogo Almeida}, \bibinfo{person}{Carroll Wainwright}, \bibinfo{person}{Pamela Mishkin}, \bibinfo{person}{Chong Zhang}, \bibinfo{person}{Sandhini Agarwal}, \bibinfo{person}{Katarina Slama}, \bibinfo{person}{Alex Ray}, {et~al\mbox{.}}} \bibinfo{year}{2022}\natexlab{}.
\newblock \showarticletitle{Training language models to follow instructions with human feedback}.
\newblock \bibinfo{journal}{\emph{Advances in neural information processing systems}}  \bibinfo{volume}{35} (\bibinfo{year}{2022}), \bibinfo{pages}{27730--27744}.
\newblock


\bibitem[Patel et~al\mbox{.}(2021)]%
        {patel2021nlp}
\bibfield{author}{\bibinfo{person}{Arkil Patel}, \bibinfo{person}{Satwik Bhattamishra}, {and} \bibinfo{person}{Navin Goyal}.} \bibinfo{year}{2021}\natexlab{}.
\newblock \showarticletitle{Are NLP models really able to solve simple math word problems?}
\newblock \bibinfo{journal}{\emph{arXiv preprint arXiv:2103.07191}} (\bibinfo{year}{2021}).
\newblock


\bibitem[Peng et~al\mbox{.}(2023b)]%
        {peng2023embeddingbased}
\bibfield{author}{\bibinfo{person}{Ruoling Peng}, \bibinfo{person}{Kang Liu}, \bibinfo{person}{Po Yang}, \bibinfo{person}{Zhipeng Yuan}, {and} \bibinfo{person}{Shunbao Li}.} \bibinfo{year}{2023}\natexlab{b}.
\newblock \bibinfo{title}{Embedding-based Retrieval with LLM for Effective Agriculture Information Extracting from Unstructured Data}.
\newblock
\newblock
\showeprint[arxiv]{2308.03107}~[cs.AI]


\bibitem[Peng et~al\mbox{.}(2023a)]%
        {peng2023geodrl}
\bibfield{author}{\bibinfo{person}{Shuai Peng}, \bibinfo{person}{Di Fu}, \bibinfo{person}{Yijun Liang}, \bibinfo{person}{Liangcai Gao}, {and} \bibinfo{person}{Zhi Tang}.} \bibinfo{year}{2023}\natexlab{a}.
\newblock \showarticletitle{GeoDRL: A Self-Learning Framework for Geometry Problem Solving using Reinforcement Learning in Deductive Reasoning}. In \bibinfo{booktitle}{\emph{Findings of the Association for Computational Linguistics: ACL 2023}}. \bibinfo{pages}{13468--13480}.
\newblock


\bibitem[Phung et~al\mbox{.}(2023)]%
        {phung2023generative}
\bibfield{author}{\bibinfo{person}{Tung Phung}, \bibinfo{person}{Victor-Alexandru Pădurean}, \bibinfo{person}{José Cambronero}, \bibinfo{person}{Sumit Gulwani}, \bibinfo{person}{Tobias Kohn}, \bibinfo{person}{Rupak Majumdar}, \bibinfo{person}{Adish Singla}, {and} \bibinfo{person}{Gustavo Soares}.} \bibinfo{year}{2023}\natexlab{}.
\newblock \bibinfo{title}{Generative AI for Programming Education: Benchmarking ChatGPT, GPT-4, and Human Tutors}.
\newblock
\newblock
\showeprint[arxiv]{2306.17156}~[cs.CY]


\bibitem[Piech et~al\mbox{.}(2015)]%
        {piech2015deep}
\bibfield{author}{\bibinfo{person}{Chris Piech}, \bibinfo{person}{Jonathan Bassen}, \bibinfo{person}{Jonathan Huang}, \bibinfo{person}{Surya Ganguli}, \bibinfo{person}{Mehran Sahami}, \bibinfo{person}{Leonidas~J Guibas}, {and} \bibinfo{person}{Jascha Sohl-Dickstein}.} \bibinfo{year}{2015}\natexlab{}.
\newblock \showarticletitle{Deep knowledge tracing}.
\newblock \bibinfo{journal}{\emph{Advances in neural information processing systems}}  \bibinfo{volume}{28} (\bibinfo{year}{2015}).
\newblock


\bibitem[Polu and Sutskever(2020)]%
        {polu2020generative}
\bibfield{author}{\bibinfo{person}{Stanislas Polu} {and} \bibinfo{person}{Ilya Sutskever}.} \bibinfo{year}{2020}\natexlab{}.
\newblock \showarticletitle{Generative language modeling for automated theorem proving}.
\newblock \bibinfo{journal}{\emph{arXiv preprint arXiv:2009.03393}} (\bibinfo{year}{2020}).
\newblock


\bibitem[Pu et~al\mbox{.}(2023)]%
        {pu2023summarization}
\bibfield{author}{\bibinfo{person}{Xiao Pu}, \bibinfo{person}{Mingqi Gao}, {and} \bibinfo{person}{Xiaojun Wan}.} \bibinfo{year}{2023}\natexlab{}.
\newblock \showarticletitle{Summarization is (almost) dead}.
\newblock \bibinfo{journal}{\emph{arXiv preprint arXiv:2309.09558}} (\bibinfo{year}{2023}).
\newblock


\bibitem[Qian et~al\mbox{.}(2023)]%
        {qian2023communicative}
\bibfield{author}{\bibinfo{person}{Chen Qian}, \bibinfo{person}{Xin Cong}, \bibinfo{person}{Cheng Yang}, \bibinfo{person}{Weize Chen}, \bibinfo{person}{Yusheng Su}, \bibinfo{person}{Juyuan Xu}, \bibinfo{person}{Zhiyuan Liu}, {and} \bibinfo{person}{Maosong Sun}.} \bibinfo{year}{2023}\natexlab{}.
\newblock \showarticletitle{Communicative agents for software development}.
\newblock \bibinfo{journal}{\emph{arXiv preprint arXiv:2307.07924}} (\bibinfo{year}{2023}).
\newblock


\bibitem[Raffel et~al\mbox{.}(2020)]%
        {raffel2020exploring}
\bibfield{author}{\bibinfo{person}{Colin Raffel}, \bibinfo{person}{Noam Shazeer}, \bibinfo{person}{Adam Roberts}, \bibinfo{person}{Katherine Lee}, \bibinfo{person}{Sharan Narang}, \bibinfo{person}{Michael Matena}, \bibinfo{person}{Yanqi Zhou}, \bibinfo{person}{Wei Li}, {and} \bibinfo{person}{Peter~J Liu}.} \bibinfo{year}{2020}\natexlab{}.
\newblock \showarticletitle{Exploring the limits of transfer learning with a unified text-to-text transformer}.
\newblock \bibinfo{journal}{\emph{The Journal of Machine Learning Research}} \bibinfo{volume}{21}, \bibinfo{number}{1} (\bibinfo{year}{2020}), \bibinfo{pages}{5485--5551}.
\newblock


\bibitem[Raheja et~al\mbox{.}(2023)]%
        {raheja2023coedit}
\bibfield{author}{\bibinfo{person}{Vipul Raheja}, \bibinfo{person}{Dhruv Kumar}, \bibinfo{person}{Ryan Koo}, {and} \bibinfo{person}{Dongyeop Kang}.} \bibinfo{year}{2023}\natexlab{}.
\newblock \showarticletitle{CoEdIT: Text Editing by Task-Specific Instruction Tuning}.
\newblock \bibinfo{journal}{\emph{arXiv preprint arXiv:2305.09857}} (\bibinfo{year}{2023}).
\newblock


\bibitem[Rajani et~al\mbox{.}(2019)]%
        {rajani2019explain}
\bibfield{author}{\bibinfo{person}{Nazneen~Fatema Rajani}, \bibinfo{person}{Bryan McCann}, \bibinfo{person}{Caiming Xiong}, {and} \bibinfo{person}{Richard Socher}.} \bibinfo{year}{2019}\natexlab{}.
\newblock \showarticletitle{Explain yourself! leveraging language models for commonsense reasoning}.
\newblock \bibinfo{journal}{\emph{arXiv preprint arXiv:1906.02361}} (\bibinfo{year}{2019}).
\newblock


\bibitem[Ren et~al\mbox{.}(2023)]%
        {ren2023investigating}
\bibfield{author}{\bibinfo{person}{Ruiyang Ren}, \bibinfo{person}{Yuhao Wang}, \bibinfo{person}{Yingqi Qu}, \bibinfo{person}{Wayne~Xin Zhao}, \bibinfo{person}{Jing Liu}, \bibinfo{person}{Hao Tian}, \bibinfo{person}{Hua Wu}, \bibinfo{person}{Ji-Rong Wen}, {and} \bibinfo{person}{Haifeng Wang}.} \bibinfo{year}{2023}\natexlab{}.
\newblock \bibinfo{title}{Investigating the Factual Knowledge Boundary of Large Language Models with Retrieval Augmentation}.
\newblock
\newblock
\showeprint[arxiv]{2307.11019}~[cs.CL]


\bibitem[Roziere et~al\mbox{.}(2023)]%
        {roziere2023code}
\bibfield{author}{\bibinfo{person}{Baptiste Roziere}, \bibinfo{person}{Jonas Gehring}, \bibinfo{person}{Fabian Gloeckle}, \bibinfo{person}{Sten Sootla}, \bibinfo{person}{Itai Gat}, \bibinfo{person}{Xiaoqing~Ellen Tan}, \bibinfo{person}{Yossi Adi}, \bibinfo{person}{Jingyu Liu}, \bibinfo{person}{Tal Remez}, \bibinfo{person}{J{\'e}r{\'e}my Rapin}, {et~al\mbox{.}}} \bibinfo{year}{2023}\natexlab{}.
\newblock \showarticletitle{Code llama: Open foundation models for code}.
\newblock \bibinfo{journal}{\emph{arXiv preprint arXiv:2308.12950}} (\bibinfo{year}{2023}).
\newblock


\bibitem[Rozière et~al\mbox{.}(2023)]%
        {rozière2023codellama}
\bibfield{author}{\bibinfo{person}{Baptiste Rozière}, \bibinfo{person}{Jonas Gehring}, \bibinfo{person}{Fabian Gloeckle}, \bibinfo{person}{Sten Sootla}, \bibinfo{person}{Itai Gat}, \bibinfo{person}{Xiaoqing~Ellen Tan}, \bibinfo{person}{Yossi Adi}, \bibinfo{person}{Jingyu Liu}, \bibinfo{person}{Tal Remez}, \bibinfo{person}{Jérémy Rapin}, \bibinfo{person}{Artyom Kozhevnikov}, \bibinfo{person}{Ivan Evtimov}, \bibinfo{person}{Joanna Bitton}, \bibinfo{person}{Manish Bhatt}, \bibinfo{person}{Cristian~Canton Ferrer}, \bibinfo{person}{Aaron Grattafiori}, \bibinfo{person}{Wenhan Xiong}, \bibinfo{person}{Alexandre Défossez}, \bibinfo{person}{Jade Copet}, \bibinfo{person}{Faisal Azhar}, \bibinfo{person}{Hugo Touvron}, \bibinfo{person}{Louis Martin}, \bibinfo{person}{Nicolas Usunier}, \bibinfo{person}{Thomas Scialom}, {and} \bibinfo{person}{Gabriel Synnaeve}.} \bibinfo{year}{2023}\natexlab{}.
\newblock \bibinfo{title}{Code Llama: Open Foundation Models for Code}.
\newblock
\newblock
\showeprint[arxiv]{2308.12950}~[cs.CL]


\bibitem[Ruis et~al\mbox{.}(2022)]%
        {ruis2022large}
\bibfield{author}{\bibinfo{person}{Laura Ruis}, \bibinfo{person}{Akbir Khan}, \bibinfo{person}{Stella Biderman}, \bibinfo{person}{Sara Hooker}, \bibinfo{person}{Tim Rockt{\"a}schel}, {and} \bibinfo{person}{Edward Grefenstette}.} \bibinfo{year}{2022}\natexlab{}.
\newblock \showarticletitle{Large language models are not zero-shot communicators}.
\newblock \bibinfo{journal}{\emph{arXiv preprint arXiv:2210.14986}} (\bibinfo{year}{2022}).
\newblock


\bibitem[Sawada et~al\mbox{.}(2023)]%
        {sawada2023arb}
\bibfield{author}{\bibinfo{person}{Tomohiro Sawada}, \bibinfo{person}{Daniel Paleka}, \bibinfo{person}{Alexander Havrilla}, \bibinfo{person}{Pranav Tadepalli}, \bibinfo{person}{Paula Vidas}, \bibinfo{person}{Alexander Kranias}, \bibinfo{person}{John~J Nay}, \bibinfo{person}{Kshitij Gupta}, {and} \bibinfo{person}{Aran Komatsuzaki}.} \bibinfo{year}{2023}\natexlab{}.
\newblock \showarticletitle{Arb: Advanced reasoning benchmark for large language models}.
\newblock \bibinfo{journal}{\emph{arXiv preprint arXiv:2307.13692}} (\bibinfo{year}{2023}).
\newblock


\bibitem[Schick et~al\mbox{.}(2024)]%
        {schick2024toolformer}
\bibfield{author}{\bibinfo{person}{Timo Schick}, \bibinfo{person}{Jane Dwivedi-Yu}, \bibinfo{person}{Roberto Dess{\`\i}}, \bibinfo{person}{Roberta Raileanu}, \bibinfo{person}{Maria Lomeli}, \bibinfo{person}{Eric Hambro}, \bibinfo{person}{Luke Zettlemoyer}, \bibinfo{person}{Nicola Cancedda}, {and} \bibinfo{person}{Thomas Scialom}.} \bibinfo{year}{2024}\natexlab{}.
\newblock \showarticletitle{Toolformer: Language models can teach themselves to use tools}.
\newblock \bibinfo{journal}{\emph{Advances in Neural Information Processing Systems}}  \bibinfo{volume}{36} (\bibinfo{year}{2024}).
\newblock


\bibitem[Shen et~al\mbox{.}(2023)]%
        {shen2023large}
\bibfield{author}{\bibinfo{person}{Chenhui Shen}, \bibinfo{person}{Liying Cheng}, \bibinfo{person}{Xuan-Phi Nguyen}, \bibinfo{person}{Yang You}, {and} \bibinfo{person}{Lidong Bing}.} \bibinfo{year}{2023}\natexlab{}.
\newblock \showarticletitle{Large language models are not yet human-level evaluators for abstractive summarization}. In \bibinfo{booktitle}{\emph{Findings of the Association for Computational Linguistics: EMNLP 2023}}. \bibinfo{pages}{4215--4233}.
\newblock


\bibitem[Sherstinsky(2020)]%
        {sherstinsky2020fundamentals}
\bibfield{author}{\bibinfo{person}{Alex Sherstinsky}.} \bibinfo{year}{2020}\natexlab{}.
\newblock \showarticletitle{Fundamentals of recurrent neural network (RNN) and long short-term memory (LSTM) network}.
\newblock \bibinfo{journal}{\emph{Physica D: Nonlinear Phenomena}}  \bibinfo{volume}{404} (\bibinfo{year}{2020}), \bibinfo{pages}{132306}.
\newblock


\bibitem[Shinn et~al\mbox{.}(2024)]%
        {shinn2024reflexion}
\bibfield{author}{\bibinfo{person}{Noah Shinn}, \bibinfo{person}{Federico Cassano}, \bibinfo{person}{Ashwin Gopinath}, \bibinfo{person}{Karthik Narasimhan}, {and} \bibinfo{person}{Shunyu Yao}.} \bibinfo{year}{2024}\natexlab{}.
\newblock \showarticletitle{Reflexion: Language agents with verbal reinforcement learning}.
\newblock \bibinfo{journal}{\emph{Advances in Neural Information Processing Systems}}  \bibinfo{volume}{36} (\bibinfo{year}{2024}).
\newblock


\bibitem[Sobania et~al\mbox{.}(2023)]%
        {sobania2023analysis}
\bibfield{author}{\bibinfo{person}{Dominik Sobania}, \bibinfo{person}{Martin Briesch}, \bibinfo{person}{Carol Hanna}, {and} \bibinfo{person}{Justyna Petke}.} \bibinfo{year}{2023}\natexlab{}.
\newblock \showarticletitle{An analysis of the automatic bug fixing performance of chatgpt}. In \bibinfo{booktitle}{\emph{2023 IEEE/ACM International Workshop on Automated Program Repair (APR)}}. IEEE, \bibinfo{pages}{23--30}.
\newblock


\bibitem[Stiennon et~al\mbox{.}(2020)]%
        {stiennon2020learning}
\bibfield{author}{\bibinfo{person}{Nisan Stiennon}, \bibinfo{person}{Long Ouyang}, \bibinfo{person}{Jeffrey Wu}, \bibinfo{person}{Daniel Ziegler}, \bibinfo{person}{Ryan Lowe}, \bibinfo{person}{Chelsea Voss}, \bibinfo{person}{Alec Radford}, \bibinfo{person}{Dario Amodei}, {and} \bibinfo{person}{Paul~F Christiano}.} \bibinfo{year}{2020}\natexlab{}.
\newblock \showarticletitle{Learning to summarize with human feedback}.
\newblock \bibinfo{journal}{\emph{Advances in Neural Information Processing Systems}}  \bibinfo{volume}{33} (\bibinfo{year}{2020}), \bibinfo{pages}{3008--3021}.
\newblock


\bibitem[Talmor et~al\mbox{.}(2018)]%
        {talmor2018commonsenseqa}
\bibfield{author}{\bibinfo{person}{Alon Talmor}, \bibinfo{person}{Jonathan Herzig}, \bibinfo{person}{Nicholas Lourie}, {and} \bibinfo{person}{Jonathan Berant}.} \bibinfo{year}{2018}\natexlab{}.
\newblock \showarticletitle{Commonsenseqa: A question answering challenge targeting commonsense knowledge}.
\newblock \bibinfo{journal}{\emph{arXiv preprint arXiv:1811.00937}} (\bibinfo{year}{2018}).
\newblock


\bibitem[Taori et~al\mbox{.}(2023)]%
        {alpacablog}
\bibfield{author}{\bibinfo{person}{Rohan Taori}, \bibinfo{person}{Ishaan Gulrajani}, \bibinfo{person}{Tianyi Zhang}, \bibinfo{person}{Yann Dubois}, \bibinfo{person}{Xuechen Li}, \bibinfo{person}{Carlos Guestrin}, \bibinfo{person}{Percy Liang}, {and} \bibinfo{person}{Tatsunori~B. Hashimoto}.} \bibinfo{year}{2023}\natexlab{}.
\newblock \bibinfo{title}{alpaca}.
\newblock \bibinfo{howpublished}{\url{https://crfm.stanford.edu/2023/03/13/alpaca.html}}.
\newblock


\bibitem[TheVicunaTeam(2023)]%
        {vicuna}
\bibfield{author}{\bibinfo{person}{TheVicunaTeam}.} \bibinfo{year}{2023}\natexlab{}.
\newblock \bibinfo{title}{vicuna}.
\newblock \bibinfo{howpublished}{\url{https://lmsys.org/blog/2023-03-30-vicuna/}}.
\newblock


\bibitem[Thompson and Weiss(2019)]%
        {thompson2019framework}
\bibfield{author}{\bibinfo{person}{Nathan~A Thompson} {and} \bibinfo{person}{David~A Weiss}.} \bibinfo{year}{2019}\natexlab{}.
\newblock \showarticletitle{A framework for the development of computerized adaptive tests}.
\newblock \bibinfo{journal}{\emph{Practical Assessment, Research, and Evaluation}} \bibinfo{volume}{16}, \bibinfo{number}{1} (\bibinfo{year}{2019}), \bibinfo{pages}{1}.
\newblock


\bibitem[TigerResearch(2023)]%
        {tigerbot}
\bibfield{author}{\bibinfo{person}{TigerResearch}.} \bibinfo{year}{2023}\natexlab{}.
\newblock \bibinfo{title}{Tigerbot}.
\newblock \bibinfo{howpublished}{\url{https://github.com/TigerResearch/TigerBot}}.
\newblock


\bibitem[Touvron et~al\mbox{.}(2023a)]%
        {touvron2023llama}
\bibfield{author}{\bibinfo{person}{Hugo Touvron}, \bibinfo{person}{Thibaut Lavril}, \bibinfo{person}{Gautier Izacard}, \bibinfo{person}{Xavier Martinet}, \bibinfo{person}{Marie-Anne Lachaux}, \bibinfo{person}{Timoth{\'e}e Lacroix}, \bibinfo{person}{Baptiste Rozi{\`e}re}, \bibinfo{person}{Naman Goyal}, \bibinfo{person}{Eric Hambro}, \bibinfo{person}{Faisal Azhar}, {et~al\mbox{.}}} \bibinfo{year}{2023}\natexlab{a}.
\newblock \showarticletitle{Llama: Open and efficient foundation language models}.
\newblock \bibinfo{journal}{\emph{arXiv preprint arXiv:2302.13971}} (\bibinfo{year}{2023}).
\newblock


\bibitem[Touvron et~al\mbox{.}(2023b)]%
        {touvron2023llama2}
\bibfield{author}{\bibinfo{person}{Hugo Touvron}, \bibinfo{person}{Louis Martin}, \bibinfo{person}{Kevin Stone}, \bibinfo{person}{Peter Albert}, \bibinfo{person}{Amjad Almahairi}, \bibinfo{person}{Yasmine Babaei}, \bibinfo{person}{Nikolay Bashlykov}, \bibinfo{person}{Soumya Batra}, \bibinfo{person}{Prajjwal Bhargava}, \bibinfo{person}{Shruti Bhosale}, {et~al\mbox{.}}} \bibinfo{year}{2023}\natexlab{b}.
\newblock \showarticletitle{Llama 2: Open foundation and fine-tuned chat models}.
\newblock \bibinfo{journal}{\emph{arXiv preprint arXiv:2307.09288}} (\bibinfo{year}{2023}).
\newblock


\bibitem[Valmeekam et~al\mbox{.}(2022)]%
        {valmeekam2022large}
\bibfield{author}{\bibinfo{person}{Karthik Valmeekam}, \bibinfo{person}{Alberto Olmo}, \bibinfo{person}{Sarath Sreedharan}, {and} \bibinfo{person}{Subbarao Kambhampati}.} \bibinfo{year}{2022}\natexlab{}.
\newblock \showarticletitle{Large Language Models Still Can't Plan (A Benchmark for LLMs on Planning and Reasoning about Change)}.
\newblock \bibinfo{journal}{\emph{arXiv preprint arXiv:2206.10498}} (\bibinfo{year}{2022}).
\newblock


\bibitem[Vu et~al\mbox{.}(2023)]%
        {vu2023freshllms}
\bibfield{author}{\bibinfo{person}{Tu Vu}, \bibinfo{person}{Mohit Iyyer}, \bibinfo{person}{Xuezhi Wang}, \bibinfo{person}{Noah Constant}, \bibinfo{person}{Jerry Wei}, \bibinfo{person}{Jason Wei}, \bibinfo{person}{Chris Tar}, \bibinfo{person}{Yun-Hsuan Sung}, \bibinfo{person}{Denny Zhou}, \bibinfo{person}{Quoc Le}, {et~al\mbox{.}}} \bibinfo{year}{2023}\natexlab{}.
\newblock \showarticletitle{Freshllms: Refreshing large language models with search engine augmentation}.
\newblock \bibinfo{journal}{\emph{arXiv preprint arXiv:2310.03214}} (\bibinfo{year}{2023}).
\newblock


\bibitem[Wang and Demszky(2023)]%
        {wang2023chatgpt}
\bibfield{author}{\bibinfo{person}{Rose~E Wang} {and} \bibinfo{person}{Dorottya Demszky}.} \bibinfo{year}{2023}\natexlab{}.
\newblock \showarticletitle{Is ChatGPT a Good Teacher Coach? Measuring Zero-Shot Performance For Scoring and Providing Actionable Insights on Classroom Instruction}.
\newblock \bibinfo{journal}{\emph{arXiv preprint arXiv:2306.03090}} (\bibinfo{year}{2023}).
\newblock


\bibitem[Wang et~al\mbox{.}(2024)]%
        {wang2024large}
\bibfield{author}{\bibinfo{person}{Shen Wang}, \bibinfo{person}{Tianlong Xu}, \bibinfo{person}{Hang Li}, \bibinfo{person}{Chaoli Zhang}, \bibinfo{person}{Joleen Liang}, \bibinfo{person}{Jiliang Tang}, \bibinfo{person}{Philip~S Yu}, {and} \bibinfo{person}{Qingsong Wen}.} \bibinfo{year}{2024}\natexlab{}.
\newblock \showarticletitle{Large Language Models for Education: A Survey and Outlook}.
\newblock \bibinfo{journal}{\emph{arXiv preprint arXiv:2403.18105}} (\bibinfo{year}{2024}).
\newblock


\bibitem[Wang et~al\mbox{.}(2023a)]%
        {wang2023scibench}
\bibfield{author}{\bibinfo{person}{Xiaoxuan Wang}, \bibinfo{person}{Ziniu Hu}, \bibinfo{person}{Pan Lu}, \bibinfo{person}{Yanqiao Zhu}, \bibinfo{person}{Jieyu Zhang}, \bibinfo{person}{Satyen Subramaniam}, \bibinfo{person}{Arjun~R Loomba}, \bibinfo{person}{Shichang Zhang}, \bibinfo{person}{Yizhou Sun}, {and} \bibinfo{person}{Wei Wang}.} \bibinfo{year}{2023}\natexlab{a}.
\newblock \showarticletitle{Scibench: Evaluating college-level scientific problem-solving abilities of large language models}.
\newblock \bibinfo{journal}{\emph{arXiv preprint arXiv:2307.10635}} (\bibinfo{year}{2023}).
\newblock


\bibitem[Wang et~al\mbox{.}(2022)]%
        {wang2022self}
\bibfield{author}{\bibinfo{person}{Xuezhi Wang}, \bibinfo{person}{Jason Wei}, \bibinfo{person}{Dale Schuurmans}, \bibinfo{person}{Quoc Le}, \bibinfo{person}{Ed Chi}, \bibinfo{person}{Sharan Narang}, \bibinfo{person}{Aakanksha Chowdhery}, {and} \bibinfo{person}{Denny Zhou}.} \bibinfo{year}{2022}\natexlab{}.
\newblock \showarticletitle{Self-consistency improves chain of thought reasoning in language models}.
\newblock \bibinfo{journal}{\emph{arXiv preprint arXiv:2203.11171}} (\bibinfo{year}{2022}).
\newblock


\bibitem[Wang et~al\mbox{.}(2023b)]%
        {wang2023codet5}
\bibfield{author}{\bibinfo{person}{Yue Wang}, \bibinfo{person}{Hung Le}, \bibinfo{person}{Akhilesh~Deepak Gotmare}, \bibinfo{person}{Nghi D.~Q. Bui}, \bibinfo{person}{Junnan Li}, {and} \bibinfo{person}{Steven C.~H. Hoi}.} \bibinfo{year}{2023}\natexlab{b}.
\newblock \bibinfo{title}{CodeT5+: Open Code Large Language Models for Code Understanding and Generation}.
\newblock
\newblock
\showeprint[arxiv]{2305.07922}~[cs.CL]


\bibitem[Wang et~al\mbox{.}(2017)]%
        {wang2017deep}
\bibfield{author}{\bibinfo{person}{Yan Wang}, \bibinfo{person}{Xiaojiang Liu}, {and} \bibinfo{person}{Shuming Shi}.} \bibinfo{year}{2017}\natexlab{}.
\newblock \showarticletitle{Deep neural solver for math word problems}. In \bibinfo{booktitle}{\emph{Proceedings of the 2017 conference on empirical methods in natural language processing}}. \bibinfo{pages}{845--854}.
\newblock


\bibitem[Wei et~al\mbox{.}(2022)]%
        {wei2022chain}
\bibfield{author}{\bibinfo{person}{Jason Wei}, \bibinfo{person}{Xuezhi Wang}, \bibinfo{person}{Dale Schuurmans}, \bibinfo{person}{Maarten Bosma}, \bibinfo{person}{Fei Xia}, \bibinfo{person}{Ed Chi}, \bibinfo{person}{Quoc~V Le}, \bibinfo{person}{Denny Zhou}, {et~al\mbox{.}}} \bibinfo{year}{2022}\natexlab{}.
\newblock \showarticletitle{Chain-of-thought prompting elicits reasoning in large language models}.
\newblock \bibinfo{journal}{\emph{Advances in Neural Information Processing Systems}}  \bibinfo{volume}{35} (\bibinfo{year}{2022}), \bibinfo{pages}{24824--24837}.
\newblock


\bibitem[Wu et~al\mbox{.}(2023)]%
        {wu2023chatgpt}
\bibfield{author}{\bibinfo{person}{Haoran Wu}, \bibinfo{person}{Wenxuan Wang}, \bibinfo{person}{Yuxuan Wan}, \bibinfo{person}{Wenxiang Jiao}, {and} \bibinfo{person}{Michael Lyu}.} \bibinfo{year}{2023}\natexlab{}.
\newblock \showarticletitle{Chatgpt or grammarly? evaluating chatgpt on grammatical error correction benchmark}.
\newblock \bibinfo{journal}{\emph{arXiv preprint arXiv:2303.13648}} (\bibinfo{year}{2023}).
\newblock


\bibitem[Wu et~al\mbox{.}(2022)]%
        {wu2022autoformalization}
\bibfield{author}{\bibinfo{person}{Yuhuai Wu}, \bibinfo{person}{Albert~Qiaochu Jiang}, \bibinfo{person}{Wenda Li}, \bibinfo{person}{Markus Rabe}, \bibinfo{person}{Charles Staats}, \bibinfo{person}{Mateja Jamnik}, {and} \bibinfo{person}{Christian Szegedy}.} \bibinfo{year}{2022}\natexlab{}.
\newblock \showarticletitle{Autoformalization with large language models}.
\newblock \bibinfo{journal}{\emph{Advances in Neural Information Processing Systems}}  \bibinfo{volume}{35} (\bibinfo{year}{2022}), \bibinfo{pages}{32353--32368}.
\newblock


\bibitem[Xia and Zhang(2023)]%
        {xia2023conversational}
\bibfield{author}{\bibinfo{person}{Chunqiu~Steven Xia} {and} \bibinfo{person}{Lingming Zhang}.} \bibinfo{year}{2023}\natexlab{}.
\newblock \showarticletitle{Conversational automated program repair}.
\newblock \bibinfo{journal}{\emph{arXiv preprint arXiv:2301.13246}} (\bibinfo{year}{2023}).
\newblock


\bibitem[Xin et~al\mbox{.}(2023)]%
        {xin2023lego}
\bibfield{author}{\bibinfo{person}{Huajian Xin}, \bibinfo{person}{Haiming Wang}, \bibinfo{person}{Chuanyang Zheng}, \bibinfo{person}{Lin Li}, \bibinfo{person}{Zhengying Liu}, \bibinfo{person}{Qingxing Cao}, \bibinfo{person}{Yinya Huang}, \bibinfo{person}{Jing Xiong}, \bibinfo{person}{Han Shi}, \bibinfo{person}{Enze Xie}, {et~al\mbox{.}}} \bibinfo{year}{2023}\natexlab{}.
\newblock \showarticletitle{LEGO-Prover: Neural Theorem Proving with Growing Libraries}.
\newblock \bibinfo{journal}{\emph{arXiv preprint arXiv:2310.00656}} (\bibinfo{year}{2023}).
\newblock


\bibitem[Xiong et~al\mbox{.}(2023)]%
        {xiong2023trigo}
\bibfield{author}{\bibinfo{person}{Jing Xiong}, \bibinfo{person}{Jianhao Shen}, \bibinfo{person}{Ye Yuan}, \bibinfo{person}{Haiming Wang}, \bibinfo{person}{Yichun Yin}, \bibinfo{person}{Zhengying Liu}, \bibinfo{person}{Lin Li}, \bibinfo{person}{Zhijiang Guo}, \bibinfo{person}{Qingxing Cao}, \bibinfo{person}{Yinya Huang}, {et~al\mbox{.}}} \bibinfo{year}{2023}\natexlab{}.
\newblock \showarticletitle{TRIGO: Benchmarking Formal Mathematical Proof Reduction for Generative Language Models}.
\newblock \bibinfo{journal}{\emph{arXiv preprint arXiv:2310.10180}} (\bibinfo{year}{2023}).
\newblock


\bibitem[Xu et~al\mbox{.}(2023)]%
        {xu2023wizardlm}
\bibfield{author}{\bibinfo{person}{Can Xu}, \bibinfo{person}{Qingfeng Sun}, \bibinfo{person}{Kai Zheng}, \bibinfo{person}{Xiubo Geng}, \bibinfo{person}{Pu Zhao}, \bibinfo{person}{Jiazhan Feng}, \bibinfo{person}{Chongyang Tao}, {and} \bibinfo{person}{Daxin Jiang}.} \bibinfo{year}{2023}\natexlab{}.
\newblock \bibinfo{title}{WizardLM: Empowering Large Language Models to Follow Complex Instructions}.
\newblock
\newblock
\showeprint[arxiv]{2304.12244}~[cs.CL]


\bibitem[Xu et~al\mbox{.}(2024)]%
        {xu2024hallucination}
\bibfield{author}{\bibinfo{person}{Ziwei Xu}, \bibinfo{person}{Sanjay Jain}, {and} \bibinfo{person}{Mohan Kankanhalli}.} \bibinfo{year}{2024}\natexlab{}.
\newblock \showarticletitle{Hallucination is inevitable: An innate limitation of large language models}.
\newblock \bibinfo{journal}{\emph{arXiv preprint arXiv:2401.11817}} (\bibinfo{year}{2024}).
\newblock


\bibitem[Yang et~al\mbox{.}(2023c)]%
        {yang2023baichuan}
\bibfield{author}{\bibinfo{person}{Aiyuan Yang}, \bibinfo{person}{Bin Xiao}, \bibinfo{person}{Bingning Wang}, \bibinfo{person}{Borong Zhang}, \bibinfo{person}{Chao Yin}, \bibinfo{person}{Chenxu Lv}, \bibinfo{person}{Da Pan}, \bibinfo{person}{Dian Wang}, \bibinfo{person}{Dong Yan}, \bibinfo{person}{Fan Yang}, {et~al\mbox{.}}} \bibinfo{year}{2023}\natexlab{c}.
\newblock \showarticletitle{Baichuan 2: Open large-scale language models}.
\newblock \bibinfo{journal}{\emph{arXiv preprint arXiv:2309.10305}} (\bibinfo{year}{2023}).
\newblock


\bibitem[Yang et~al\mbox{.}(2023b)]%
        {yang2023leandojo}
\bibfield{author}{\bibinfo{person}{Kaiyu Yang}, \bibinfo{person}{Aidan~M Swope}, \bibinfo{person}{Alex Gu}, \bibinfo{person}{Rahul Chalamala}, \bibinfo{person}{Peiyang Song}, \bibinfo{person}{Shixing Yu}, \bibinfo{person}{Saad Godil}, \bibinfo{person}{Ryan Prenger}, {and} \bibinfo{person}{Anima Anandkumar}.} \bibinfo{year}{2023}\natexlab{b}.
\newblock \showarticletitle{Leandojo: Theorem proving with retrieval-augmented language models}.
\newblock \bibinfo{journal}{\emph{arXiv preprint arXiv:2306.15626}} (\bibinfo{year}{2023}).
\newblock


\bibitem[Yang et~al\mbox{.}(2023a)]%
        {yang2023gpt}
\bibfield{author}{\bibinfo{person}{Zhen Yang}, \bibinfo{person}{Ming Ding}, \bibinfo{person}{Qingsong Lv}, \bibinfo{person}{Zhihuan Jiang}, \bibinfo{person}{Zehai He}, \bibinfo{person}{Yuyi Guo}, \bibinfo{person}{Jinfeng Bai}, {and} \bibinfo{person}{Jie Tang}.} \bibinfo{year}{2023}\natexlab{a}.
\newblock \showarticletitle{GPT Can Solve Mathematical Problems Without a Calculator}.
\newblock \bibinfo{journal}{\emph{arXiv preprint arXiv:2309.03241}} (\bibinfo{year}{2023}).
\newblock


\bibitem[Yao et~al\mbox{.}(2024)]%
        {yao2024tree}
\bibfield{author}{\bibinfo{person}{Shunyu Yao}, \bibinfo{person}{Dian Yu}, \bibinfo{person}{Jeffrey Zhao}, \bibinfo{person}{Izhak Shafran}, \bibinfo{person}{Tom Griffiths}, \bibinfo{person}{Yuan Cao}, {and} \bibinfo{person}{Karthik Narasimhan}.} \bibinfo{year}{2024}\natexlab{}.
\newblock \showarticletitle{Tree of thoughts: Deliberate problem solving with large language models}.
\newblock \bibinfo{journal}{\emph{Advances in Neural Information Processing Systems}}  \bibinfo{volume}{36} (\bibinfo{year}{2024}).
\newblock


\bibitem[Yasunaga et~al\mbox{.}(2021)]%
        {yasunaga2021qa}
\bibfield{author}{\bibinfo{person}{Michihiro Yasunaga}, \bibinfo{person}{Hongyu Ren}, \bibinfo{person}{Antoine Bosselut}, \bibinfo{person}{Percy Liang}, {and} \bibinfo{person}{Jure Leskovec}.} \bibinfo{year}{2021}\natexlab{}.
\newblock \showarticletitle{QA-GNN: Reasoning with language models and knowledge graphs for question answering}.
\newblock \bibinfo{journal}{\emph{arXiv preprint arXiv:2104.06378}} (\bibinfo{year}{2021}).
\newblock


\bibitem[Ye et~al\mbox{.}(2023)]%
        {ye2023mplug}
\bibfield{author}{\bibinfo{person}{Qinghao Ye}, \bibinfo{person}{Haiyang Xu}, \bibinfo{person}{Jiabo Ye}, \bibinfo{person}{Ming Yan}, \bibinfo{person}{Haowei Liu}, \bibinfo{person}{Qi Qian}, \bibinfo{person}{Ji Zhang}, \bibinfo{person}{Fei Huang}, {and} \bibinfo{person}{Jingren Zhou}.} \bibinfo{year}{2023}\natexlab{}.
\newblock \showarticletitle{mPLUG-Owl2: Revolutionizing Multi-modal Large Language Model with Modality Collaboration}.
\newblock \bibinfo{journal}{\emph{arXiv preprint arXiv:2311.04257}} (\bibinfo{year}{2023}).
\newblock


\bibitem[Yu et~al\mbox{.}(2023)]%
        {yu2023metamath}
\bibfield{author}{\bibinfo{person}{Longhui Yu}, \bibinfo{person}{Weisen Jiang}, \bibinfo{person}{Han Shi}, \bibinfo{person}{Jincheng Yu}, \bibinfo{person}{Zhengying Liu}, \bibinfo{person}{Yu Zhang}, \bibinfo{person}{James~T Kwok}, \bibinfo{person}{Zhenguo Li}, \bibinfo{person}{Adrian Weller}, {and} \bibinfo{person}{Weiyang Liu}.} \bibinfo{year}{2023}\natexlab{}.
\newblock \showarticletitle{Metamath: Bootstrap your own mathematical questions for large language models}.
\newblock \bibinfo{journal}{\emph{arXiv preprint arXiv:2309.12284}} (\bibinfo{year}{2023}).
\newblock


\bibitem[Yuan et~al\mbox{.}(2023)]%
        {yuan2023well}
\bibfield{author}{\bibinfo{person}{Zheng Yuan}, \bibinfo{person}{Hongyi Yuan}, \bibinfo{person}{Chuanqi Tan}, \bibinfo{person}{Wei Wang}, {and} \bibinfo{person}{Songfang Huang}.} \bibinfo{year}{2023}\natexlab{}.
\newblock \showarticletitle{How well do Large Language Models perform in Arithmetic tasks?}
\newblock \bibinfo{journal}{\emph{arXiv preprint arXiv:2304.02015}} (\bibinfo{year}{2023}).
\newblock


\bibitem[Zelikman et~al\mbox{.}(2022)]%
        {zelikman2022star}
\bibfield{author}{\bibinfo{person}{Eric Zelikman}, \bibinfo{person}{Yuhuai Wu}, \bibinfo{person}{Jesse Mu}, {and} \bibinfo{person}{Noah~D Goodman}.} \bibinfo{year}{2022}\natexlab{}.
\newblock \showarticletitle{Star: Bootstrapping reasoning with reasoning, 2022}.
\newblock \bibinfo{journal}{\emph{URL https://arxiv. org/abs/2203.14465}} (\bibinfo{year}{2022}).
\newblock


\bibitem[Zeng et~al\mbox{.}(2022)]%
        {zeng2022glm}
\bibfield{author}{\bibinfo{person}{Aohan Zeng}, \bibinfo{person}{Xiao Liu}, \bibinfo{person}{Zhengxiao Du}, \bibinfo{person}{Zihan Wang}, \bibinfo{person}{Hanyu Lai}, \bibinfo{person}{Ming Ding}, \bibinfo{person}{Zhuoyi Yang}, \bibinfo{person}{Yifan Xu}, \bibinfo{person}{Wendi Zheng}, \bibinfo{person}{Xiao Xia}, {et~al\mbox{.}}} \bibinfo{year}{2022}\natexlab{}.
\newblock \showarticletitle{Glm-130b: An open bilingual pre-trained model}.
\newblock \bibinfo{journal}{\emph{arXiv preprint arXiv:2210.02414}} (\bibinfo{year}{2022}).
\newblock


\bibitem[Zhang et~al\mbox{.}(2023b)]%
        {zhang2023evaluating}
\bibfield{author}{\bibinfo{person}{Beichen Zhang}, \bibinfo{person}{Kun Zhou}, \bibinfo{person}{Xilin Wei}, \bibinfo{person}{Wayne~Xin Zhao}, \bibinfo{person}{Jing Sha}, \bibinfo{person}{Shijin Wang}, {and} \bibinfo{person}{Ji-Rong Wen}.} \bibinfo{year}{2023}\natexlab{b}.
\newblock \showarticletitle{Evaluating and Improving Tool-Augmented Computation-Intensive Math Reasoning}.
\newblock \bibinfo{journal}{\emph{arXiv preprint arXiv:2306.02408}} (\bibinfo{year}{2023}).
\newblock


\bibitem[Zhang et~al\mbox{.}(2024)]%
        {zhang2024geoeval}
\bibfield{author}{\bibinfo{person}{Jiaxin Zhang}, \bibinfo{person}{Zhongzhi Li}, \bibinfo{person}{Mingliang Zhang}, \bibinfo{person}{Fei Yin}, \bibinfo{person}{Chenglin Liu}, {and} \bibinfo{person}{Yashar Moshfeghi}.} \bibinfo{year}{2024}\natexlab{}.
\newblock \showarticletitle{GeoEval: Benchmark for Evaluating LLMs and Multi-Modal Models on Geometry Problem-Solving}.
\newblock \bibinfo{journal}{\emph{arXiv preprint arXiv:2402.10104}} (\bibinfo{year}{2024}).
\newblock


\bibitem[Zhang et~al\mbox{.}(2023a)]%
        {zhang2023multi}
\bibfield{author}{\bibinfo{person}{Ming-Liang Zhang}, \bibinfo{person}{Fei Yin}, {and} \bibinfo{person}{Cheng-Lin Liu}.} \bibinfo{year}{2023}\natexlab{a}.
\newblock \showarticletitle{A Multi-Modal Neural Geometric Solver with Textual Clauses Parsed from Diagram}.
\newblock \bibinfo{journal}{\emph{arXiv preprint arXiv:2302.11097}} (\bibinfo{year}{2023}).
\newblock


\bibitem[Zhang et~al\mbox{.}(2023b)]%
        {zhang2023retrieve}
\bibfield{author}{\bibinfo{person}{Peitian Zhang}, \bibinfo{person}{Shitao Xiao}, \bibinfo{person}{Zheng Liu}, \bibinfo{person}{Zhicheng Dou}, {and} \bibinfo{person}{Jian-Yun Nie}.} \bibinfo{year}{2023}\natexlab{b}.
\newblock \bibinfo{title}{Retrieve Anything To Augment Large Language Models}.
\newblock
\newblock
\showeprint[arxiv]{2310.07554}~[cs.IR]


\bibitem[Zhang et~al\mbox{.}(2023a)]%
        {zhang2023planning}
\bibfield{author}{\bibinfo{person}{Shun Zhang}, \bibinfo{person}{Zhenfang Chen}, \bibinfo{person}{Yikang Shen}, \bibinfo{person}{Mingyu Ding}, \bibinfo{person}{Joshua~B Tenenbaum}, {and} \bibinfo{person}{Chuang Gan}.} \bibinfo{year}{2023}\natexlab{a}.
\newblock \showarticletitle{Planning with large language models for code generation}.
\newblock \bibinfo{journal}{\emph{arXiv preprint arXiv:2303.05510}} (\bibinfo{year}{2023}).
\newblock


\bibitem[zhang et~al\mbox{.}(2023)]%
        {zhang2023}
\bibfield{author}{\bibinfo{person}{Sabrina zhang}, \bibinfo{person}{Daksha Yadav}, {and} \bibinfo{person}{Tom Jin}.} \bibinfo{year}{2023}\natexlab{}.
\newblock \showarticletitle{Cash transaction booking via retrieval augmented LLM}. In \bibinfo{booktitle}{\emph{KDD 2023 Workshop on Robust NLP for Finance (RobustFin)}}.
\newblock
\urldef\tempurl%
\url{https://www.amazon.science/publications/cash-transaction-booking-via-retrieval-augmented-llm}
\showURL{%
\tempurl}


\bibitem[Zhou et~al\mbox{.}(2023a)]%
        {zhou2023solving}
\bibfield{author}{\bibinfo{person}{Aojun Zhou}, \bibinfo{person}{Ke Wang}, \bibinfo{person}{Zimu Lu}, \bibinfo{person}{Weikang Shi}, \bibinfo{person}{Sichun Luo}, \bibinfo{person}{Zipeng Qin}, \bibinfo{person}{Shaoqing Lu}, \bibinfo{person}{Anya Jia}, \bibinfo{person}{Linqi Song}, \bibinfo{person}{Mingjie Zhan}, {et~al\mbox{.}}} \bibinfo{year}{2023}\natexlab{a}.
\newblock \showarticletitle{Solving challenging math word problems using gpt-4 code interpreter with code-based self-verification}.
\newblock \bibinfo{journal}{\emph{arXiv preprint arXiv:2308.07921}} (\bibinfo{year}{2023}).
\newblock


\bibitem[Zhou et~al\mbox{.}(2023b)]%
        {zhou2023language}
\bibfield{author}{\bibinfo{person}{Andy Zhou}, \bibinfo{person}{Kai Yan}, \bibinfo{person}{Michal Shlapentokh-Rothman}, \bibinfo{person}{Haohan Wang}, {and} \bibinfo{person}{Yu-Xiong Wang}.} \bibinfo{year}{2023}\natexlab{b}.
\newblock \showarticletitle{Language agent tree search unifies reasoning acting and planning in language models}.
\newblock \bibinfo{journal}{\emph{arXiv preprint arXiv:2310.04406}} (\bibinfo{year}{2023}).
\newblock


\bibitem[Zhou et~al\mbox{.}(2022b)]%
        {zhou2022least}
\bibfield{author}{\bibinfo{person}{Denny Zhou}, \bibinfo{person}{Nathanael Sch{\"a}rli}, \bibinfo{person}{Le Hou}, \bibinfo{person}{Jason Wei}, \bibinfo{person}{Nathan Scales}, \bibinfo{person}{Xuezhi Wang}, \bibinfo{person}{Dale Schuurmans}, \bibinfo{person}{Claire Cui}, \bibinfo{person}{Olivier Bousquet}, \bibinfo{person}{Quoc Le}, {et~al\mbox{.}}} \bibinfo{year}{2022}\natexlab{b}.
\newblock \showarticletitle{Least-to-most prompting enables complex reasoning in large language models}.
\newblock \bibinfo{journal}{\emph{arXiv preprint arXiv:2205.10625}} (\bibinfo{year}{2022}).
\newblock


\bibitem[Zhou et~al\mbox{.}(2022a)]%
        {zhou2022docprompting}
\bibfield{author}{\bibinfo{person}{Shuyan Zhou}, \bibinfo{person}{Uri Alon}, \bibinfo{person}{Frank~F Xu}, \bibinfo{person}{Zhiruo Wang}, \bibinfo{person}{Zhengbao Jiang}, {and} \bibinfo{person}{Graham Neubig}.} \bibinfo{year}{2022}\natexlab{a}.
\newblock \showarticletitle{Docprompting: Generating code by retrieving the docs}.
\newblock \bibinfo{journal}{\emph{arXiv preprint arXiv:2207.05987}} (\bibinfo{year}{2022}).
\newblock


\end{thebibliography}

\appendix

\end{document}